%% file: neurips_2026.tex
\documentclass{article}

\PassOptionsToPackage{numbers, compress}{natbib}
\usepackage[main, final]{neurips_2026}
\usepackage{times}
\usepackage{amsthm}
\usepackage{float}
\usepackage{tablefootnote}
\usepackage{fvextra}

\usepackage{amsmath}
\usepackage[utf8]{inputenc} 
\usepackage[T1]{fontenc}    
\usepackage{hyperref}       
\usepackage{url}            
\usepackage{booktabs}       
\usepackage{amsfonts}       
\usepackage{nicefrac}       
\usepackage{microtype}      
\usepackage[table]{xcolor}
\usepackage{xcolor}         
\usepackage{graphicx}
\usepackage{wrapfig}
\usepackage{subcaption}
\usepackage{enumitem}
\usepackage{adjustbox}
\usepackage{multirow}
\usepackage{tcolorbox}
\tcbuselibrary{breakable,listings}
\usepackage{float}
\usepackage{etoc}

\definecolor{boxteal}{RGB}{0,128,128}
\definecolor{maroon}{RGB}{128,0,0}
\definecolor{violetborder}{HTML}{874FFF}
\definecolor{violetfill}{HTML}{DCCCFF}
\definecolor{pinkborder}{HTML}{F849C1}
\definecolor{pinkfill}{HTML}{FFC2EC}
\definecolor{brownborder}{HTML}{9C4F00}
\definecolor{brownfill}{HTML}{EED3B3}

\title{TSFM Meets LLM: \underline{\textbf{Co}}ntext as \underline{\textbf{Co}}variates}

\author{%
  Hemanth Ram\thanks{Equal Contribution} \\
  \And
  Sai Shankar Narasimhan\footnotemark[1] \\
  \And
  Shubhankar Agarwal \\
  \And
  Sandeep Chinchali \\
  \And
  \normalfont The University of Texas at Austin, Austin, TX, 78712. \\
  \texttt{\{hemanthramgk,nsaishankar,somi.agarwal,sandeepc\}@utexas.edu}
}

\begin{document}

\theoremstyle{plain}
\newtheorem{theorem}{Theorem}[section]
\newtheorem{proposition}[theorem]{Proposition}
\newtheorem{lemma}[theorem]{Lemma}
\newtheorem{corollary}[theorem]{Corollary}
\theoremstyle{definition}
\newtheorem{definition}[theorem]{Definition}
\newtheorem{assumption}[theorem]{Assumption}
\theoremstyle{remark}
\newtheorem{remark}[theorem]{Remark}

\tcbset{
  theostyle/.style={
    colback=pinkfill!25!white,   
    colframe=pinkborder!100!black, 
    coltitle=black          
  }
}

\newtcolorbox{theorembox}[1][]{
  theostyle,
  title=#1
}

\maketitle
\setcounter{footnote}{0}

\input{abstract}

\etocdepthtag.toc{mtchapters}

\input{sections/1_introduction}
\input{sections/2_related_works}
\input{sections/3_problem_formulation}

\input{sections/4_synthetic_data_generation}

\input{sections/5_model_arch}
\input{sections/6_experiments}

{
\small
\bibliographystyle{unsrtnat}
\bibliography{references/external}
}

\newpage

\input{sections/appendix}

\input{checklist.tex}

\end{document}

%% file: abstract.tex
\begin{abstract}
Textual information is ubiquitous in real-world time series datasets, appearing as news, reports, and annotations that hold key predictive signals not deducible from numerical history alone. Large Language Models (LLMs) interpret text but lack numerical precision, while Time Series Foundation Models (TSFMs) are quantitatively accurate but cannot reason about the effects of text. Moreover, the lack of large-scale paired multimodal datasets limits progress in bridging this gap. In this work, we address this challenge by providing a scalable approach for generating realistic multimodal datasets by annotating real-world time series with LLMs. Building on this, we propose \textbf{Co}ntext as \textbf{Co}variates (CoCo), a forecasting framework in which an LLM distills its textual reasoning into numerically grounded forecast and confidence covariates that guide a TSFM backbone, thereby exploiting the strengths of both LLMs and TSFMs. We align the two models in stages, first fine-tuning the LLM via Group Relative Policy Optimization (GRPO) with a reward tied to covariate-induced improvement in forecasting performance, and then fine-tuning the TSFM on covariates generated by the aligned LLM. Our experiments show that models trained on our synthetic multimodal corpus generalize to real benchmarks spanning finance, economics, weather, traffic, and security. Notably, CoCo with \texttt{Qwen3-4B-Instruct-2507} as the LLM and Chronos2 as the TSFM achieves up to 12\% MAE improvement over state-of-the-art TSFMs and LLM forecasters, such as GPT\text{-}4o. Code\footnote{https://github.com/UTAustin-SwarmLab/CoCo}, checkpoints, and the evaluation datasets\footnote{https://huggingface.co/collections/SWARM-Lab/coco} are publicly available.
\end{abstract}

%% file: sections/1_introduction.tex
\section{Introduction}
\label{sec:introduction}

\begin{center}
\emph{``\textbf{Forecast Apple's stock price if revenue from iPhone sales is 15\% higher than the target}''}\\[2pt]
\emph{``\textbf{Forecast crude oil prices given the recent tariffs on the OPEC countries}''}
\end{center}

These contextual forecasting examples across domains show that the textual context is deeply 
intertwined with real-world time series and can provide signals about future changes that 
are not captured by past numerical values alone (check Figure \ref{fig:selling_qualitative}). For example, in the finance domain, analyst notes, 
government policies, and even social media sentiments influence market dynamics, 
and are commonly used to predict stock price movements \citep{fin1, fin2, fin3, fin4}. 
Similarly, in consumer markets, online reviews and social media discussions are used to forecast future product 
demand \citep{consumer1, consumer2}. 

Despite the availability of textual information, the state-of-the-art (SOTA) univariate 
(TimesFM \citep{timesfm}) and multivariate (Chronos2 \citep{chronos2}) 
time series foundation models (TSFMs) operate solely on unimodal numerical inputs. Figure \ref{fig:rank_mae_plot} shows that incorporating text embeddings into SOTA multivariate forecasting models does not necessarily yield multimodal gains, even when the text clearly contains useful information. Given the substantive progress in the 
understanding and reasoning capabilities of large language models (LLMs) 
\citep{qwen3, openai2024gpt4technicalreport}, we address the following key question in this work: 
\emph{How can we leverage LLMs to process textual annotations or context and improve the performance of TSFMs?} Current efforts to build such context-aware TSFMs struggle with two major barriers:

\begin{itemize}[leftmargin=18pt, itemsep=1pt]
    \item \textbf{Insufficient Multi-Domain Multimodal Training Data:} Learning complex and lagged relationships between text and future numerical values requires large volumes of aligned multimodal data from different domains. However, existing multi-domain datasets such as TimeMMD \citep{timemmd} and Context-Is-Key \citep{williams2025context} contain only a few thousand aligned samples. Large-scale datasets \citep{fnspid} are heavily restricted to a single sub-domain, limiting cross-domain generalization. 

    \item \textbf{LLM-Centric Multimodal Forecasting approaches neglect the predictive power of TSFMs:} Numerous prior works \citep{park2025revisitingllmszeroshottimeseries, yu2023temporal, wang2024news, hybridmmf, jin2024timellmtimeseriesforecasting, captime, jiang2025explainable} serialize time series as tokens and rely primarily on LLM prediction. While this enables the use of textual context, it fails to leverage the forecasting inductive biases and uncertainty modeling strengths of dedicated TSFMs. 

\end{itemize}

\input{figures_latex/intro}

Motivated by these gaps, we make three contributions.

\begin{itemize}[leftmargin=18pt, itemsep=2pt, topsep=2pt]
    \item \textbf{Scalable Synthetic Multimodal Data.} We introduce a pipeline that annotates real-world time series with LLM-generated text, producing large-scale, text-aligned datasets with varied lagged correlations. We empirically show that augmenting scarce and often domain-restricted real multimodal data with synthetic annotations yields strong zero-shot forecasting performance.

    \item \textbf{CoCo Architecture and Staged Fine-tuning.} We propose CoCo (\textbf{Co}ntext as \textbf{Co}variates), where an LLM distills text into forecast and confidence covariates that guide a Chronos2 backbone. The LLM is fine-tuned via Group Relative Policy Optimization (GRPO) using a reward tied to the forecasting performance, after which Chronos2 is fine-tuned with the LLM-generated covariates. 
    

    \item \textbf{Multimodal Forecasting Gains.} CoCo outperforms SOTA TSFMs and LLMs across 30 datasets and 95 settings (Figure \ref{fig:rank_mae_plot}) spanning finance, weather, macroeconomics, traffic, and security domains, achieving up to 70\% improvement over GPT-4o in traffic forecasting (Figure \ref{fig:selling_quantitative}).
\end{itemize}

    



%% file: figures_latex/intro.tex
\begin{figure}[t]
    \centering
    \includegraphics[width=0.9\textwidth]{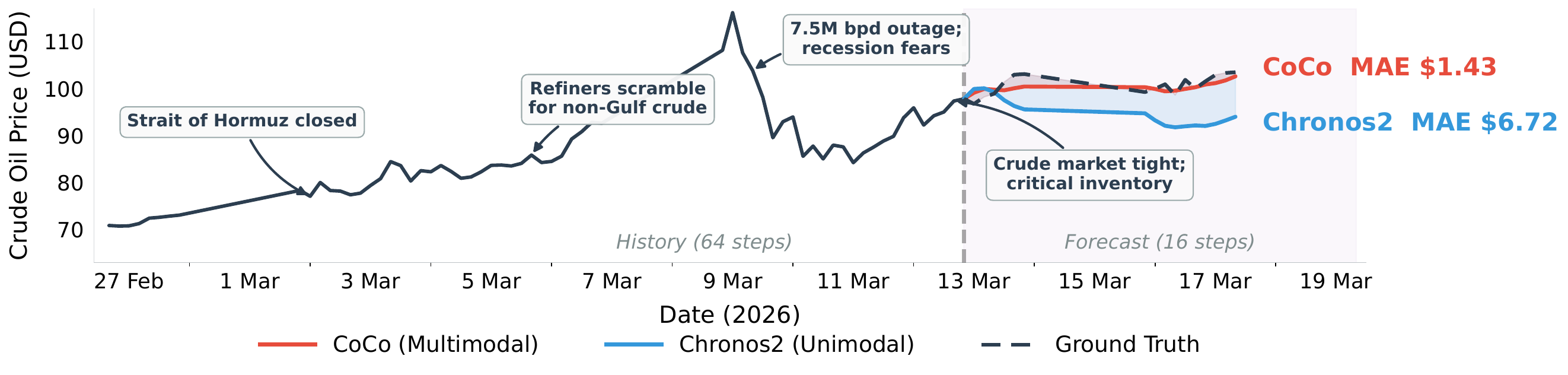}
    \caption{
    \textbf{Effective use of context is essential for accurate time series forecasting.} In crude oil price forecasting, textual context (e.g., recession fears on March 9) aligns with sudden price shifts. Our approach, \textbf{Co}ntext as \textbf{Co}variates (CoCo), leverages this context to produce accurate forecasts, while unimodal models such as Chronos2 fail to capture context-induced trends.}
    \label{fig:selling_qualitative}
    \vspace{-0.2in}
\end{figure}

%% file: sections/2_related_works.tex
\section{Related Works}
\label{sec:related_works}
\noindent\textbf{Multimodal datasets for text-aided forecasting.}
Prior works largely study \emph{text-aided} time series prediction, where auxiliary textual context improves forecasts across domains \citep{liu2024echo, sawhney2021fast, wang2024news, xu2018stock, wu2018hybrid, chen2023chatgpt}.
However, the nature of text supervision varies significantly across domains. For example, finance offers abundant dynamic or \emph{time-varying} annotations (\textit{e.g.,} news/social media/search trends) enabling large-scale datasets for stock price and movement predictions \citep{fnspid, liu2024echo, sawhney2021fast, wang2024news, xu2018stock, wu2018hybrid, chen2023chatgpt}. On the other hand, healthcare datasets typically provide \emph{static/global} patient information in the form of medical records aligned with physiological signals and are commonly used for classification or anomaly detection \citep{wang2020mimic, smigiel2021ecg}.
Other settings (\textit{e.g.,} weather, system monitoring, e-commerce) similarly incorporate static descriptors or dynamic textual context, but remain largely domain-specific \citep{zheng2025lemmarcalargemultimodalmultidomain, chen2024terra, xubeyond, zhou2025motimedatasetsuitemultimodal}.

While TimeMMD \citep{timemmd} and Context-Is-Key \citep{williams2025context} provide a collection of multimodal datasets across domains, they are primarily used for benchmarking purposes, and not large-scale training, due to their small sizes.
To address these gaps, \textbf{we focus on leveraging LLMs to generate synthetic multimodal datasets} with realistic time-varying textual annotations aligned to real-world time series. This scales the volume of multimodal datasets, enabling large-scale training. 

\noindent\textbf{Multimodal forecasting approaches.}
Prior methods can be broadly grouped into \emph{zero-shot (training-free)} and \emph{dataset-specific, training-based} approaches.
Zero-shot methods primarily use LLMs by encoding time series and other modalities as textual prompts, enabling LLMs to reason over temporal patterns and contextual information \citep{guo2024towards, kong2025time, wang2024news, xie2023wall, yu2023temporal}. Such approaches offer flexibility across domains, but \citet{park2025revisitingllmszeroshottimeseries} show that they are often sensitive to noise and typically underperform with respect to lightweight or domain-specific forecasting models \citep{zeng2022transformerseffectivetimeseries} on standard benchmarks.

\input{figures_latex/selling_quantitative}
Training-based approaches learn joint representations of time series and text by fine-tuning LLMs \citep{li2024urbangpt,liu2023large,jia2024gpt4mts,jin2024timellmtimeseriesforecasting}, aligning text and time series embeddings \citep{han2026unica,li2026language,liu2024unitime, hong2025unifyanchorcontextawaretransformer}, or employing fusion architectures \citep{liu2025timecmallmempoweredmultivariatetime, liu2023large, chattopadhyay2024context,lee2025timecap,sawhney2020deep}. Several works explore vision-based \citep{daswani2024plotsunlocktimeseriesunderstanding, li2023timeseriesimagesvision, zhou2025llmsunderstandtimeseries} or vision--language-based \citep{zhong2025time} representations by encoding time series as images or plots, or closed-loop reasoning frameworks that iteratively align text, predictions, and explanations \citep{jiang2025explainable}.
These methods are either domain-specific or require per-dataset training.

Overall, previous works primarily (i) focus on LLMs as the key forecasters, (ii) rely on dataset-specific multimodal training, and (iii) do not exploit SOTA TSFMs \citep{chronos2,timesfm,grinsztajn2026tabpfn25advancingstateart}. In contrast, \textbf{we develop a zero-shot, domain-agnostic setup} that combines the strengths of LLMs and TSFMs to generate context-aware forecasts. We refer the readers to Appendix \ref{sec:extended_related_works} for a detailed discussion.

%% file: figures_latex/selling_quantitative.tex
\begin{wrapfigure}{r}{0.5\textwidth}
\vspace{-4mm}
\centering
\includegraphics[width=\linewidth]{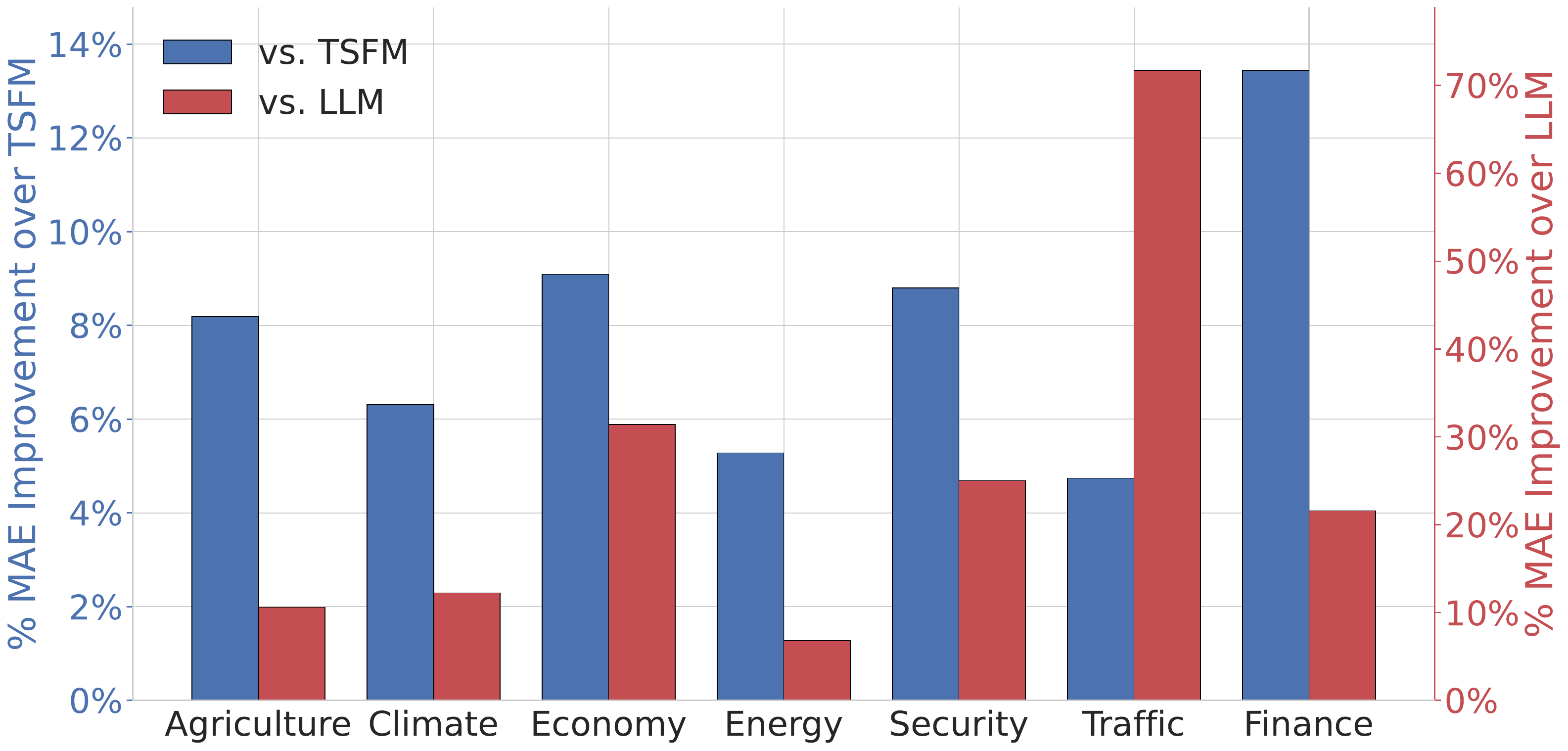}
\caption{\textbf{CoCo outperforms SOTA TSFMs (Chronos2) and closed-source LLMs (GPT-4o) across datasets from diverse domains.} CoCo effectively utilizes LLMs to distill textual annotations into usable forecast covariates that guide powerful TSFMs to generate accurate forecasts.}
\label{fig:selling_quantitative}
\vspace{-5mm}
\end{wrapfigure}

%% file: sections/3_problem_formulation.tex
\section{Problem Formulation}
\label{sec:prob_formulation}

Unimodal time series forecasting models take historical numerical series $X^H \in \mathcal{X}^H = \mathbb{R}^{l}$, where $l$ is the length of the context window, as 
input and generate the forecast $X^F \in \mathcal{X}^F = \mathbb{R}^{h}$, where $h$ is the horizon or forecast length. Typically, 
the sample at timestamp $t$ is denoted as $(X^H_t, X^F_t)$, 
where $X^H_t = [{x}_{t-l+1}, {x}_{t-l+2}, \ldots, {x}_t] \in \mathcal{X}^H$ and 
$X^F_t = [{x}_{t+1}, {x}_{t+2}, \ldots, {x}_{t+h}] \in \mathcal{X}^F$. Thus, the unimodal forecasting model 
parameterized by $\theta$ is given by $f^{\mathrm{uni}}_\theta : \mathcal{X}^H \rightarrow \mathcal{X}^F$.

In the multimodal setting, the textual annotations, denoted by 
${S} \in \mathcal{S} =\mathbb{N}^{l \times N^{\mathrm{token}}_{\mathrm{max}}}$, are available for each timestamp in the context for the most general setting, with up to $N^{\mathrm{token}}_{\mathrm{max}}$ tokens in each annotation. 
As with the unimodal case, the sample at timestamp $t$ in a dataset is denoted as $(X^H_t, S_t, X^F_t)$, 
where $S_t = [{s}_{t-l+1}, {s}_{t-l+2}, \ldots, {s}_t] \in \mathcal{S} $. Note that at any given timestamp 
$t$, $s_t$ can contain a summary of the past numerical values and the predictive cues to accurately forecast 
the future values. To this end, the multimodal forecasting model, parameterized by $\theta$, is denoted as 
$f^{\mathrm{multi}}_\theta : \mathcal{X}^H \times \mathcal{S} \rightarrow \mathcal{X}^F$.

\begin{center}
\begin{tcolorbox}[width=\linewidth, colback=brownfill, 
colframe=brownborder,
boxrule=1pt,
arc=3pt]
\textbf{Problem Statement:} We aim to develop a multimodal forecasting 
model $f^{\mathrm{multi}}_\theta$ such that the inclusion of textual context ${S}$ considerably improves 
the forecasting accuracy compared to the unimodal approach $f^{\mathrm{uni}}_\theta$ during zero-shot inference. 
More formally, given a dataset $D = \{(X^H_t, S_t, X^F_t)\}_{t=1}^{t=N}$, where $N$ is the 
total number of samples, and a metric $\mathcal{M}(\hat{{X}}^F, {X}^F)$ that evaluates the error in the 
forecast $\hat{{X}}^F$ relative to the ground truth $X^F$, we aim to develop $f^{\mathrm{multi}}_\theta$ such 
that $\mathcal{M}(f^{\mathrm{multi}}_\theta({X^H}, {S}), {X^F}) \leq \mathcal{M}(f^{\mathrm{uni}}_\theta({X^H}), {X^F}).$  
\end{tcolorbox}
\end{center}

%% file: sections/4_synthetic_data_generation.tex
\section{Synthetic Multimodal Dataset Generation}
\label{sec:dataset_gen}

We propose a scalable approach to address the scarcity of multi-domain multimodal datasets by generating textual annotations for real-world time series using LLMs, motivated by the following \textbf{key intuition}: given numerical trajectories and domain descriptions, LLMs can produce semantically meaningful explanations for observed temporal trends (see Figure~\ref{fig:synthetic_annotation_example}).

\input{figures_latex/toy_example}

For each dataset, we randomly sample windows of fixed length and provide the numerical values, timestamps, and dataset description to an LLM (\texttt{Qwen3-30B-A3B-Instruct-2507}~\citep{qwen3}, referred to as Qwen3-30B). The model is prompted to generate context-rich annotations for each timestamp $t$ describing plausible events or exogenous indicators explaining the observations up to $t$. Additionally, the LLM produces speculative descriptions of the future trajectory without generating explicit numerical forecasts. We added key indications to the prompt to ensure sufficient diversity in the generated annotations, and refer the readers to Appendix \ref{sec:synthetic_annotation_gen} for the actual prompt input to the LLM. 

To assess the realism of the generated annotations, we compare the cosine similarity between the normalized BERT \citep{bert} embeddings of the true future and the predicted forecast description, for real and synthetic cases. The significant overlap in resulting distributions (Figure~\ref{fig:assumption1_part_proof}) indicates synthetic annotations capture the forecast with levels of uncertainty similar to those observed in real context.

To this end, we make the following observation: \textbf{sim-to-real transfer}, \emph{i.e.,} training on synthetic and testing on real-world text annotations, is feasible because both annotation types encode similar domain-agnostic descriptions of forecast trends (e.g., \texttt{argmax} in the second forecast timestamp) in domain-specific terms (e.g., due to tariffs, crude oil prices are expected to peak in the next two days). We validate this observation using a toy example (Figure~\ref{fig:sim_to_real_toy}). Consider the  pharmaceutical sales of a company following a Seasonal Autoregressive Integrated Moving Average (SARIMA) process with shocks, explained by analyst notes. For each shock, an LLM generates a plausible domain-specific annotation. We fit SARIMA models with extracted sentiment features (scalar, domain-agnostic value ranging from -1 to 1) from both real and synthetic annotations and observe similar MAE on real annotations during the test period, demonstrating that synthetic text preserves forecast-relevant semantics useful for accurate prediction. A detailed description is provided in Appendix \ref{app:toy_description}.

To formalize this observation, we introduce a domain-agnostic contextual variable $Z \in \mathcal{S}$ (e.g., spikes or gradual trends). Let the history, forecast, and the forecast's domain-agnostic descriptions be sampled from a joint data distribution $p(X^H, Z, X^F)$, where the future $X^F$ depends on history $X^H$ and domain-agnostic context $Z$. The real annotations are naturally aligned with and describe the underlying future trends $Z$ in domain-specific terms ($S^r \sim p_r(S^r \mid X^H, Z)$). The synthetic annotations are generated as $S^s \sim p_s(S^s \mid X^H, Z, X^F)$, reflecting the use of future information by the LLM. To this end, for a multimodal forecaster $f^{\mathrm{multi}}_\theta(X^H,S)=h^{\mathrm{multi}}_\theta(X^H,\phi(S))$, we define the real and synthetic population risks as
\begin{equation*}
\resizebox{0.95\linewidth}{!}{$
R_{\mathrm{multi}}^r(\theta)
=
\mathbb{E}_{p,p_r}
\left[
\mathcal{M}\!\left(h^{\mathrm{multi}}_\theta(X^H,\phi(S^r)),X^F\right)
\right], \
R_{\mathrm{multi}}^s(\theta)
=
\mathbb{E}_{p,p_s}
\left[
\mathcal{M}\!\left(h^{\mathrm{multi}}_\theta(X^H,\phi(S^s)),X^F\right)
\right].
$}
\end{equation*}
Here, $\phi:\mathcal{S}\rightarrow\mathbb{R}^d$ is a text encoder with embedding dimension $d$, and $\mathbb{E}$ is the expectation operator. 

\begin{theorembox}
\begin{assumption}[Approximate semantic representation]\label{assump:semantic}
There exists a semantic embedding $\psi(Z) \in \mathbb{R}^d$ of the latent context and representation error bounds $\varepsilon_r, \varepsilon_s \geq 0$ such that
\begin{equation}
    \|\phi(S^r) - \psi(Z)\| \leq \varepsilon_r, \qquad \|\phi(S^s) - \psi(Z)\| \leq \varepsilon_s.
\end{equation}
\end{assumption}

\begin{assumption}[Lipschitz dependence on text representation]\label{assump:lipschitz}
There exists a constant $L > 0$ such that for all $X^H$, $X^F$, and $h^{\mathrm{multi}}_\theta : \mathcal{X}^H \times \mathbb{R}^d \rightarrow \mathcal{X}^F$, and all $u, v \in \mathbb{R}^d$,
\begin{equation}
    |\mathcal{M}(h^{\mathrm{multi}}_\theta(X^H, u), X^F) - \mathcal{M}(h^{\mathrm{multi}}_\theta(X^H, v), X^F)| \leq L \|u - v\|.
\end{equation}
\end{assumption}
\end{theorembox}

Here, $|\cdot|$ represents the absolute value and $\| \cdot \|$ represents the $l_2$ norm functions. \textbf{Assumption~\ref{assump:semantic}} states that both real ($S^r$) and synthetic ($S^s$) annotations lie close to the domain-agnostic forecast description $Z$ in a latent space. Now, we provide empirical evidence to back this assumption. First, real-world text annotations ($S^r$) are typically speculations that encode the future ($Z$) and thus imply a small $\varepsilon_r$. To justify small $\varepsilon_s$, we refer to Figure \ref{fig:assumption1_part_proof}. The strong overlap in cosine similarity distributions between the embeddings of forecast predictions derived from real ($\phi(S^r)$) and synthetic annotations ($\phi(S^s)$) with the true future descriptions ($\psi(Z)$), \emph{i.e.,} $\phi(S^r)^T\psi(Z)$ and $\phi(S^s)^T\psi(Z)$, indicates that synthetic annotations recover the forecast nearly as well as real ones, suggesting $\varepsilon_s \approx \varepsilon_r$. Additionally, the generation process grounds synthetic text in numerical trajectories and domain descriptions. Here, $\phi(S)$ and $\psi(Z)$ correspond to normalized BERT embeddings of predicted and true domain-agnostic forecast descriptions, respectively.

\textbf{Assumption~\ref{assump:lipschitz}} ensures that small perturbations in the text embedding produce limited changes in forecasting errors. Tables~\ref{tab:rephrasing_effect}, \ref{tab:text_perturbation}, and \ref{tab:adversial_text} show that perturbing real-world textual annotations causes negligible variation in forecasting error when the perturbations largely preserve forecast-relevant semantics. This empirical result points towards local smoothness in the loss space with respect to small perturbations in text annotations for reasonably good forecasters. 



\begin{theorembox}
\begin{lemma}[Representation mismatch bound]\label{lemma:mismatch}
Under \textbf{Assumptions~\ref{assump:semantic}} and~\textbf{\ref{assump:lipschitz}},
\begin{equation}
    |R_{\mathrm{multi}}^r(\theta) - R_{\mathrm{multi}}^s(\theta)| \leq L(\varepsilon_r + \varepsilon_s).
\end{equation}
\end{lemma}

\begin{theorem}[Synthetic-to-real transfer]\label{thm:transfer}
Let $\hat{\theta}_s \in \arg\min_{\theta \in \Theta} R_{\mathrm{multi}}^s(\theta)$ and $\theta^{\star}_r \in \arg\min_{\theta \in \Theta} R_{\mathrm{multi}}^r(\theta)$. Under \textbf{Assumptions~\ref{assump:semantic}} and~\textbf{\ref{assump:lipschitz}},
\begin{equation}\label{eq:transfer_bound}
    R_{\mathrm{multi}}^r(\hat{\theta}_s) \leq R_{\mathrm{multi}}^r(\theta^{\star}_r) + 2L(\varepsilon_r + \varepsilon_s).
\end{equation}
\end{theorem}
\end{theorembox}

\textbf{Lemma~\ref{lemma:mismatch}} states that for a forecaster, if real and synthetic embeddings are close to the same latent semantic representation (small $\varepsilon_r$ and $\varepsilon_s$), their risks remain close. \textbf{Theorem~\ref{thm:transfer}} states that a model trained with synthetic text achieves near-optimal performance under real text, with a bounded gap of $2L(\varepsilon_r+\varepsilon_s)$ relative to the best achievable real-text risk. In practice, this gap depends on (i) how effectively the encoder $\phi$ aligns real and synthetic text in a shared semantic space (whether $\varepsilon_r \simeq \varepsilon_s$), and (ii) how well synthetic annotations preserve the forecast-relevant semantics of $Z$ (whether $\|\phi(S^s) - \psi(Z)\|$ is small). Our results in Section~\ref{sec:experiments} demonstrate that models trained on synthetic text generalize effectively to real-world multimodal datasets. Proofs of \textbf{Lemma~\ref{lemma:mismatch}} and \textbf{Theorem~\ref{thm:transfer}} and a detailed discussion on the assumptions are provided in Appendix~\ref{appendix:theory_proofs}.

Our formulation is closely related to domain adaptation and invariant representations \citep{arjovsky2019invariant}, where domain generalization is achieved by aligning representations or assuming a shared latent variable. Similarly, sim-to-real transfer \citep{tobin2017domain} relies on learning representations robust to shifts between synthetic and real data. We adapt these concepts to multimodal forecasting and realistic synthetic annotation generation for domains where there is a lack of real textual annotations, with valid assumptions. Note that in our problem formulation and theoretical justification, the risk or the metric $\mathcal{M}(\hat{{X}}^F, {X}^F)$ can be the mean squared error (MSE) or the mean absolute error (MAE) between the ground truth and the forecast. Without any loss of generality, the theory and algorithms developed in this work can be extended to the multivariate case.

%% file: figures_latex/toy_example.tex
\begin{figure}[t]
    \centering
    \includegraphics[width=\textwidth]{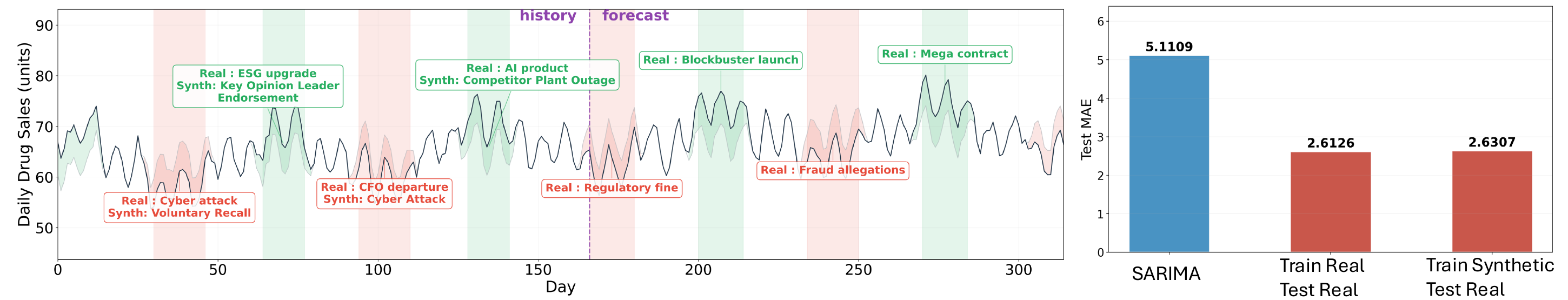}
    \caption{
    \textbf{Training on synthetic annotations yields real-world multimodal gains.} Using pharmaceutical sales data annotated with both real news and synthetic text, a SARIMA model with sentiment features achieves similar MAE when trained on synthetic versus real annotations, as both encode the same domain-agnostic forecast trends, thereby enabling \textbf{sim-to-real} transfer. 
    }
    \label{fig:sim_to_real_toy}
    \vspace{-0.2in}
\end{figure}

%% file: sections/5_model_arch.tex
\section{\textbf{Co}ntext as \textbf{Co}variates}\label{sec:model}
\input{figures_latex/model_arch}

We introduce CoCo (\textbf{Co}ntext as \textbf{Co}variates), a multimodal forecasting framework that combines LLMs' ability to interpret text with TSFMs' numerical precision (Figure~\ref{fig:model_arch}). CoCo uses the LLM to generate a text-aware forecast capturing broad contextual trends and a per-step confidence estimate, which are provided as covariates to guide a multivariate TSFM (M-TSFM), such as Chronos2, toward precise numerical forecasts. We first describe the covariate construction.

\textbf{LLM Forecast Covariate:} Our main intuition is to model the forecast covariate as a noisy version of the next timestep. The forecast covariate, $M_t \in \mathbb{R}^{(l+h)}$, is constructed from the numerical history $X^H_t = [x_{t-l+1}, \ldots, x_t]$ and the text-aware LLM forecast $\theta_{\mathrm{LLM}}(X^H_t, S_t) = [m_{t+1}, \ldots, m_{t+h}]$ as:
\begin{equation}
    M_t = \bigl[x_{t-l+2} + \epsilon_2,\; x_{t-l+3} + \epsilon_3,\; \ldots,\; x_t + \epsilon_l,\; m_{t+1},\; m_{t+2},\; \ldots,\; m_{t+h},\; m_{t+h} + \epsilon_{\text{end}}\bigr],
    \label{eq:forecast_covariate}
\end{equation}
where $\epsilon_i, \epsilon_{\text{end}} \sim \mathcal{N}(0, \sigma_\epsilon^2)$ are independent and identically distributed (i.i.d.) noise terms and we assign a very small value to $\sigma_\epsilon^2$. In the history portion, the covariate at position $i$ is a noisy version of the next observed value $x_i$. The future portion carries the LLM's median forecast directly. This construction allows M-TSFMs to learn through in-context learning that $M_t$ provides a noisy version of the true forecast, enabling text-induced trends in $[m_{t+1}, \ldots, m_{t+h}]$ to be included in the final forecast. In other words, this setup allows M-TSFMs to interpret $M_t$ as a noisy next-step predictor.

\textbf{LLM Confidence Covariate:} We model the confidence covariate to provide the M-TSFM with a per-step signal of how much to trust the LLM forecast. To measure confidence, we generate $N_\mathrm{rollouts}$ independent rollouts for the same multimodal input sample $(X^H_t, S_t)$, where the $i$-th rollout is denoted as $[m^{(i)}_{t+1}, \ldots, m^{(i)}_{t+h}]$. The confidence at future timestep $t+j$ ($1 \leq j \leq h$) is given by $c_{t+j} = \frac{\sigma^2\!\left(\{m^{(i)}_{t+j}\}_{i=1}^{N_\mathrm{rollouts}}\right)}{\sigma^2(X^H_t)}$,
where $\sigma$ denotes the standard deviation operator and dividing by $\sigma^2(X^H_t)$ makes $c_{t+j}$ scale invariant. To this end, we define the confidence covariate as
\begin{equation}
    C_t = [\epsilon_2,\; \epsilon_3,\; \ldots,\; \epsilon_l,\; c_{t+1},\; c_{t+2},\; \ldots,\; c_{t+h}, \epsilon_{\text{end}}].
    \label{eq:confidence_covariate}
\end{equation}
Small $c_{t+j}$ indicates higher LLM confidence at step $t+j$. The history portion of the confidence covariate is filled with $\epsilon_i \sim \mathcal{N}(0, \sigma_\epsilon^2)$, where we assign a very small value to $\sigma_\epsilon^2$ to teach the M-TSFM that the history portion of $M_t$ resembles an LLM forecast with a very high confidence score. 

Conditioned on $M_t$ and $C_t$, an M-TSFM ($\theta_\mathrm{M-TSFM}$) produces the forecast $\hat{X}^F_t = \theta_\mathrm{M-TSFM}(X^H_t \oplus \emptyset  \| M_t \| C_t)$. Here, $X^H_t \oplus \emptyset$ indicates padding the history $X^H_t$ for $h$ forecast timestamps and $\|$ denotes column-wise concatenation, such that $X^H_t \oplus \emptyset \| M_t \| C_t \in \mathbb{R}^{(l+h) \times 3}$ stacks the padded history and covariate channels along the feature dimension. For ease of notation, we drop $\emptyset$ and use $\hat{X}^F_t = \theta_\mathrm{M-TSFM}(X^H_t \| M_t \| C_t)$ in the rest of the manuscript.

We fine-tune both the LLM ($\theta_{\mathrm{LLM}}$) and the M-TSFM ($\theta_{\mathrm{M-TSFM}}$) in sequence instead of end-to-end joint fine-tuning. The LLM is fine-tuned first against a pretrained M-TSFM, and the M-TSFM is subsequently fine-tuned using covariates pre-generated from the fine-tuned LLM. 

\textbf{LLM Fine-Tuning:} We fine-tune the LLM with GRPO~\citep{shao2024deepseekmathpushinglimitsmathematical} to produce forecast covariates with text-induced trends by optimizing it against a pretrained M-TSFM, using forecasting improvement as the reward. Although the M-TSFM is held fixed, it can still respond to the covariate since $M_t$ is constructed as a noisy next-step continuation aligned with its pretraining inductive biases (Table~\ref{tab:additional_llms_tsfms}). This allows the model to partially utilize $M_t$ via in-context learning, making the relative forecasting error a meaningful reward signal. For a sample $(X^H_t, S_t, X^F_t)$, we generate $N_\mathrm{rollouts}$ forecasts $\{\theta_{\mathrm{LLM}}^{(g)}(X^H_t, S_t)\}_{g=1}^{N\mathrm{rollouts}}$ and construct covariates $M_t^{(g)}$. The reward and GRPO advantage are:
\begin{equation}
\begin{aligned}
r^{(g)}_t &= \frac{1}{1 + \dfrac{\|\theta_{\mathrm{M-TSFM}}(X^H_t \| M_t^{(g)}) - X^F_t\|_1}{\|\theta_{\mathrm{M-TSFM}}(X^H_t) - X^F_t\|_1 + \delta}},
\quad
A^{(g)}_t &= \frac{r^{(g)}_t - \frac{1}{N_\mathrm{rollouts}}\sum_{i=1}^{N_\mathrm{rollouts}}r^{(i)}_t}{\sigma(\left\{r_t^{(i)}\right\}_{i=1}^{N_\mathrm{rollouts}}) + \delta},
\label{eq:grpo_reward}
\end{aligned}
\end{equation}
where a very small $\delta > 0$ prevents division by zero ($\delta = 10^{-8}$). The reward lies in $(0,1)$, approaching $1$ when the covariate improves forecasting and $0.5$ when it provides no benefit. The confidence covariate $C_t$ is excluded from reward computation due to the cost of additional rollouts. We denote $\theta_\mathrm{M-TSFM}(X^H_t)$, $\theta_\mathrm{M-TSFM}(X^H_t \| M_t)$, and $\theta_\mathrm{M-TSFM}(X^H_t \| M_t \| C_t)$ as the unimodal, forecast-covariate, and full multimodal predictions, respectively.

\textbf{M-TSFM Fine-Tuning:} Following LLM fine-tuning, we pre-generate the full covariate set $\{M_t, C_t\}_{t=1}^N$ for all $N$ training samples using the fine-tuned LLM. The M-TSFM is then fine-tuned by minimizing the MSE between its covariate-conditioned forecast and the ground truth: $\frac{1}{N} \sum_{t=1}^{N} \left\| \theta_{\mathrm{M-TSFM}}(X^H_t \| M_t \| C_t) - X^F_t \right\|^2.$ We note that fine-tuning is essential to teach the M-TSFM to interpret $M_t$ as a soft next-step predictor and to modulate the reliance on $M_t$ according to the per-step confidence signal $C_t$, producing forecasts that are context-aware and numerically precise.

This two-stage approach avoids the instability of optimizing against a moving target and keeps training tractable. The architecture choices and the hyperparameters are described in detail in Appendix \ref{app:training_description}.

%% file: figures_latex/model_arch.tex
\begin{figure}[t]
    \centering
    \includegraphics[width=\textwidth]{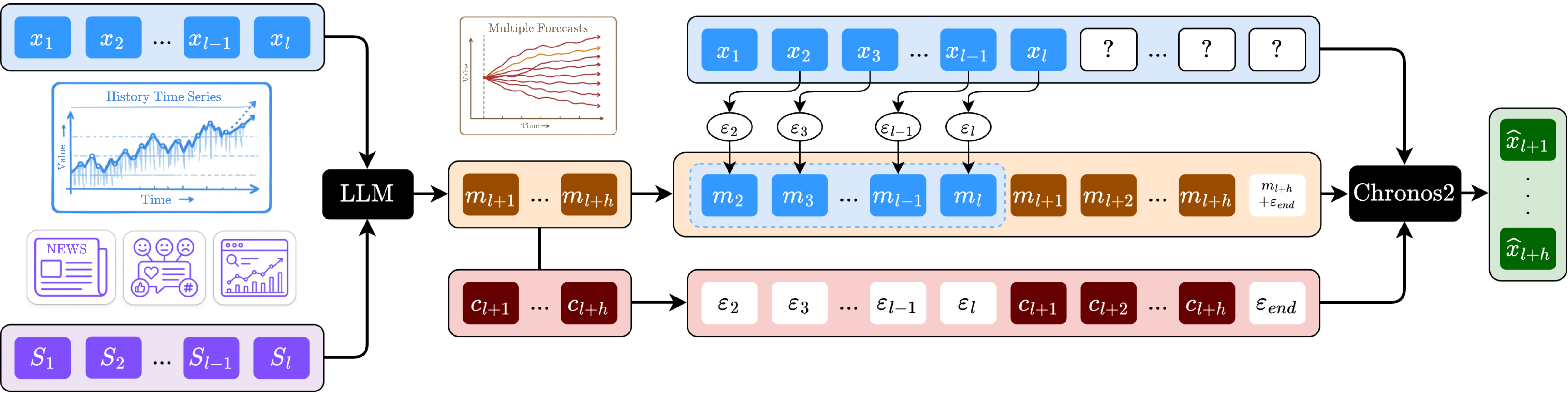}
    \caption{
    \textbf{Context as Covariates (CoCo) Description.} The LLM 
    $\theta_{\mathrm{LLM}}$ takes the numerical history $X^H_t$ (blue) and textual context 
    $S_t$ (violet) as input and produces a forecast covariate $M_t$ 
    (Equation~\ref{eq:forecast_covariate}, brown) and a rollout-based confidence 
    covariate $C_t$ (Equation~\ref{eq:confidence_covariate}, maroon), which are 
    concatenated with $X^H_t$ and passed to Chronos2 for multivariate forecasting. Our \textbf{key technical intuition} is to utilize the confidence covariate (obtained from the per-step variance in the LLM's predictions) to teach Chronos2 when to trust the LLM and when not to trust it (low variance $\rightarrow$ trust the LLM, as text induces a more decisive forecast).}
    \label{fig:model_arch}
\end{figure}


%% file: sections/6_experiments.tex
\section{Experiments}
\label{sec:experiments}
\input{figures_latex/rank_mae_bar_plot}
We use \texttt{Qwen3-4B-Instruct-2507}~\citep{qwen3} (Qwen3-4B) as the LLM and Chronos2 as the M-TSFM in CoCo, with both models fine-tuned as described in Section~\ref{sec:model}.

\textbf{Training Datasets and Setup:} We generated synthetic annotations (Section~\ref{sec:dataset_gen}) for 36,826 time series datasets ($\sim$12.4M rows) spanning retail, macroeconomics, environment, and finance, and curated $\sim$150,000 rows of real annotations from the finance \citep{fnspid} domain. Details of the datasets, annotation pipeline, and training (Section \ref{sec:model}) are provided in Appendices~\ref{sec:training_datasets} and~\ref{app:training_description}.

\textbf{Evaluation Datasets:} We evaluate CoCo on a wide range of 30 datasets (over 16,719 samples) spanning macroeconomics, finance, retail, climate, security, and traffic domains. These datasets are partly curated from existing benchmarks, such as TimeMMD \citep{timemmd} (8 datasets) and FNSPID \citep{fnspid} (11 datasets). Additionally, we curated text annotations for 11 real-world time series datasets with time-stamped news articles extracted using the Media Cloud API \citep{mediacloud}. A   description of our evaluation datasets is provided in Appendix \ref{sec:eval_datasets}. Our experiments used 64, 128, and 256 as context lengths and 16 as the forecast length. The TimeMMD dataset is also evaluated under the original paper settings (shown as TimeMMD*).

\input{tables/iclr/mae_setting_zero_shot}

\input{tables/iclr/mae_setting_finetuning}

\textbf{Baselines.} We compare CoCo against 27 methods, categorized by modality and adaptation: eight zero-shot unimodal, seven zero-shot multimodal, six unimodal fine-tuning, and six multimodal fine-tuning approaches. For a controlled comparison, every fine-tuned baseline uses the same synthetic text--time-series pairs as CoCo. Appendix~\ref{app:baselines} provides the complete list and implementation details.

\textbf{Metrics.} We report the per-dataset and aggregated (weighted equally to avoid bias from differing sample sizes) mean MAE. To aggregate relative forecasting performance across different settings, we report two complementary metrics: \emph{Average Rank}, for which the models are ranked (best model ranked 1) in each setting, and the average rank across all the settings is reported; and \emph{ELO Score} (detailed in Appendix \ref{app:evaluation_metrics}). For statistical significance, we report the Wilcoxon p-value based on MAE computed across settings. The quantitative comparisons are summarized in Tables \ref{tab:mae_setting_summary_zs} and \ref{tab:mae_setting_summary_ft} with dataset-level metrics for each baseline category shown in Tables~\ref{tab:class_mae_pooled_95_zero_shot_unimodal}-\ref{tab:class_mae_pooled_95_multimodal_finetuning}. For the Context-Is-Key benchmark, we report the Region of Interest Continuous Ranked Probability Score (RCRPS). 

\textbf{CoCo outperforms SOTA TSFMs and LLM-based forecasters consistently.} We refer the readers to Tables \ref{tab:mae_setting_summary_zs} and \ref{tab:mae_setting_summary_ft}, and Figure \ref{fig:rank_mae_plot}. The comparison against the SOTA TSFMs reveals the multimodal gains that can be achieved from textual annotations. Improvements over LLM-based forecasters showcase the ability of CoCo to effectively combine textual context and time series representations. Similarly, significant improvements over multimodal fine-tuning approaches elucidate the right approach to utilize textual annotations to improve SOTA TSFMs.

\textbf{Synthetic annotations enable generalization to domains without real multimodal data.} Table~\ref{tab:text_in_finetuning} shows that fine-tuning on real FNSPID data achieves strong in-domain performance (MAE 0.891) but degrades on out-of-distribution TimeMMD (MAE 1.528). In contrast, fine-tuning on synthetic annotations (Section \ref{sec:dataset_gen}) yields comparable FNSPID performance (MAE 0.903) and improved TimeMMD performance (MAE 1.365). Finally, fine-tuning on both real and synthetic annotations achieves the best overall results (TimeMMD MAE 1.354) without degrading FNSPID (MAE 0.894). These results show that fine-tuning on our synthetic annotations can enable generalization to domains lacking large-scale multimodal datasets, thereby showcasing \textbf{sim-to-real} transfer in multimodal forecasting. Notably, neither supervised fine-tuning nor reinforcement learning (RL) fine-tuning without text achieves comparable gains, confirming that both the GRPO reward signal and the textual annotations are essential for fine-tuning.

\input{tables/effect_of_cov_and_ft}

\textbf{CoCo's gains are strongly dependent on how effectively the LLM utilizes the text.} We use the relative MAE improvement in the LLM forecast when the textual context $S_t$ is provided versus withheld, as a measure of text usefulness. Positive values indicate that the text helped the LLM produce a better forecast, while negative values indicate misleading or incorrect text. Figure \ref{fig:text_use_ablation} (left) shows that text usefulness is a strong indicator of CoCo's gains. Figure \ref{fig:text_use_ablation} (right) shows that when text is highly useful, performance peaks at high LLM confidence and degrades as confidence decreases. Conversely, when text is not useful, high LLM confidence is actively harmful, since Chronos2 is misled by confidently wrong covariates. This demonstrates that $C_t$ acts as an effective gating mechanism, allowing Chronos2 to selectively incorporate $M_t$ based on its per-step reliability.

\textbf{Both fine-tuning stages and both covariates are crucial for forecasting performance.} Tables \ref{tab:finetuning_ablation_sumary}, \ref{tab:covariate_finetuning_usefulness} and \ref{tab:finetuned_components_ablation} show that fine-tuning both LLM and Chronos2, and providing both the covariates, yields the largest improvement (3.36\% over the Chronos2 baseline), with each component contributing incrementally. Adding the confidence covariate on top of the forecast covariate consistently improves performance across all fine-tuning configurations, except for the zero-shot setting, where neither the LLM has learned to provide good covariates nor Chronos2 has learned to interpret them.

\input{figures_latex/generalization_heatmap}

\textbf{CoCo generalizes to multiple LLM $\times$ M-TSFM pairs.} Tables~\ref{tab:additional_llms_tsfms} and~\ref{tab:coco_with_toto}, and Figure~\ref{fig:framework_generalization} demonstrate that CoCo is not specific to Qwen3-4B or Chronos2. Across five LLM configurations and three M-TSFM architectures, the forecast covariate $M_t$ improves every evaluated LLM $\times$ M-TSFM combination, while the confidence covariate $C_t$ provides further gains with zero-shot Toto and fine-tuned Chronos2. Although Chronos2 is fine-tuned using rollouts pre-generated only by the fine-tuned Qwen3-4B, it generalizes to covariates produced by all evaluated zero-shot LLMs. The fine-tuned Qwen3-4B also outperforms the larger zero-shot LLMs as a standalone forecaster (Figure~\ref{fig:standalone_llm_forecast}).

\textbf{CoCo improves text-time series alignment in pretrained LLMs.} We evaluate this using the Context-Is-Key benchmark (CIK, Tables \ref{tab:cik_summary} and \ref{tab:cik}), where accurate forecasting requires strong cross-modal alignment between text and time series due to text-induced trends and spikes. The RCRPS metric further penalizes incorrect trends and shifted spikes. The CoCo fine-tuned LLM consistently outperforms its pretrained counterpart, showing that GRPO fine-tuning improves the LLM's ability to combine numerical forecasting with text-based reasoning. Notably, CoCo fine-tuned Qwen3-4B matches the performance of GPT-4o. We refer readers to Appendix~\ref{app:cik} for detailed results.

\input{figures_latex/text_usefulness}

\textbf{Limitations.} First, we note that CoCo inference is slower than unimodal TSFMs due to the additional LLM inference required for extracting predictive signals from text (Table~\ref{tab:latency}), a capability fundamentally absent in TSFMs. Further, CoCo assumes that textual annotations are generally predictive of future trends. When the text is misleading, the LLM can generate confidently incorrect covariates that Chronos2 cannot fully correct, thereby degrading performance (Table~\ref{tab:adversial_text}, Figure~\ref{fig:text_use_ablation} (left)). Finally, CoCo is designed for settings where numerical history and textual context carry complementary predictive signals. In settings where text is the sole driver of forecast outcomes, the TSFM backbone contributes less, and we observe that CoCo's gains over just the LLM are limited (Appendix~\ref{app:cik}). 

\input{tables/cik_summary}

\textbf{Additional Experiments.}
Table~\ref{tab:post_june_2024_zero_shot_multimodal} shows that CoCo continues to outperform zero-shot multimodal baselines on samples strictly after June 2024, suggesting robustness beyond potential pretraining overlap. To reduce sensitivity to dataset scale and outliers in aggregate MAE, Table~\ref{tab:mase_setting_summary} reports the geometric mean of MASE across datasets, following GIFT-Eval~\citep{aksu2024gift}, and yields conclusions consistent with MAE. Table~\ref{tab:average_rank_all_models} further shows that CoCo's relative performance remains consistent across raw and normalized evaluation metrics. Table~\ref{tab:covariate_finetuning_usefulness} demonstrates that the strongest multimodal performance is obtained by fine-tuning both the LLM and M-TSFM and using both $M_t$ and $C_t$. Finally, across 30 Media Cloud settings, CoCo outperforms GPT-5 in $70\%$ ($21/30$) of cases and achieves up to an $11\%$ MAE improvement on steel-price forecasting (Table~\ref{tab:gpt5_zero_shot_multimodal}). Table \ref{tab:stochasticity_rollouts} shows CoCo's robustness to the stochasticity induced by the LLM rollouts. Table~\ref{tab:covariate_baselines_combined} shows that increasing the number of rollouts beyond $N_{\mathrm{rollouts}}=5$ provides minimal gains. We observe that generating textual annotations from historical numerical values alone produces textual annotations that do not reflect the real forecast trends. On the other hand, future-conditioned textual annotation generation reflects the true forecast trends better (Table~\ref{tab:annotation_probe_accuracy}).

\section{Conclusion and Future Work}
In this work, we first proposed a simple yet effective recipe for generating large-scale multimodal time series datasets. Specifically, we focused on text as the alternative modality. Secondly, we proposed \textbf{Co}ntext as \textbf{Co}variates, a novel multimodal architecture for incorporating orthogonal textual information to refine the forecasts from SOTA TSFMs. CoCo achieves up to 12\% MAE improvement over state-of-the-art TSFMs and LLM forecasters. As future work, we aim to add a text reliability assessment step to CoCo to improve the unimodal forecast only when text is useful and not misleading. This can be empirically assessed using the relationship between the text and time series in the context or history. Additionally, the current setup involves fine-tuning the LLM and M-TSFM sequentially. To this end, we aim to explore the possibility of jointly fine-tuning both models.   

\begin{ack}
This work was supported in part by the National Science Foundation under Grant No. 2148186. Any opinions, findings, and conclusions or recommendations expressed in this material are those of the authors and do not necessarily reflect the views of the National Science Foundation.
\end{ack}



%% file: figures_latex/rank_mae_bar_plot.tex
\begin{figure}[t]
    \centering
    \includegraphics[width=0.9\textwidth]{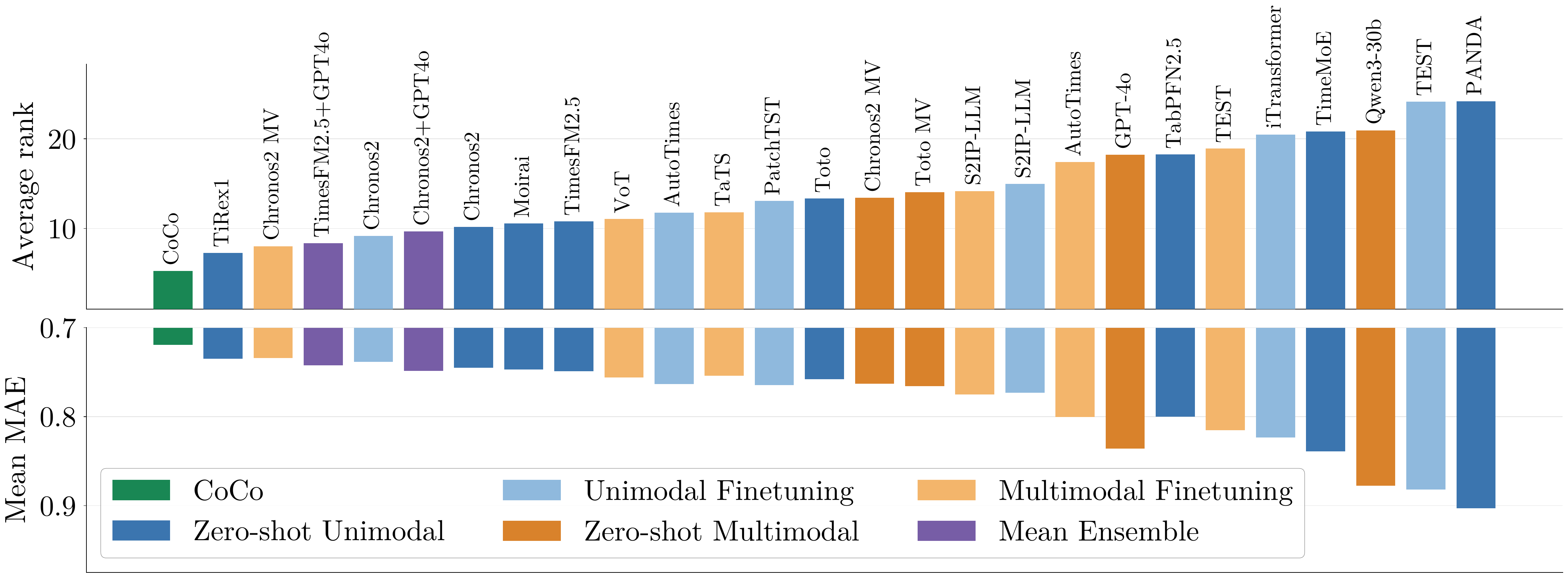}
    \caption{\textbf{CoCo improves over unimodal and multimodal fine-tuning approaches, including SOTA TSFMs on real-world multimodal datasets.} CoCo (ours) achieves the best average rank and mean MAE across 30 datasets (95 settings) spanning diverse domains and context lengths against 26 (TimeOmni omitted for scale) approaches.
    These gains demonstrate the effectiveness of GRPO-based fine-tuning along with our novel two-step fine-tuning. 
    }
    \label{fig:rank_mae_plot}
\end{figure}


%% file: tables/iclr/mae_setting_zero_shot.tex
\begin{table*}[t]
\centering
\footnotesize
\renewcommand{\arraystretch}{1.08}
\setlength{\tabcolsep}{3pt}
\resizebox{\textwidth}{!}{%
\begin{tabular}{@{}l ccc ccc ccc c@{\hspace{10pt}}ccccc@{}}
\toprule
 & \multicolumn{10}{c}{Mean MAE ($\downarrow$)} & \multicolumn{5}{c}{Aggregate Performance} \\
\cmidrule(lr){2-11} \cmidrule(l){12-16}
 & \multicolumn{3}{c}{64--16 Setting} & \multicolumn{3}{c}{128--16 Setting} & \multicolumn{3}{c}{256--16 Setting} & Native & Overall & Average & ELO & CoCo & Wilcoxon \\
\cmidrule(lr){2-4} \cmidrule(lr){5-7} \cmidrule(lr){8-10} \cmidrule(lr){11-11} \cmidrule(l){12-16}
Model & Media Cloud & FNSPID & TimeMMD & Media Cloud & FNSPID & TimeMMD & Media Cloud & FNSPID & TimeMMD & TimeMMD$^{*}$ & MAE $\downarrow$ & Rank $\downarrow$ & Rating $\uparrow$ & Wins & $p$ \\
\midrule
\rowcolor{gray!15} \textbf{CoCo (Ours)} & \textbf{0.809} & \underline{0.907} & \textbf{0.811} & \textbf{0.567} & \textbf{0.613} & \textbf{0.672} & \textbf{0.446} & \underline{0.453} & \textbf{0.678} & \textbf{1.329} & \textbf{0.719} & \textbf{5.24} & \textbf{1838} & -- & -- \\
\midrule
\multicolumn{16}{@{}l}{\textbf{Zero-shot Unimodal}} \\
TiRex1 & 0.835 & 0.915 & \underline{0.824} & 0.580 & \underline{0.616} & \underline{0.696} & \underline{0.450} & \textbf{0.450} & 0.738 & 1.352 & \underline{0.735} & \underline{7.26} & \underline{1750} & 60/95 & 5.0e-04 \\
Chronos2 & 0.847 & 0.942 & 0.832 & 0.590 & 0.631 & 0.701 & 0.461 & 0.462 & \underline{0.718} & 1.362 & 0.745 & 10.18 & 1643 & 81/95 & 2.0e-12 \\
Moirai & 0.843 & 0.923 & 0.843 & 0.576 & 0.647 & 0.708 & 0.461 & 0.468 & 0.726 & 1.379 & 0.747 & 10.57 & 1638 & 78/95 & 9.2e-13 \\
TimesFM2.5 & 0.861 & 0.941 & 0.824 & 0.597 & 0.643 & 0.699 & 0.468 & 0.469 & 0.732 & 1.342 & 0.749 & 10.78 & 1629 & 72/95 & 7.8e-09 \\
Toto & 0.865 & 0.955 & 0.836 & 0.589 & 0.660 & 0.710 & 0.473 & 0.484 & 0.737 & 1.358 & 0.758 & 13.36 & 1548 & 80/95 & 8.1e-13 \\
TabPFN2.5 & 0.908 & 0.968 & 0.864 & 0.647 & 0.662 & 0.724 & 0.632 & 0.538 & 0.770 & 1.390 & 0.800 & 18.27 & 1385 & 90/95 & 1.4e-16 \\
TimeMoE & 1.001 & 1.050 & 0.939 & 0.647 & 0.710 & 0.768 & 0.653 & 0.574 & 0.750 & 1.378 & 0.839 & 20.82 & 1292 & 88/95 & 3.3e-15 \\
PANDA & 1.072 & 1.133 & 1.060 & 0.770 & 0.790 & 0.856 & 0.565 & 0.547 & 0.812 & 1.499 & 0.903 & 24.21 & 1134 & 93/95 & 4.1e-17 \\
\midrule
\multicolumn{16}{@{}l}{\textbf{Zero-shot Multimodal}} \\
TimesFM2.5+GPT-4o & \underline{0.829} & \textbf{0.898} & 0.857 & \underline{0.575} & 0.625 & 0.720 & 0.456 & 0.456 & 0.742 & 1.385 & 0.742 & 8.35 & 1712 & 60/95 & 5.0e-03 \\
Chronos2+GPT-4o & 0.831 & 0.910 & 0.864 & 0.582 & 0.629 & 0.718 & 0.460 & 0.465 & 0.734 & 1.412 & 0.748 & 9.65 & 1664 & 65/95 & 6.3e-05 \\
Chronos2 MV & 0.892 & 0.947 & 0.870 & 0.607 & 0.639 & 0.715 & 0.516 & 0.475 & 0.721 & \underline{1.341} & 0.763 & 13.42 & 1542 & 74/95 & 1.8e-10 \\
Toto MV & 0.902 & 0.962 & 0.844 & 0.600 & 0.668 & 0.706 & 0.499 & 0.487 & 0.719 & 1.348 & 0.766 & 14.04 & 1518 & 75/95 & 6.1e-10 \\
GPT-4o & 0.913 & 0.966 & 1.035 & 0.644 & 0.680 & 0.853 & 0.513 & 0.502 & 0.845 & 1.586 & 0.836 & 18.23 & 1388 & 85/95 & 8.9e-15 \\
Qwen3-30B & 0.949 & 0.985 & 1.080 & 0.691 & 0.717 & 0.927 & 0.546 & 0.539 & 0.902 & 1.631 & 0.878 & 20.96 & 1293 & 94/95 & 2.8e-17 \\
TimeOmni & 1.414 & 1.126 & 1.515 & 1.099 & 0.736 & 8.543 & 0.760 & 0.564 & 1.490 & 1.748 & 1.756 & 25.67 & 1054 & 95/95 & 2.6e-17 \\
\bottomrule
\end{tabular}%
}
\caption{\textbf{CoCo outperforms SOTA zero-shot unimodal and multimodal forecasting approaches on real-world multimodal datasets.} CoCo (ours) achieves the best mean MAE across 30 datasets (95 settings) spanning diverse domains and context lengths. It outperforms SOTA TSFMs such as TiRex1 in over $63\%$ (60/95) of settings. Similarly, CoCo surpasses constructed SOTA zero-shot multimodal baselines (TimesFM2.5 + GPT-4o) in 60/95 settings with a significant improvement in the aggregate MAE and the average rank metrics, highlighting effective \textbf{sim-to-real} transfer.}
\label{tab:mae_setting_summary_zs}
\end{table*}

%% file: tables/iclr/mae_setting_finetuning.tex
\begin{table*}[t]
\centering
\footnotesize
\renewcommand{\arraystretch}{1.08}
\setlength{\tabcolsep}{3pt}
\resizebox{\textwidth}{!}{%
\begin{tabular}{@{}l ccc ccc ccc c@{\hspace{10pt}}ccccc@{}}
\toprule
 & \multicolumn{10}{c}{Mean MAE ($\downarrow$)} & \multicolumn{5}{c}{Aggregate Performance} \\
\cmidrule(lr){2-11} \cmidrule(l){12-16}
 & \multicolumn{3}{c}{64--16 Setting} & \multicolumn{3}{c}{128--16 Setting} & \multicolumn{3}{c}{256--16 Setting} & Native & Overall & Average & ELO & CoCo & Wilcoxon \\
\cmidrule(lr){2-4} \cmidrule(lr){5-7} \cmidrule(lr){8-10} \cmidrule(lr){11-11} \cmidrule(l){12-16}
Model & Media Cloud & FNSPID & TimeMMD & Media Cloud & FNSPID & TimeMMD & Media Cloud & FNSPID & TimeMMD & TimeMMD$^{*}$ & MAE $\downarrow$ & Rank $\downarrow$ & Rating $\uparrow$ & Wins & $p$ \\
\midrule
\rowcolor{gray!15} \textbf{CoCo (Ours)} & \textbf{0.809} & 0.907 & \underline{0.811} & \textbf{0.567} & \textbf{0.613} & \underline{0.672} & \textbf{0.446} & \textbf{0.453} & \underline{0.678} & \textbf{1.329} & \textbf{0.719} & \textbf{5.24} & \textbf{1838} & -- & -- \\
\midrule
\multicolumn{16}{@{}l}{\textbf{Unimodal Fine-tuning}} \\
Chronos2 & 0.851 & 0.924 & \textbf{0.810} & 0.583 & 0.637 & \textbf{0.670} & \underline{0.460} & 0.469 & \textbf{0.675} & 1.389 & 0.738 & 9.18 & 1677 & 64/95 & 1.2e-05 \\
AutoTimes & 0.851 & 0.916 & 0.901 & 0.585 & 0.633 & 0.773 & 0.515 & 0.473 & 0.784 & 1.337 & 0.763 & 11.75 & 1588 & 78/95 & 3.2e-10 \\
PatchTST & 0.858 & 0.927 & 0.903 & 0.603 & 0.638 & 0.762 & 0.469 & 0.492 & 0.772 & \underline{1.332} & 0.764 & 13.06 & 1559 & 89/95 & 7.2e-15 \\
S2IP-LLM & 0.893 & 0.939 & 0.892 & 0.619 & 0.635 & 0.736 & 0.492 & 0.500 & 0.768 & 1.363 & 0.773 & 14.98 & 1491 & 87/95 & 3.5e-15 \\
iTransformer & 0.894 & 0.962 & 0.920 & 0.694 & 0.756 & 0.812 & 0.541 & 0.579 & 0.808 & 1.354 & 0.823 & 20.49 & 1300 & 92/95 & 6.1e-17 \\
TEST & 0.962 & 1.040 & 1.026 & 0.679 & 0.731 & 0.860 & 0.588 & 0.609 & 0.880 & 1.590 & 0.882 & 24.14 & 1145 & 94/95 & 2.7e-17 \\
\midrule
\multicolumn{16}{@{}l}{\textbf{Multimodal Fine-tuning}} \\
Chronos2 MV \ \ \  \ \ \ \ \ \ \ \ \ \ \ \  & 0.845 & \underline{0.905} & 0.834 & \underline{0.573} & \underline{0.625} & 0.697 & 0.462 & \underline{0.461} & 0.702 & 1.334 & \underline{0.734} & \underline{8.02} & \underline{1714} & 68/95 & 1.8e-05 \\
VoT & \underline{0.836} & \textbf{0.888} & 0.885 & 0.594 & 0.628 & 0.753 & 0.530 & 0.488 & 0.745 & 1.339 & 0.756 & 11.06 & 1614 & 71/95 & 1.5e-08 \\
TaTS & 0.883 & 0.936 & 0.848 & 0.594 & 0.637 & 0.702 & 0.509 & 0.473 & 0.707 & 1.343 & 0.754 & 11.79 & 1592 & 70/95 & 1.3e-08 \\
S2IP-LLM & 0.858 & 0.910 & 0.896 & 0.610 & 0.647 & 0.758 & 0.528 & 0.546 & 0.771 & 1.338 & 0.775 & 14.18 & 1518 & 85/95 & 1.3e-13 \\
AutoTimes & 0.884 & 0.949 & 0.973 & 0.626 & 0.663 & 0.835 & 0.552 & 0.520 & 0.790 & 1.350 & 0.801 & 17.40 & 1414 & 83/95 & 2.3e-13 \\
TEST & 0.857 & 0.970 & 0.977 & 0.636 & 0.706 & 0.840 & 0.556 & 0.544 & 0.813 & 1.390 & 0.815 & 18.93 & 1359 & 91/95 & 2.7e-16 \\
\bottomrule
\end{tabular}%
}
\caption{\textbf{CoCo outperforms SOTA compute-matched and fine-tuned forecasting approaches on real-world multimodal datasets.} Similar to Table \ref{tab:mae_setting_summary_zs}, here, we show CoCo's improvements over fine-tuned and compute-matched baselines. Observe how, even after fine-tuning, both unimodal and multimodal baselines struggle to match CoCo's performance.}
\label{tab:mae_setting_summary_ft}
\end{table*}

%% file: tables/effect_of_cov_and_ft.tex
\begin{wraptable}{r}{0.5\linewidth}
\centering
\small
\begin{tabular}{lcc}
\toprule
\textbf{Fine-tuning} & $M_t$ & Both $M_t$ + $C_t$ \\
\midrule
Only $\theta_{\mathrm{LLM}}$ & 0.21 & 1.64 \\
Only $\theta_{\mathrm{M-TSFM}}$ & 2.45 & 2.55 \\
Both $\theta_{\mathrm{LLM}}$ and $\theta_{\mathrm{M-TSFM}}$ & 2.6 & \textbf{3.36} \\
\bottomrule
\end{tabular}
\caption{\textbf{Using both $M_t$ and $C_t$ consistently outperforms $M_t$ alone across all fine-tuning configurations}. We show the \% MAE improvement over SOTA unimodal models. Sequential fine-tuning of both LLM $\theta_{\mathrm{LLM}}$ and M-TSFM $\theta_{\mathrm{M-TSFM}}$ yields the best \% MAE improvement (Table \ref{tab:covariate_finetuning_usefulness}).}
\label{tab:finetuning_ablation_sumary}
\vspace{-4mm}
\end{wraptable}


%% file: figures_latex/generalization_heatmap.tex
\begin{wrapfigure}[21]{r}{0.5\textwidth}
    \centering
    \includegraphics[width=\linewidth]{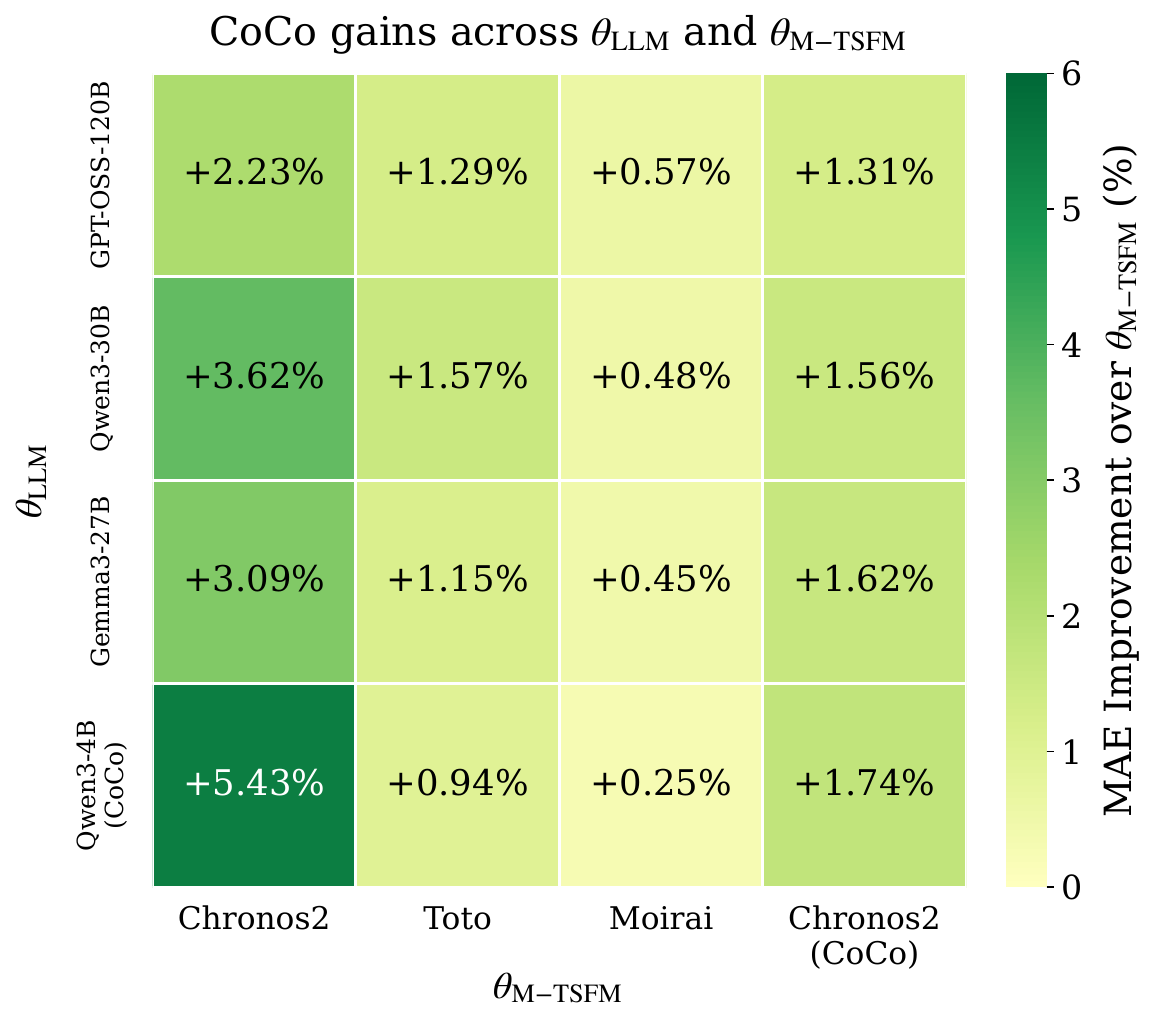}
    \caption{\textbf{CoCo generalizes across model choices.}
    MAE improvement over each numerical-only M-TSFM baseline on Media Cloud (64-16 setting).
    }
    \label{fig:framework_generalization}
\end{wrapfigure}

\noindent

%% file: figures_latex/text_usefulness.tex
\begin{figure}[t]
    \centering
    \includegraphics[width=0.95\textwidth]{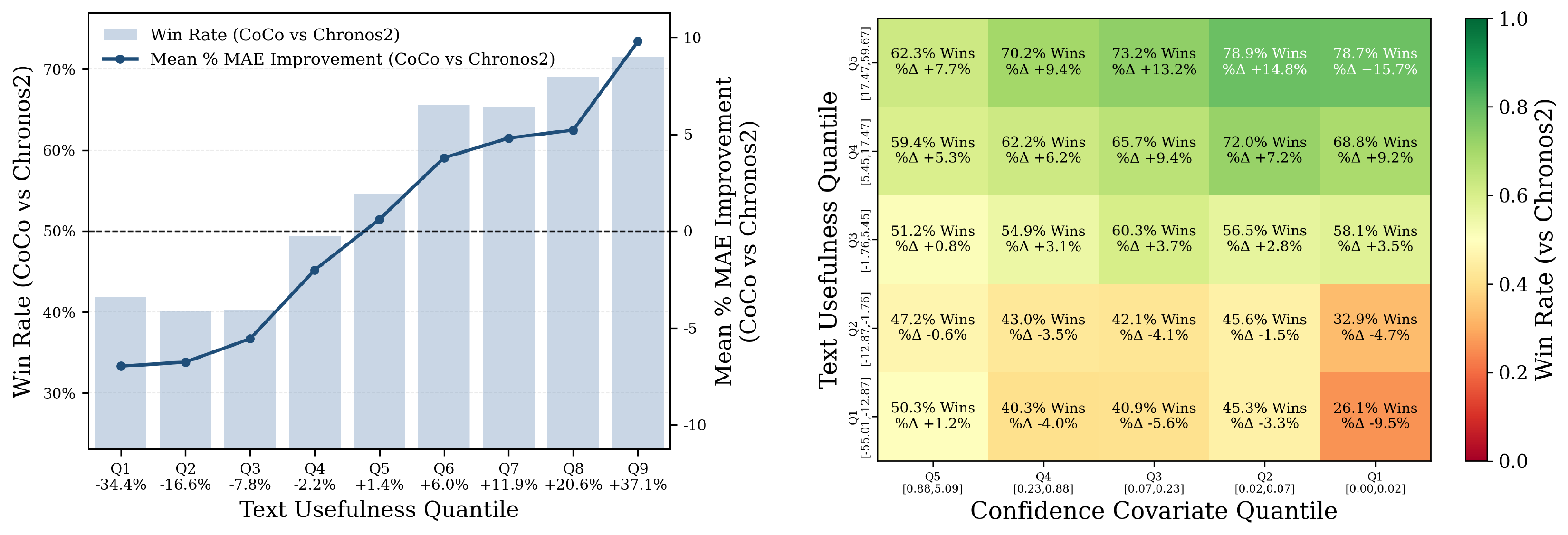}
    \caption{
    \textbf{(Left) CoCo consistently outperforms SOTA unimodal models (Chronos2) when text is useful (Q5--Q9) and underperforms when it is not (Q1--Q3).} Text usefulness is defined as the relative MAE improvement of the LLM forecast when the context $S_t$ is provided versus 
withheld. \textbf{(Right) Furthermore, for the set of windows where text is useful, CoCo significantly outperforms unimodal models when the rollout variance from the LLM (Confidence $C_t$) is very low.}}
    \label{fig:text_use_ablation}
\end{figure}

%% file: tables/cik_summary.tex
\begin{wraptable}[19]{r}{0.52\linewidth}
\centering
\small
\begin{tabular}{lcc}
\toprule
\textbf{Model} & \textbf{Mean RCRPS} $\downarrow$ \\
\midrule
Qwen3-4B
    & 0.358 \\
Qwen3-4B + Reasoning 
    & 0.189  \\
Qwen3-4B (CoCo) 
    & 0.301 \\
Qwen3-4B (CoCo) + Reasoning 
    & 0.161  \\
\midrule
Chronos2 (Unimodal) 
    & 0.265 \\
\midrule
CoCo  
    & 0.256 \\
\midrule
\textit{GPT-4o (reference)} 
    & \textit{0.191 } \\
\bottomrule
\end{tabular}
\caption{\textbf{LLMs fine-tuned with CoCo have improved text-time series alignment and outperform their pretrained counterparts.} Notably, Qwen3-4B fine-tuned with CoCo, with reasoning, matches the zero-shot results from GPT-4o. We refer to Table \ref{tab:cik} for the detailed results.}
\label{tab:cik_summary}
\vspace{-3mm}
\end{wraptable}

%% file: sections/appendix.tex
\appendix
\etocdepthtag.toc{mtappendix}

\input{appendix/overview}

\input{appendix/1_extended_related_works}

\input{appendix/9_proofs}

\input{appendix/11_synthetic_example}

\input{tables/rephrasing_effect}

\input{tables/rebuttal/text_perturbation_analysis}

\input{tables/rebuttal/representation_distance}

\input{appendix/8_context_is_key}

\section{Training}

\input{appendix/2_training_description}

\input{appendix/3_training_dataset}

\section{Evaluation}

\input{appendix/4_evaluation_datasets}

\input{appendix/rebuttal/2_baselines}

\input{appendix/5_evaluation_metrics}

\input{appendix/rebuttal/1_stochasticity}



\section{Extended Results}

\input{tables/iclr/mae_setting_summary}
\input{tables/iclr/dataset_mae_zero_shot_unimodal}
\input{tables/iclr/dataset_mae_zero_shot_multimodal}
\input{tables/iclr/dataset_mae_unimodal_finetuning}
\input{tables/iclr/dataset_mae_multimodal_finetuning}
\input{tables/iclr/post_june_2024_zero_shot_multimodal}
\input{tables/iclr/mase_setting_summary}
\input{tables/iclr/average_rank_all_models}
\input{tables/iclr/gpt5_zero_shot_multimodal}

\input{tables/adversial_text}

\input{tables/covariate_usefulness}
\input{tables/rollout_k_effect}
\input{tables/text_in_finetuning}

\input{tables/rebuttal/tsfm_llm_generalization}
\input{tables/coco_with_toto}
\input{tables/rebuttal/leakage_probe_analysis}
\input{tables/rebuttal/finetuned_component_ablation}
\input{figures_latex/standalone_llm_forecast}
\input{tables/latency}

\input{appendix/6_llm_prompts}

\input{appendix/12_broader_impacts}

%% file: appendix/overview.tex
\begin{Large}
 \textbf{Appendix Overview}
\end{Large}

\etocdepthtag.toc{mtappendix}
\etocsettagdepth{mtchapters}{none}   
\etocsettagdepth{mtappendix}{subsection} 

\etocsettocstyle{}{}   
\tableofcontents                      

\newpage

%% file: appendix/1_extended_related_works.tex
\section{Extended Related Works}
\label{sec:extended_related_works}
\subsection{Multimodal Datasets for Training and Evaluation}
\label{sec:extended_related_datasets}
Multiple prior works have proposed multimodal time series datasets with primary focus on textual annotations for forecasting. The nature and richness of textual annotations vary widely across domains. In particular, numerous works \cite{liu2024echo, sawhney2021fast, wang2024news, xu2018stock, wu2018hybrid, chen2023chatgpt} focus on stock price prediction using auxiliary information, such as news articles, Twitter sentiments, and Google search trends. This is primarily due to the abundance of rich and time-varying textual annotations that describe the trends in stock price values, such as the daily opening, closing, high, and low prices. For example, FNSPID \cite{fnspid} contains stock prices and aligned textual annotations for over 4,000 companies from 1999 to 2023. In contrast, textual annotations in the healthcare domain, \cite{wang2020mimic, smigiel2021ecg} are typically global and static, in the form of patient medical records for electrocardiogram (ECG) signals. Such datasets have been primarily used for tasks like classification and anomaly detection. Similarly, \citet{zheng2025lemmarcalargemultimodalmultidomain} have a large collection of performance metrics time series that are logged from complex systems for root-cause analysis. In the weather prediction domain, \citet{chen2024terra} provide static descriptions of the geography, while \citet{xubeyond} focus on time-varying textual annotations that aid in forecasting temperature, rain levels, and other weather metrics. More recently, \citet{zhou2025motimedatasetsuitemultimodal} proposed Mo-Time, a multimodal dataset with static annotations spanning e-commerce and social media user behavior domains. From these prior works, our key takeaway is that large-scale multimodal datasets are generally domain-specific and localized (annotations are available for a short time frame and are restricted in terms of geographic locations). Additionally, based on the domains, the annotations can be static or time-varying.

TimeMMD \cite{timemmd} and Context-Is-Key \cite{williams2025context} provide a collection of multimodal datasets that span healthcare, environment, finance, and retail domains. However, these efforts are restricted in the sizes of their individual datasets, and are mainly suitable for evaluation and not large-scale training. Additionally, curating these datasets is expensive owing to the significant human effort that was required. 

\subsection{Multimodal Forecasting Approaches} 
\label{sec:extended_related_approaches}
We group the existing works into zero-shot or training-free and dataset-specific training-based approaches. A common methodology for training-free approaches involves repurposing LLMs for time series forecasting, given their ability to process textual inputs efficiently. 

\subsubsection{Zero-Shot Approaches}

\citet{guo2024towards} perform multimodal traffic flow prediction by converting information from other modalities into textual prompts for an LLM to capture complex patterns in time series and associate them with textual annotations. \citet{kong2025time} fine-tuned LLMs (domain- or dataset-agnostic) with Low-Rank Adaptation (LoRA) for forecasting, imputation, and classification, by converting multimodal time series datasets and tasks into a dataset with questions and answers.

\citet{wang2024news} used reasoning agents to filter relevant news articles or textual annotations along with domain- or dataset-agnostic fine-tuning and evaluation steps that update LLMs to effectively utilize time series and textual annotations, and evaluate whether all relevant annotations have been retrieved, respectively. The authors show that the proposed approach generalizes for the electricity, finance, and traffic domains. 

In the finance domain, \citet{xie2023wall} evaluated ChatGPT's \cite{openai2024gpt4technicalreport} ability to forecast stock price movements. However, the results indicate that this approach does not perform better than multimodal approaches based on domain-specific training for the same task. \citet{yu2023temporal} focused on price prediction of NASDAQ-100 stocks using historical stock price data, company metadata, and financial news, with LLMs.

\citet{hybridmmf} proposed Time Text Corpus, a multimodal dataset with around 10,000 timestamps from the weather and the healthcare domains. Additionally, the authors introduced the Hybrid Multi-Modal Forecaster (Hybrid-MMF) which involves a two-stage training setup to jointly predict text and time series. However, the experiments are limited in both context (history, up to 7 timestamps) and forecast lengths.

\subsubsection{Dataset-Specific Training-Based Approaches}

\citet{li2026language} employed a learnable text encoder to project the time-varying textual annotations into the embedding space of learnable time series forecasting models, thereby enabling multimodal training for forecasting.

\citet{li2024urbangpt} presented UrbanGPT, a hybrid approach that integrates a spatiotemporal dependency encoder with the instruction-tuning paradigm in LLMs with the objective of forecasting traffic flow patterns in New York. This provides the ability for LLMs to understand complex spatial and temporal relationships in multivariate time series datasets. 

\citet{liu2023large} demonstrate the ability of LLMs to perform classification and prediction for physiological and medical time series using a few-shot tuning approach. 

\citet{chattopadhyay2024context} introduce ContextFormer, a dataset-specific fine-tuning procedure for unimodal time series forecasting models with textual annotations provided as embedding inputs from open-source LLMs.

\citet{jia2024gpt4mts} fine-tuned a pretrained GPT2 \cite{radford2019language} model with time series and textual annotations for large-scale events datasets, curated using the Global Database of Events, Language, and Tone (GDELT) tool. \citet{jia2024gpt4mts} generated relevant time-varying textual annotations from GDELT by filtering the required news articles using cosine similarity with respect to the natural language description of the forecasting target. The quantitative results indicate that the use of highly useful text annotations indeed helps in forecasting events more accurately than unimodal models trained on the same dataset. 

\citet{liu2025timecmallmempoweredmultivariatetime} employed a cross-modal alignment setup between the time series and text embeddings, for textual annotations that describe the trend and seasonality in the time series. Here, \citet{liu2025timecmallmempoweredmultivariatetime} trained a time series embedder from scratch and utilized off-the-shelf LLMs to extract text embeddings. Though the approach outperforms other SOTA trainable approaches, the text annotations mainly provide a summary of the time series and do not provide auxiliary information. Similarly, \citet{ekambaram2020attention} proposed an attention-based encoder-decoder framework for image-based product sales forecasting, while also including holidays, discounts, etc.

\citet{jiang2025explainable} proposed TimeXL, where a multimodal encoder processes both time series and textual inputs to generate initial forecasts and aided explanations. Then, a prediction LLM refines the forecasts by reasoning over or aligning the time series predictions with the explanations. Consequently, a reflection LLM compares the
predicted forecast against the ground truth to identify textual inconsistencies. Overall, TimeXL proposes a closed-loop framework, where the prediction LLM aligns the text and time series predictions, and the refinement LLM iteratively enhances text quality.

TimeLLM \cite{jin2024timellmtimeseriesforecasting} learns patch embeddings for time series to align time series representations with the token embedding space of LLMs, so that LLMs can be utilized to forecast or predict the next time series patch. Additionally, TimeLLM deals with static textual annotations, which are broad descriptions of the domain and the geography corresponding to the time series dataset. However, this approach does not deal with complex time-varying annotations that provide auxiliary information to aid forecasting.

TimeCAP \cite{lee2025timecap} utilizes two LLM agents, one summarizes the time series and the other uses aligned representations (learned using a multimodal encoder) of time series and its summary to forecast. Similar to other training-based approaches, TimeCAP performs dataset-specific training.

UniTime \cite{liu2024unitime} and \citet{hong2025unifyanchorcontextawaretransformer} deal with utilizing dataset descriptions (static textual annotations) to better learn time series forecasting for individual domains. Again, these annotations usually describe the dataset domain and the geographic region corresponding to where the dataset was collected, instead of time-varying auxiliary information, as in the case of finance datasets. \citet{hong2025unifyanchorcontextawaretransformer} also generate similar synthetic annotations for time series windows. However, they do not show the forecast values as inputs and primarily summarize the historical values.

Recently, UniCA \cite{han2026unica} proposed a unified covariate adaptation framework for TSFMs, where heterogeneous modalities such as text, categorical metadata, and numerical covariates are projected into a shared temporal representation space and fused into pretrained TSFMs using lightweight attention-based adapters. While UniCA enables covariate-aware forecasting across domains, the approach still relies on dataset-specific fine-tuning.

TimeVLM \cite{zhong2025time} uses pre-trained Vision-Language Models (VLMs), where the language encoders capture contextual information and the vision encoders capture trend and seasonality using time series features represented as images. The approach learns the parameters of the fusion layers that enrich time series representations with the image and the language embeddings for each dataset. Again, the annotations are restricted to the descriptions of the dataset domain. \citet{daswani2024plotsunlocktimeseriesunderstanding} and \citet{li2023timeseriesimagesvision} also suggest similar empirical conclusions regarding viewing time series as images or plots. \citet{daswani2024plotsunlocktimeseriesunderstanding} benchmarked the performances of the SOTA LLMs, such as GPT-4o \cite{openai2024gpt4technicalreport}, in various time series tasks, including classification, detection, and prediction. Similarly, \citet{li2023timeseriesimagesvision} used pre-trained vision encoders for time series classification tasks. \citet{zhou2025llmsunderstandtimeseries} explored anomaly detection using multimodal LLMs with time series inputs in the form of images.

\citet{sawhney2020deep} proposed an early fusion approach that combines learnable representations for time series and text annotations to predict the stock price movements. Similarly, \citet{chen2023chatgpt} employed ChatGPT \cite{openai2024gpt4technicalreport} to predict possible stocks that might be affected for the next day, along with the direction of change (positive or negative). This is combined with learnable time series embeddings, obtained from historical time series values, to predict the movements.

\citet{narasimhan2024time, narasimhan2025constrained} and \citet{ge2025t2shighresolutiontimeseries} focus on developing dataset-specific models for multimodal time series generation. Here, note that the inputs provided to the model are non-causal, \emph{i.e.,} the inputs describe the values of all the timestamps in the time series to be generated. In contrast, to evaluate forecasting, we require annotations that are causal and contain predictions or speculations about the forecast timestamps.

Although most of the approaches mentioned above reuse SOTA LLMs for forecasting, \citet{park2025revisitingllmszeroshottimeseries} highlighted the key issues that limit the ability of LLMs, such as sensitivity to noise. Additionally, \citet{park2025revisitingllmszeroshottimeseries} show the inferior performance of LLMs in comparison to light-weight linear models (DLinear) on standard time series forecasting benchmarks.

From all these approaches, we find some common trends: (i) Zero-shot / few-shot approaches mainly rely on the predictive power of LLMs; (ii) Most approaches are domain-specific and fine-tune LLMs/VLMs for individual datasets or domains; (iii) Previous works did not utilize powerful TSFMs such as Chronos2 \cite{chronos2}, TimesFM2.5 \cite{timesfm}, TabPFN-TS \cite{hollmann2023tabpfntransformersolvessmall}. We differentiate our approach from these prior works by seeking to develop a zero-shot solution that corrects/adapts unimodal forecasts from SOTA TSFMs using textual annotations. In addition, we are seeking to develop a framework that can work in multiple domains.

%% file: appendix/9_proofs.tex
\input{figures_latex/cosine_similarity_test}

\section{Proofs for Theorems in Section~\ref{sec:dataset_gen}}\label{appendix:theory_proofs}

\input{figures_latex/assumption1_proof}

Under our latent model, we have
history, forecast description, and the forecast sampled from $p(X^H,Z,X^F)$. Real annotations are
sampled as $S^r \sim p_r(S^r \mid X^H,Z)$, while synthetic annotations are sampled as
$S^s \sim p_s(S^s \mid X^H,Z,X^F)$. We define the real and synthetic
population risks as
\begin{align}
R_{\mathrm{multi}}^r(\theta)
&=
\mathbb{E}_{(X^H,Z,X^F)\sim p,\; S^r \sim p_r(\cdot \mid X^H,Z)}
\left[
\mathcal{M}\!\left(h^{\mathrm{multi}}_\theta(X^H,\phi(S^r)),X^F\right)
\right], \\
R_{\mathrm{multi}}^s(\theta)
&=
\mathbb{E}_{(X^H,Z,X^F)\sim p,\; S^s \sim p_s(\cdot \mid X^H,Z,X^F)}
\left[
\mathcal{M}\!\left(h^{\mathrm{multi}}_\theta(X^H,\phi(S^s)),X^F\right)
\right].
\end{align}

Here, note that real annotations are causal and carry  information about the forecast trends in domain-specific terms. Therefore, there is no conditioning on $X^F$ in $S^r \sim p_r(S^r \mid X^H,Z)$. However, during training, the synthetic annotations are generated using an LLM which considers the history $X^H$ and the forecast $X^F$ to generate the domain-specific description. Additionally, we note that $Z$ can be extracted from $X^F$ easily by computing certain domain-agnostic statistical features. Therefore, for synthetic annotations, we have $S^s \sim p_s(S^s \mid X^H,Z,X^F)$. 

For analysis, we also define the risk with the true domain-agnostic forecast description.
\begin{equation}
R_{\mathrm{multi}}^Z(\theta)
=
\mathbb{E}_{(X^H,Z,X^F)\sim p}
\left[
\mathcal{M}\!\left(h^{\mathrm{multi}}_\theta(X^H,\psi(Z)),X^F\right)
\right].
\end{equation}

\begin{proof}[\textbf{Proof of Lemma~\ref{lemma:mismatch}}]
We first introduce $R_{\mathrm{multi}}^Z(\theta)$.
\begin{align}
|R_{\mathrm{multi}}^r(\theta)-R_{\mathrm{multi}}^s(\theta)|
= |R_{\mathrm{multi}}^r(\theta)- R_{\mathrm{multi}}^Z(\theta) + R_{\mathrm{multi}}^Z(\theta) -R_{\mathrm{multi}}^s(\theta)|
\end{align}
Now, using the triangle inequality, we have
\begin{align}
|R_{\mathrm{multi}}^r(\theta)-R_{\mathrm{multi}}^s(\theta)|
&\leq
|R_{\mathrm{multi}}^r(\theta)-R_{\mathrm{multi}}^Z(\theta)|
+
|R_{\mathrm{multi}}^Z(\theta)-R_{\mathrm{multi}}^s(\theta)|.
\label{eq:lemma_ineq}
\end{align}

For the first term, we have
\begin{align}
&|R_{\mathrm{multi}}^r(\theta)-R_{\mathrm{multi}}^Z(\theta)| \nonumber\\
&=
\Bigg|
\mathbb{E}_{p,\,p_r}
\Big[
\mathcal{M}\!\left(h^{\mathrm{multi}}_\theta(X^H,\phi(S^r)),X^F\right)
-
\mathcal{M}\!\left(h^{\mathrm{multi}}_\theta(X^H,\psi(Z)),X^F\right)
\Big]
\Bigg| \nonumber\\
&\leq
\mathbb{E}_{p,\,p_r}
\Big[
\Big|
\mathcal{M}\!\left(h^{\mathrm{multi}}_\theta(X^H,\phi(S^r)),X^F\right)
-
\mathcal{M}\!\left(h^{\mathrm{multi}}_\theta(X^H,\psi(Z)),X^F\right)
\Big|
\Big] \nonumber\\
&\leq
L\,
\mathbb{E}_{p,\,p_r}
\left[
\|\phi(S^r)-\psi(Z)\|
\right] \label{eq:proof_one_mid} \\
&\leq
L\varepsilon_r.
\end{align}

Here the first inequality is from Jensen's inequality, the second inequality follows from \textbf{Assumption~\ref{assump:lipschitz}} and the
last inequality follows from \textbf{Assumption~\ref{assump:semantic}}. 

Similarly, for the second term in Equation \ref{eq:lemma_ineq}, we have
\begin{align}
|R_{\mathrm{multi}}^s(\theta)-R_{\mathrm{multi}}^Z(\theta)|
&\leq
L\,
\mathbb{E}_{p,\,p_s}
\left[
\|\phi(S^s)-\psi(Z)\|
\right]
\leq
L\varepsilon_s.
\label{eq:proof_one_end}
\end{align}
Combining the two bounds gives
\begin{equation}
|R_{\mathrm{multi}}^r(\theta)-R_{\mathrm{multi}}^s(\theta)|
\leq
L(\varepsilon_r+\varepsilon_s).
\end{equation}
\end{proof}

\begin{proof}[\textbf{Proof of Theorem~\ref{thm:transfer}}]
We decompose $R_{\mathrm{multi}}^r(\hat{\theta}_s)-R_{\mathrm{multi}}^r(\theta_r^\star)$ as

\begin{align}
R_{\mathrm{multi}}^r(\hat{\theta}_s)-R_{\mathrm{multi}}^r(\theta_r^\star) 
&=
\underbrace{
R_{\mathrm{multi}}^r(\hat{\theta}_s)-R_{\mathrm{multi}}^s(\hat{\theta}_s)
}_{\text{(I)}} \\
&+
\underbrace{
R_{\mathrm{multi}}^s(\hat{\theta}_s)-R_{\mathrm{multi}}^s(\theta_r^\star)
}_{\text{(II)}} \\
&+
\underbrace{
R_{\mathrm{multi}}^s(\theta_r^\star)-R_{\mathrm{multi}}^r(\theta_r^\star)
}_{\text{(III)}}.
\end{align}

Term (II) is non-positive since $\hat{\theta}_s$ minimizes
$R_{\mathrm{multi}}^s$ over $\Theta$. By \textbf{Lemma~\ref{lemma:mismatch}}, terms (I)
and (III) are each bounded above by $L(\varepsilon_r+\varepsilon_s)$. Therefore,
\begin{equation}
R_{\mathrm{multi}}^r(\hat{\theta}_s)
\leq
R_{\mathrm{multi}}^r(\theta_r^\star)
+
2L(\varepsilon_r+\varepsilon_s).
\end{equation}
\end{proof}

Observe from Table \ref{tab:rephrasing_effect} that changing the text only affects the MAE numbers in the third decimal for the LLM forecast as well as CoCo. To this end, we note that for reasonable forecasters, the smoothness property is generally satisfied in practice. As we explained in Section \ref{sec:dataset_gen}, the key ingredient for sim-to-real transfer is the proximity of real and synthetic annotations to the real future description in some latent space. Given the existence of foundation models that can act as powerful encoders, if the annotations contain speculations of the expected trend in the forecast, it is fair to expect small values for $\varepsilon_s$ and $\varepsilon_r$, indicating good proximity in the latent space. Consequently, the bounded gap of $2L(\varepsilon_r+\varepsilon_s)$ tends to zero for very small $\varepsilon_s$ and $\varepsilon_r$, indicating that the model trained on synthetic annotations will perform as well as the model trained on real annotations. Additionally, our claim of real and synthetic annotations capturing similar domain-agnostic predictions in Section \ref{sec:dataset_gen} is visually depicted in Figure \ref{fig:tsne_real_synth}.

\input{figures_latex/tsne_real_synthetic}

\subsection{Assumption Validity Discussion}\label{appendix:assumptions_discussion}

\textbf{Assumption \ref{assump:semantic}} focuses on the proximity of domain-agnostic forecast descriptions inferred from real and synthetic annotations to the domain-agnostic description of the true future in a common representation space. In Figure \ref{fig:assumption1_part_proof}, we show the strong overlap in cosine similarity distributions between the embeddings of the verbal description of the forecast predictions derived from real and synthetic annotations. Note that the cosine similarity is with respect to the true future descriptions. As mentioned in Section \ref{sec:dataset_gen}, this indicates that synthetic annotations recover the forecast nearly as well as real ones, suggesting $\varepsilon_s \approx \varepsilon_r$. Additionally, the generation process grounds synthetic text in numerical trajectories and domain descriptions. In our experiments, the embeddings correspond to normalized BERT embeddings, and consequently, the cosine similarity and the Euclidean distance (used in the assumption) are equivalent measures.
 
To this end, we directly quantify these representation distances on our real-world evaluation datasets using the same unit-normalized BERT embeddings, on a per-domain basis, in Table \ref{tab:forecast_description_distance}. Across all domains, synthetic annotations exhibit nearly the same distance to the true forecast description as real annotations. Therefore, the significant overlap in the cosine similarity distribution (Figure \ref{fig:assumption1_part_proof}) with a very high mean ($\sim$ 0.93) and the similar representation distances between the real and synthetic annotations from the true forecast descriptions (Table \ref{tab:forecast_description_distance}) convey the same idea that real and synthetic annotations encode domain-agnostic information similar to the true forecast description. 

Additionally, we note that the Equations \ref{eq:proof_one_mid} and \ref{eq:proof_one_end} require only the expected representation distances to be bounded rather than the per-sample bound stated in Assumption \ref{assump:semantic}. Therefore, relaxing Assumption \ref{assump:semantic} to reflect the bound on the expected representation distances between the real and synthetic forecast descriptions and the true forecast descriptions would still be sufficient to prove Lemma \ref{lemma:mismatch} and Theorem \ref{thm:transfer}.

\textbf{Assumption~\ref{assump:lipschitz}} focuses on ensuring that small perturbations in the text embedding produce limited changes in forecasting errors. Our empirical evidence in Tables~\ref{tab:rephrasing_effect}, \ref{tab:text_perturbation}, and \ref{tab:adversial_text} shows that local perturbations that do not adversely affect the forecast-relevant semantics cause negligible variation in forecasting error. Conversely, for perturbations that alter the forecast relevant semantics (see Tables \ref{tab:text_perturbation} and \ref{tab:adversial_text}), we observe a significant change in the forecasting performance. Consequently, we note that this empirical result points only towards local smoothness in the loss space with respect to small perturbations in text annotations for reasonably good forecasters rather than the uniform Lipschitz condition currently stated in Assumption \ref{assump:lipschitz}. 

Overall, the theoretical bound on the risk in Theorem \ref{thm:transfer} remains conditional on these assumptions and should be interpreted as explaining when sim-to-real transfer is expected to hold, rather than guaranteeing it for arbitrary cases.

%% file: figures_latex/cosine_similarity_test.tex
\begin{figure}
\centering
\includegraphics[width=0.7\linewidth]{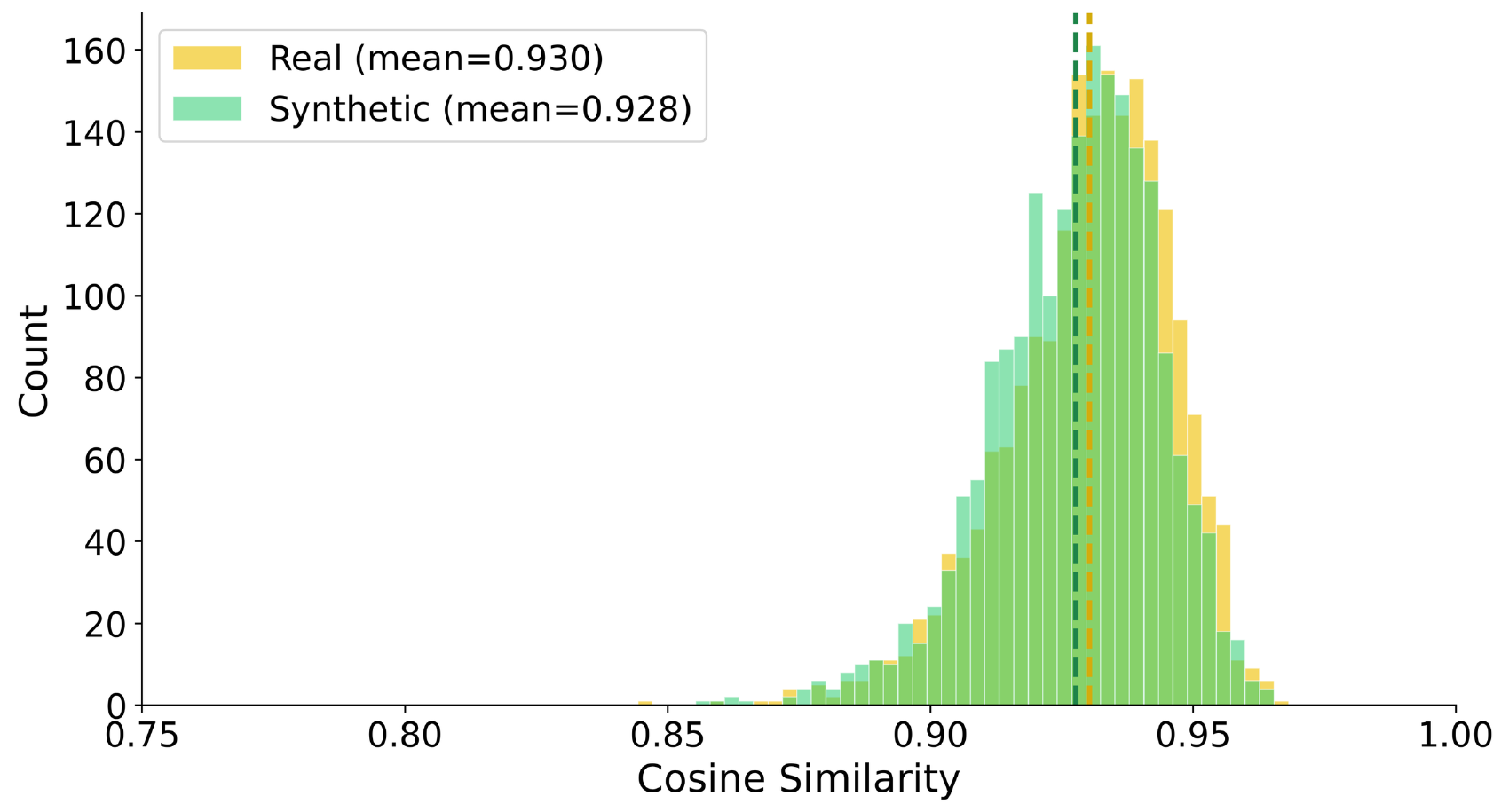}
\caption{\textbf{Real and synthetic annotations contain similar predictive powers to estimate the future.} The significant overlap of cosine similarity distributions between the true and the predicted forecast descriptions for synthetic and real annotations validates our synthetic annotation strategy. For the real cosine similarity distribution, we compare the normalized BERT embeddings for the domain-agnostic forecast predictions from real text with the true domain-agnostic forecast description. Similarly, for the synthetic cosine similarity distribution, we compare the normalized BERT embeddings for the domain-agnostic forecast predictions from synthetic text with the true domain-agnostic forecast description.  For this experiment, we consider the Media Cloud datasets for which the real annotations are available, and we generated synthetic annotations for those datasets.}
\label{fig:assumption1_part_proof}
\vspace{-6mm}
\end{figure}

%% file: figures_latex/assumption1_proof.tex
    


%% file: figures_latex/tsne_real_synthetic.tex
\begin{figure}
\centering
\includegraphics[width=0.9\linewidth]{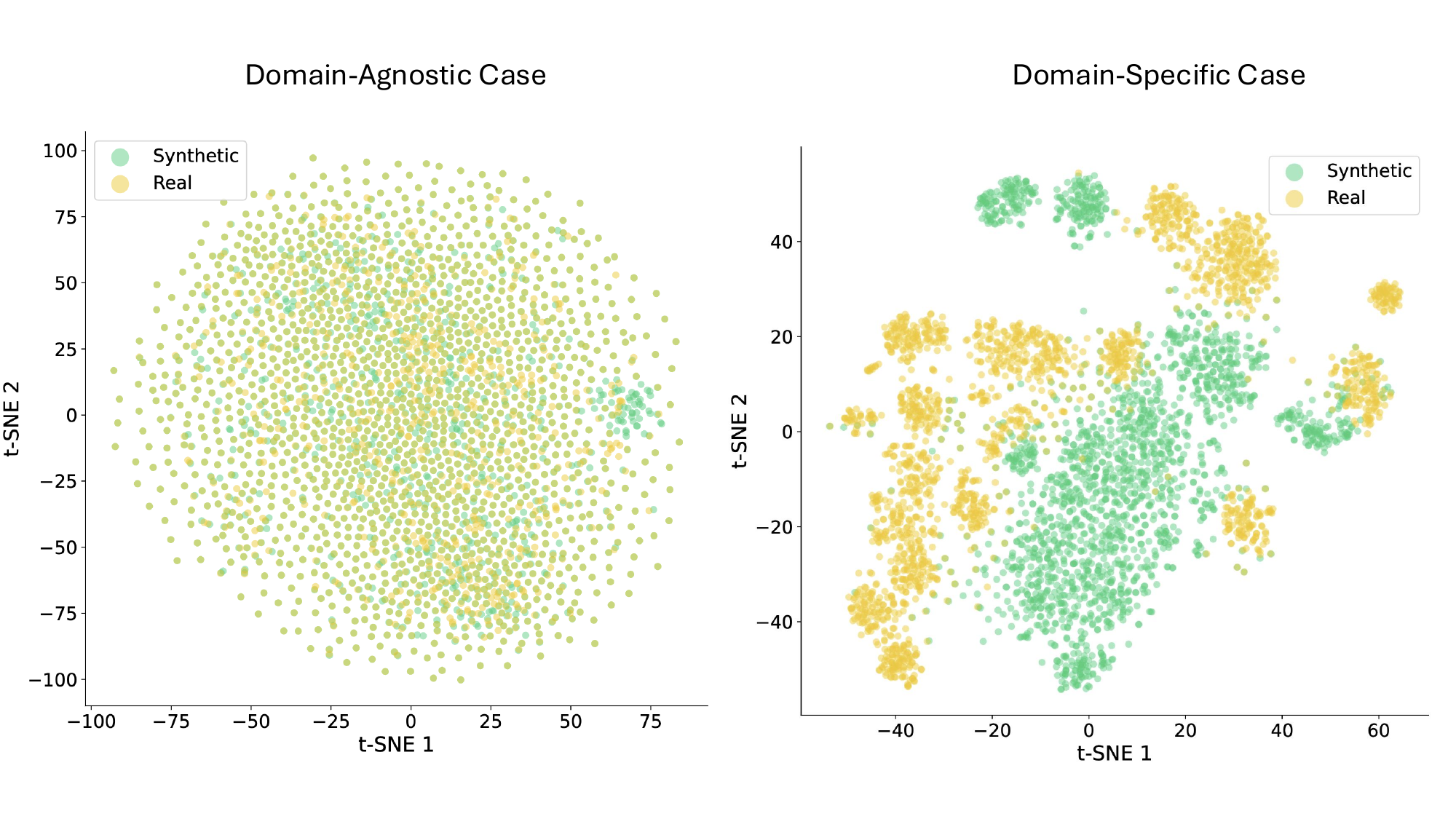}
\caption{\textbf{Real and synthetic annotations contain similar domain-agnostic descriptions, but differ in semantics.} Here, we show the t-SNE plots for concatenated embeddings of real and synthetic annotations. We consider the Media Cloud datasets for which the real annotations are available, and we generated synthetic annotations for those datasets. On the left, we compare the embeddings of real domain-agnostic predictions concatenated with true forecast descriptions against the synthetic counterpart. Similarly, on the right, we compare the embeddings of the real domain-specific prediction concatenated with the true forecast descriptions against the synthetic counterpart. Note that there is near-perfect overlap in the domain-agnostic case and clear separation in the domain-specific case. This conveys that for the same true forecast description (through concatenation), the domain-agnostic predictions overlap indicating that real and synthetic annotations provide similar domain-agnostic prediction. In contrast, the domain-specific predictions vary significantly, indicating that both real and synthetic domain-specific predictions have different semantic constructions.}
\label{fig:tsne_real_synth}
\vspace{-6mm}
\end{figure}

%% file: appendix/11_synthetic_example.tex
\section{Synthetic Toy Example Description}
\label{app:toy_description}

We construct a controlled, fully synthetic toy series of length $T = 1100$ daily steps that combines a SARIMA-like latent process with discrete, text-driven shocks.  The latent baseline $y^{\mathrm{base}}_t$ is generated
recursively from a linear trend, weekly seasonality, and an AR(1) autoregressive component centered on a fixed level:
\begin{equation}
  y^{\mathrm{base}}_t
  = \mu + s\,t
    + \phi\,\bigl(y^{\mathrm{base}}_{t-1} - \mu\bigr)
    + A\sin\!\left(\frac{2\pi t}{P}\right)
    + \varepsilon_t,
  \qquad \varepsilon_t \sim \mathcal{N}(0, \sigma^2),
  \label{eq:toy-baseline}
\end{equation}
with parameters $\mu = 50$ (bias), $s = 0.008$ (linear trend slope),
$\phi = 0.6$ (AR coefficient), seasonal amplitude $A = 3.0$ and periodicity $P = 7$, and $\sigma = 1.2$,
initialized at $y^{\mathrm{base}}_0 = \mu$. 

The text component is added through a curated catalog of 30 discrete events, each represented by a tuple
$(t_s, t_e, \ell, \mathrm{label}, \mathrm{text})$ with start day $t_s$, end day $t_e$, signed sentiment level $\ell \in [-1, 1]$, a short tag, and a natural-language description of the event in an analyst style. The first 25
events are used for training and the last 5 are reserved as held-out test windows. From these events we construct a piecewise-constant sentiment signal $Z_t$ for length $T$. Every $K$ timestamps before the start and after the end of the event, $Z_t$ linearly increases from 0 to $\ell$ and linearly decreases from $\ell$ to 0, respectively. During the event $Z_t$ stays constant at $\ell$. We set $K$ to 4. Consequently, the observed series is
\begin{equation}
  Y_t = y^{\mathrm{base}}_t + \beta\, Z_t,
  \label{eq:toy-observation}
\end{equation}
with text coupling strength $\beta = 8.0$, which makes the text-driven
deviations clearly visible against the natural fluctuations of the base signal as described in Equation \ref{eq:toy-baseline}. 

We first consider training on real and testing on real (TRTR). By training, we refer to regressing the parameters $\mu, s, \phi, A,$ and $\beta$. We use the \texttt{least\_squares} function from \texttt{scipy}. This allows us to obtain the estimate of the underlying model used for data generation. With this model, we simulate the held-out test horizon and compute the MAE for test event timestamps. During training, we do not utilize the true sentiment levels. Instead, we use Sentence Transformer to estimate the sentiment level.

To showcase the effectiveness of training on synthetic annotations, we use LLMs to annotate the events, using the same approach described in Section \ref{sec:dataset_gen}. We extract sentiment levels from the generated synthetic annotations using the same approach as in TRTR and estimate $\mu, s, \phi, A,$ and $\beta$. Consequently, we use these estimated parameters to simulate the test events with sentiments extracted from real text and compute the MAE (TSTR). 

The results in Figure \ref{fig:sim_to_real_toy} show that the TRTR and TSTR MAEs are very similar, thereby establishing the veracity of our \textbf{sim-to-real} approach for multimodal forecasting, for this specific toy scenario.

%% file: tables/rephrasing_effect.tex
\begin{table}[t]
    \centering
    \small
    \renewcommand{\arraystretch}{1.12}
    \resizebox{0.6\textwidth}{!}{%
    \begin{tabular}{lcccccccccccc}
    \toprule
    & \multicolumn{2}{c}{MAE ($\downarrow$)} \\
    \cmidrule(lr){2-3}
     Dataset & \multicolumn{1}{c}{LLM Forecast ($\theta_\mathrm{LLM}(X^H_t, S_t)$)}
     & \multicolumn{1}{c}{CoCo ($\theta_\mathrm{M-TSFM}{}(X^H_t||M_t||C_t)$)} \\
    \midrule
    Raw
     & 0.878 & \textbf{0.858}
     \\
    Rephrased
     & \textbf{0.875} & 0.859
     \\
    \bottomrule
    \end{tabular}
    }%
    \vspace{0.1in}
    \caption{\textbf{CoCo is less sensitive to text rephrasing.} We observe that the MAE of the LLM and the MAE of CoCo change marginally with rephrasing as long as the text conveys the same information. This validates \textbf{Assumption \ref{assump:lipschitz}} (smoothness assumption) indicating that small perturbations in the context $S_t$ do not significantly affect the forecasting performance.}
    \label{tab:rephrasing_effect}
\end{table}

%% file: tables/rebuttal/text_perturbation_analysis.tex
\begin{table}[t]
    \centering
    \small
    \renewcommand{\arraystretch}{1.12}
    \resizebox{0.6\textwidth}{!}{%
    \begin{tabular}{lcccccccccccc}
    \toprule
     Perturbation & \multicolumn{1}{c}{\% Change in the LLM Forecast ($\theta_\mathrm{LLM}(X^H_t, S_t)$)} MAE ($\downarrow$) \\
    \midrule
    Rephrased
     & 0.3
     \\
        Random Word Deletion ($5\%$)
     & 0.8
     \\
        Less Verbose Annotations
     & 1.1
     \\
        Sentence Reordering
     & 1.02
     \\
        Using Annotations From Other Datasets
     & 5.0
     \\
        Misleading Text
     & 7.9
     \\
    \bottomrule
    \end{tabular}
    }%
    \vspace{0.1in}
    \caption{\textbf{CoCo exhibits local smoothness with respect to text perturbations.} For simple perturbations that do not affect the forecast trend indicators (first 4 rows), note that the change in the LLM forecast MAE is less than or around $1\%$, indicating local smoothness towards mild perturbations. However, for perturbations that change the underlying forecast (last two rows), the change is significant.}
    \label{tab:text_perturbation}
\end{table}

%% file: tables/rebuttal/representation_distance.tex
\begin{table}[t]
\centering
\resizebox{0.6\linewidth}{!}{
\begin{tabular}{lccc}
\toprule
\textbf{Domain} &
\begin{tabular}[c]{@{}c@{}}
\textbf{Mean Distance Between}\\
\textbf{Real and True Forecast}\\
\textbf{Description Representation}
\end{tabular} &
\begin{tabular}[c]{@{}c@{}}
\textbf{Mean Distance Between}\\
\textbf{Synthetic and True Forecast}\\
\textbf{Description Representation}
\end{tabular} &
\begin{tabular}[c]{@{}c@{}}
$\boldsymbol{\Delta}$\\
\textbf{Mean}
\end{tabular} \\
\midrule
\textbf{All domains} & 0.3715 & 0.3774 & +0.0059 \\
Agriculture          & 0.3718 & 0.3735 & +0.0017 \\
Energy               & 0.3744 & 0.3745 & +0.0001 \\
Macro Indicators     & 0.3720 & 0.3805 & +0.0085 \\
Retail               & 0.3759 & 0.3744 & -0.0015 \\
Finance              & 0.3707 & 0.3775 & +0.0068 \\
\bottomrule
\end{tabular}
}
\caption{\textbf{Our synthetic annotations encode similar information about the true future as the real annotations.} We empirically show this in Figure \ref{fig:assumption1_part_proof}. However, as Assumption \ref{assump:semantic} uses the Euclidean distance measure, we also provide a per-domain breakdown of the $l_2$ norm between the real and synthetic predicted representations and the representation of the true forecast description. Here, the representation space used is the unit normalized BERT embedding space.}
\label{tab:forecast_description_distance}
\end{table}

%% file: appendix/8_context_is_key.tex
\section{Evaluation on Context is Key (CiK)}
\label{app:cik}
Context is Key (CiK)~\cite{williams2025context} is a time series forecasting benchmark that pairs numerical 
data with diverse types of carefully crafted textual context, where crucially every task requires understanding textual context to be solved successfully. We evaluate on CiK to understand the behavior of CoCo's components in a setting that is deliberately adversarial to unimodal numerical forecasting. We first establish where Chronos2 stands on the CiK leaderboard as a unimodal baseline. Chronos2 achieves an average RCRPS of 0.265, outperforming Chronos-Large (0.326), TimeGEN  (0.353), Moirai-Large (0.520), all statistical baselines (ARIMA: 
0.475, ETS: 0.530), and the majority of LLM-based methods on the leaderboard. This is notable given that CiK is explicitly designed  so that numerical history alone is insufficient.



\input{tables/context_is_key_llm}

Table~\ref{tab:cik} reports the average RCRPS (Region of Interest Continuous Ranked Probability Score) on CiK for 
Qwen3-4b under four configurations: base model without reasoning (as evaluated on CiK), base model with a short chain-of-thought reasoning prompt, 
CoCo fine-tuned model without reasoning, and 
CoCo fine-tuned model prompted for reasoning. The reasoning prompt 
used is:


\begin{quote}
    \textit{``Before the forecast, provide a reasoning section between <reasoning> and </reasoning> tags.''}
\end{quote}


Adding a chain-of-thought reasoning prompt reduces RCRPS by $47\%$ for the base model (0.358 $\rightarrow$ 0.189) and by $46\%$ for the 
fine-tuned model (0.301 $\rightarrow$ 0.161). This improvement from a single prompt modification substantially exceeds the gain from fine-tuning alone in this setting. This suggests that CiK rewards 
explicit chain-of-thought reasoning. Fine-tuning provides complementary gains on top of reasoning. While the reasoning prompt dominates, fine-tuning still contributes meaningfully when combined with reasoning (0.189 $\rightarrow$ 0.161, a further $15\%$ improvement). The fine-tuned model with reasoning matches GPT-4o, despite being much smaller in parameter count. This confirms that GRPO fine-tuning for covariate generation teaches transferable forecasting capabilities that generalize to a held-out benchmark the model was never trained on.


Fundamentally, CiK differs from our main evaluation benchmarks in a key respect: every CiK task is constructed so that textual context is the \emph{primary} source of predictive signal, with 
numerical history deliberately insufficient. In this regime, the TSFM backbone adds limited value. In contrast, our evaluation datasets contain naturally occurring noisy text where numerical history remains genuinely 
informative, the complementary setting that CoCo is 
designed for, and where CoCo consistently outperforms both the LLM and the TSFM in isolation.

%% file: tables/context_is_key_llm.tex
\begin{table}[H]
\centering
\small
\renewcommand{\arraystretch}{1.15}

\resizebox{\textwidth}{!}{
\begin{tabular}{lcccccc}
\toprule
 & \textbf{Average} & \textbf{Intemporal} & 
\textbf{Historical} & \textbf{Future} & \textbf{Covariate} & 
\textbf{Causal} \\
 \textbf{Model} & \textbf{RCRPS} $\downarrow$ & \textbf{Information} & 
\textbf{Information} & \textbf{Information} & \textbf{Information} & 
\textbf{Information} \\
\midrule
Qwen3-4b (Base) 
    & 0.358 ± 0.031 & 0.390 ± 0.045 & 0.247 ± 0.023 
    & 0.301 ± 0.042 & 0.310 ± 0.035 & 0.700 ± 0.149 \\
Qwen3-4b (CoCo Fine-tuned) 
    & 0.301 ± 0.027 & 0.306 ± 0.040 & 0.219 ± 0.019 
    & 0.276 ± 0.036 & 0.280 ± 0.033 & 0.675 ± 0.142 \\
Qwen3-4b (Base) + Reasoning 
    & 0.189 ± 0.036&0.159 ± 0.011&0.134 ± 0.011&\textbf{0.088 ± 0.012}&0.204 ± 0.047&0.523 ± 0.214 \\
Qwen3-4b (CoCo Fine-tuned) + Reasoning 
    & \textbf{0.161 ± 0.013}&\textbf{0.154 ± 0.010}&\textbf{0.130 ± 0.012}&0.090 ± 0.012&\textbf{0.171 ± 0.016}&\textbf{0.378 ± 0.064} \\
\midrule
Chronos2 (Unimodal) 
    & 0.265 ± 0.016 & 0.254 ± 0.021 & 0.166 ± 0.017 
    & 0.293 ± 0.031 & 0.239 ± 0.018 & 0.438 ± 0.066 \\
\midrule
CoCo (Fine-tuned + Reasoning) 
    & 0.256 ± 0.017&0.242 ± 0.023&0.137 ± 0.011&0.281 ± 0.029&0.234 ± 0.019&0.438 ± 0.074 \\
\midrule
\textit{GPT-4o (reference)} 
    & \textit{0.191 ± 0.004} & \textit{0.218 ± 0.007} 
    & \textit{0.118 ± 0.001} & \textit{0.121 ± 0.001} 
    & \textit{0.143 ± 0.003} & \textit{0.363 ± 0.011} \\
\textit{Llama-3.1-405B (reference)} 
    & \textit{0.143 ± 0.006} & \textit{0.174 ± 0.010} 
    & \textit{0.146 ± 0.001} & \textit{0.075 ± 0.005} 
    & \textit{0.144 ± 0.008} & \textit{0.303 ± 0.037} \\
\bottomrule
\end{tabular}
}
\vspace{0.1in}
\caption{\textbf{RCRPS on the CiK benchmark~\cite{williams2025context}}. We provide a breakdown of RCRPS by information type, and evaluate the effect of fine-tuning with CoCo on the reasoning capacity of the LLM. \textbf{Bold} denotes the best result among our models per column.  Reference numbers for GPT-4o and Llama-3.1-405B are from the official CiK leaderboard (\url{https://github.com/servicenow/context-is-key-forecasting}) and shown in \textit{italics} for context. Lower is better. Our key takeaway from the results is that LLMs fine-tuned with CoCo consistently outperform their base, pre-trained counterpart. Additionally, a Qwen3-4B model that is fine-tuned with CoCo, when prompted for reasoning, can match the performance of powerful closed-source LLMs, such as GPT-4o.}
\label{tab:cik}
\end{table}

%% file: appendix/2_training_description.tex
\subsection{Training Description}
\label{app:training_description}

We fine-tune $\theta_{\mathrm{LLM}}$ and $\theta_{\mathrm{M-TSFM}}$ in two sequential stages, as described in Section~\ref{sec:model}. Both stages use parameter-efficient LoRA adapters \cite{lora} so that the base weights remain frozen at their public checkpoints, and only the adapters are written to disk and exchanged across the training/inference processes.

\textbf{Window Sampling and Normalization:} Each training window is sampled by independently drawing a context length $l \in \{64, 96, 128, 192, 256\}$ and a forecast horizon $h \in \{16, 24, 32, 40\}$ uniformly at random, and then choosing a random valid start index in the source series. This exposes both stages to the full range of $(l, h)$ combinations rather than a single fixed pair. Sampling longer histories is deliberately included even though Chronos2's unimodal accuracy improves rapidly with $l$. It forces the LLM to remain useful in the regime where the numerical baseline is already strong, and yields a forecast covariate $M_t$ whose marginal benefit (Equation~\ref{eq:forecast_covariate}) does not collapse as $l$ grows. For every sampled window, the numerical history $X^H_t = [x_{t-l+1}, \ldots, x_t]$ is z-score standardized using per-window statistics $\mu_t$ and $\sigma_t$ (mean and standard deviation) computed over the context portion only, and the ground truth $X^F_t = [x_{t+1}, \ldots, x_{t+h}]$ is normalized with the same $(\mu_t, \sigma_t)$. The LLM is conditioned on raw (unscaled) values in the prompt but is supervised in the normalized space (i.e., the reward MAE in Section~\ref{sec:model} is computed after applying $(\mu_t, \sigma_t)$ to the parsed forecast), while Chronos2 operates entirely in the per-window normalized space. The evaluation MAEs reported are in the normalized space, so that they are comparable across datasets. We observe that the relative performance of models does not change significantly for metrics computed in the de-normalized space as well (Table \ref{tab:average_rank_all_models}).

\textbf{LLM Fine-Tuning:} The base LLM $\theta_{\mathrm{LLM}}$ is initialized from \texttt{Qwen3-4B-Instruct-2507} and adapted with rank-$32$ LoRA modules ($\alpha=64$, dropout $0$, no bias) attached to all attention and MLP projection matrices ($q$, $k$, $v$, $o$, gate, up, down) across the $36$ transformer blocks, yielding $\sim\!61.6$M trainable parameters ($\sim\!1.5\%$ of the $\sim\!4$B base). 
We optimize the GRPO objective of Section~\ref{sec:model} with $N_\mathrm{rollouts}=8$ rollouts per prompt sampled at temperature $1.0$, a prompt batch size of $256$, and AdamW at a learning rate of $10^{-6}$, for up to $500$ optimizer steps. The reward $r^{(g)}_t$ is computed against the \emph{frozen} pretrained Chronos2 (\texttt{amazon/chronos-2}) and the resulting GRPO advantage $A^{(g)}_t$ updates only the LoRA adapters of $\theta_{\mathrm{LLM}}$. Training prompts are drawn at a $0.5$/$0.5$ ratio from our synthetic corpus and FNSPID real-world annotations with a periodicity stratifier that balances periodic and aperiodic series.
We run the training loop with PrimeRL \cite{primerl}, splitting GPUs into $2$ trainers and $2$ vLLM \cite{vllm} inference workers (4 H100s). 

\textbf{Textual Context:} Although the LLM prompt always includes the full numerical history $X_t$ of length $l$, the textual annotation $S_t$ is restricted to only the most recent $K_\mathrm{train}=10$ timesteps during fine-tuning. Concretely, only the text fields associated with $\{x_{t-K_\mathrm{train}+1}, \ldots, x_t\}$ are inserted into the prompt's annotation block. Older annotations, if present, are discarded. We deliberately use a smaller $K_\mathrm{train}$ at training time than the $K_\mathrm{eval}=16$ used at evaluation. Textual annotations are dense and verbose, and including all $16$ steps in every prompt, in combination with the longest contexts $l=256$, the $N_\mathrm{rollouts}=8$ on-policy completions per GRPO update, and LoRA activations stored across transformer blocks, pushed the per-rollout sequence length above the $8{,}192$-token budget that fits within the LoRA fine-tuning memory on our $2$-trainer setup. Truncating to $K_\mathrm{train}=10$ keeps the per-rollout sequence length comfortably under that budget while preserving the most causally relevant annotations. At inference time, this constraint disappears and we expand the textual window back to $K_\mathrm{eval}=16$, matching the value used by all closed-source LLM baselines for a fair comparison.

\textbf{Chronos2 Fine-Tuning:} Once $\theta_{\mathrm{LLM}}$ has converged, we pre-generate the covariate set $\{M_t, C_t\}_{t=1}^N$ from Equations~\ref{eq:forecast_covariate} and \ref{eq:confidence_covariate} using $N_\mathrm{rollouts}=5$ rollouts per sample for the confidence channel. The forecaster $\theta_{\mathrm{M-TSFM}}$ is initialized from \texttt{amazon/chronos-2} ($\sim\!119.5$M parameters, $d_{\text{model}}=768$, $d_{\text{ff}}=3072$, $12$ encoder blocks) and adapted with rank-$8$ LoRA modules ($\alpha=16$, dropout $0$, no bias) attached to the self-attention $q$, $k$, $v$, $o$ projections in every encoder block and to the output projection of the patch-embedding head,
totaling $\sim\!1.21$M trainable parameters ($\sim\!1.0\%$ of the base). The base weights are kept frozen and only the LoRA adapters are trained by minimizing $\frac{1}{N} \sum_{t=1}^{N} \left\| \theta_{\mathrm{M-TSFM}}(X^H_t \| M_t \| C_t) - X^F_t \right\|^2$ with AdamW for $10,000$ steps, with model selection by validation MAE on a held-out $5\%$ split. The fine-tuned adapters are merged with the base \texttt{amazon/chronos-2} weights only at inference time.

\textbf{Noise Level for Covariate Construction.} The forecast covariate $M_t$ (Eq.~\ref{eq:forecast_covariate}) and confidence covariate $C_t$ (Eq.~\ref{eq:confidence_covariate}) include 
i.i.d.\ noise $\epsilon_i \sim \mathcal{N}(0, \sigma_\epsilon^2)$ in the history portion to allow Chronos2 to interpret it as a 
noisy next-step predictor. Table~\ref{tab:noise_level} shows 
the effect of $\sigma_\epsilon$ on forecasting performance. Neither removing noise entirely ($\sigma_\epsilon = 0.00$) nor using high noise levels ($\sigma_\epsilon = 1.00$) yields the best performance. The former provides no signal diversity for Chronos2 to calibrate against, while the latter obscures the next-step information in the history portion. A small noise level of $\sigma_\epsilon = 0.05$ achieves the best MAE and is used throughout all experiments for both fine-tuning and inference.

\input{figures_latex/grpo_reward}
\input{tables/noise_table}

%% file: figures_latex/grpo_reward.tex
\begin{figure}[h]
    \centering
    \includegraphics[width=0.65\linewidth]{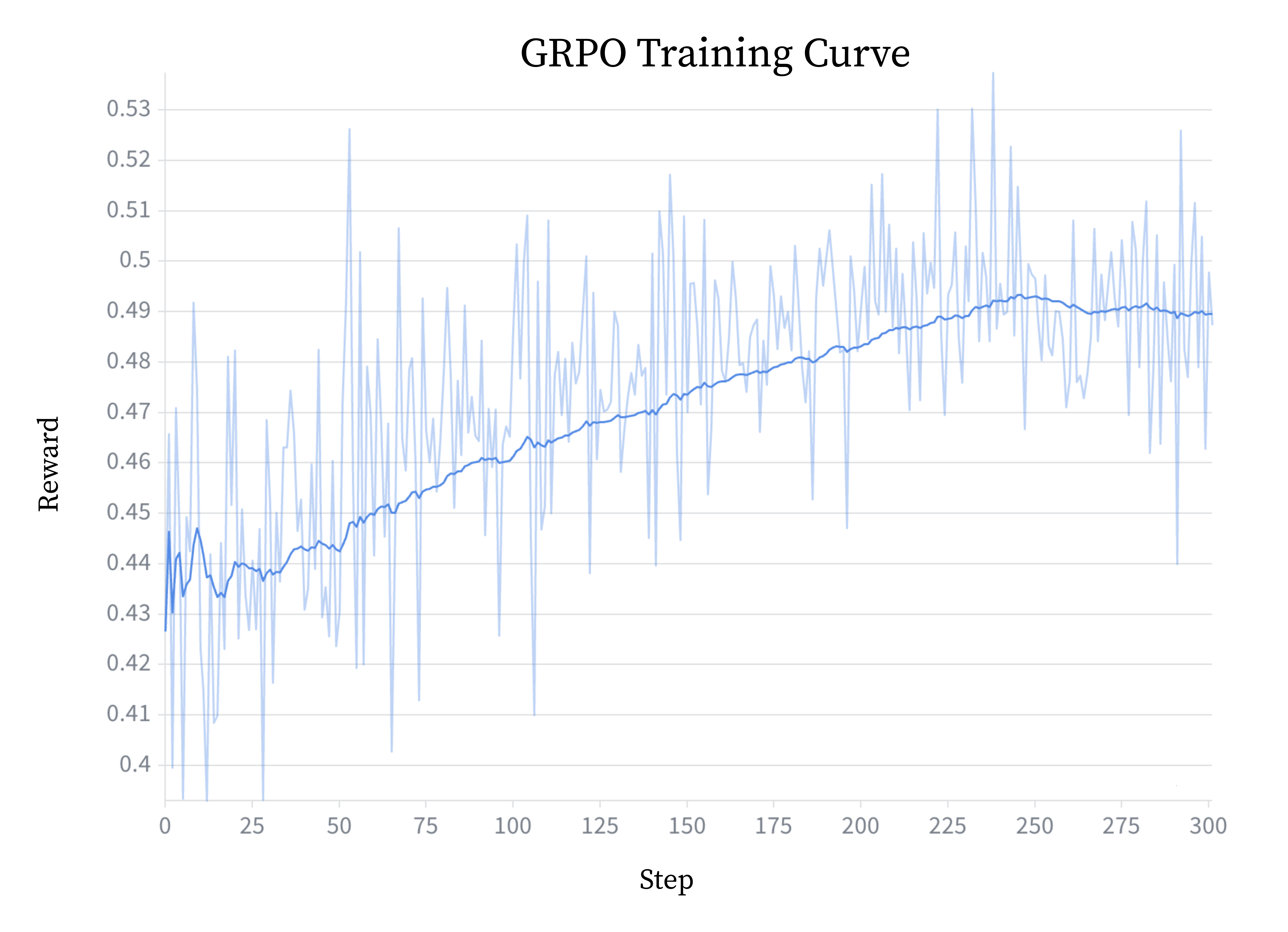}
    \caption{Average GRPO training reward $r_t$ 
(Eq.~\ref{eq:grpo_reward}) over training steps, computed as the mean over $N_\text{rollouts}$ rollouts per prompt in a batch. The reward increases steadily reflecting that the LLM progressively learns to generate forecast covariates $M_t$ that provide better forecasts ($\theta_\mathrm{M-TSFM}(X^H_t||M_t)$) relative to the unimodal baseline ($\theta_\mathrm{M-TSFM}(X^H_t)$).
}
    \label{fig:wordcloud}
\end{figure}

%% file: tables/noise_table.tex
\begin{table}[H]
    \centering
    \small
    \renewcommand{\arraystretch}{1.12}
    \resizebox{0.2\textwidth}{!}{%
    \begin{tabular}{lcccccccccccc}
    \toprule
    \textbf{$\sigma_\epsilon$}
     & CoCo MAE ($\downarrow$)
     \\
    \midrule
    0.00
     & 0.810 
     \\
    0.05
     & \textbf{0.801} 
     \\
    0.25
     & 0.802 
     \\
     0.50
     & 0.805
     \\
     1.00
     & 0.814
     \\
    \bottomrule
    \end{tabular}
    }%
    \vspace{0.1in}

    \caption{\textbf{Effect of noise standard deviation $\sigma_\epsilon$ on CoCo's forecasting performance.} $\sigma_\epsilon$ controls the noise added to the history portion of the forecast covariate $M_t$ (Eq.~\ref{eq:forecast_covariate}) and the confidence covariate $C_t$ (Eq.~\ref{eq:confidence_covariate}). Results are for 16-step forecasts with context length 64 on the Media Cloud dataset, using the base pretrained Chronos2 as the TSFM backbone. \textbf{Bold} denotes the best result. $\sigma_\epsilon = 0.05$ is selected and used in all experiments.}
    \label{tab:noise_level}
\end{table}

%% file: appendix/3_training_dataset.tex
\subsection{Training Datasets}\label{sec:training_datasets}

For training, as described in Section~\ref{sec:dataset_gen}, we annotate real-world time series datasets using a large language model (\textit{Qwen3-30B-A3B-Instruct-2507}) with the prompt detailed in Section~\ref{sec:synthetic_annotation_gen}. Specifically, we provide windows of 30 timestamps to the LLM for annotation in each batch for a dataset. Since we evaluate forecast horizons of up to 16 steps, this procedure yields annotations with varying degrees of predictive information. In particular, the first annotation within a window has access to the full future horizon and may therefore contain stronger predictive signals, whereas annotations closer to the end of the window are increasingly speculative and may carry little to no predictive power.

\input{figures_latex/synthetic_example}

We focus on the macroeconomic domain, as the relevant datasets are publicly available, abundant, and well-studied. Importantly, macroeconomic time series are known to be influenced by textual information from news and reports, making them a natural testbed for evaluating multimodal models that incorporate textual context. The primary sources of real-world time series used in our study are:
\begin{itemize}[leftmargin=18pt, itemsep=2pt, topsep=2pt]
    \item \textbf{FRED}: A public-domain repository of domestic and international economic indicators maintained by the Federal Reserve Bank of St.\ Louis. We select around 1200 datasets from the domains included in the FRED-MD collection, which is curated for empirical analysis in economics and finance. These datasets span multiple frequencies and economic domains. 
    \item \textbf{Other Sources}: We collect time series from two major European statistical agencies: Eurostat, the statistical office of the European Union; and Statistics Denmark (StatBank), Denmark's national statistical institute. From these sources, we collect time series datasets corresponding to various economic indicators, such as energy utility, unemployment rates, etc. These sources offer datasets at multiple frequencies (daily to monthly).
\end{itemize}

\input{tables/training_datasets}

A summary of the number of datasets generated from each source, with the frequency and the number of samples annotated, is provided in Table \ref{tab:training_datasets_summary}. Leveraging the synthetic annotation process, we were able to create a large-scale multimodal training dataset with more than 12 million samples, spanning over 36,968 target variables.

To mitigate memorization and unintended exposure to test sets, we enforce a strict date cutoff of December 2022 when creating the training dataset. While some existing multimodal datasets we use for evaluation contain samples from before this cutoff, the datasets used (summarized in Table~\ref{tab:eval_data_summary}) also consist of time series from diverse domains, which are substantially out-of-distribution relative to the datasets used during training. To further enforce a strict separation between training and evaluation and to explicitly address the knowledge cutoff issue of LLMs, we additionally curate a new multimodal evaluation suite consisting exclusively of data from 2024–2025. This dataset enables evaluation under both temporal shift and domain mismatch, providing a test of generalization, and is described in detail in Section~\ref{sec:eval_datasets}.

Apart from the synthetic dataset generated, we also include a subset of the FNSPID dataset during training to evaluate the usefulness of synthetic annotations when compared to real-world text for training (results in Table~\ref{tab:text_in_finetuning}). The top 150 stocks where the text availability is highest are chosen for training. Since FNSPID is also part of the evaluation set, we exclude the stocks in the evaluation set from the training set to prevent leakage during training.

Additionally, we use the Beijing Air Quality dataset \cite{beijing_multi-site_air_quality_501} as part of real-world multimodal datasets used along with our synthetically annotated multimodal datasets. The dataset contains air quality metrics and environmental variables recorded at 12 different stations. For each metric/variable, we construct textual annotations using the remaining features. The dataset has time series samples with hourly frequency.

%% file: figures_latex/synthetic_example.tex
\begin{figure*}[t]
    \centering
    \includegraphics[width=0.95\linewidth]{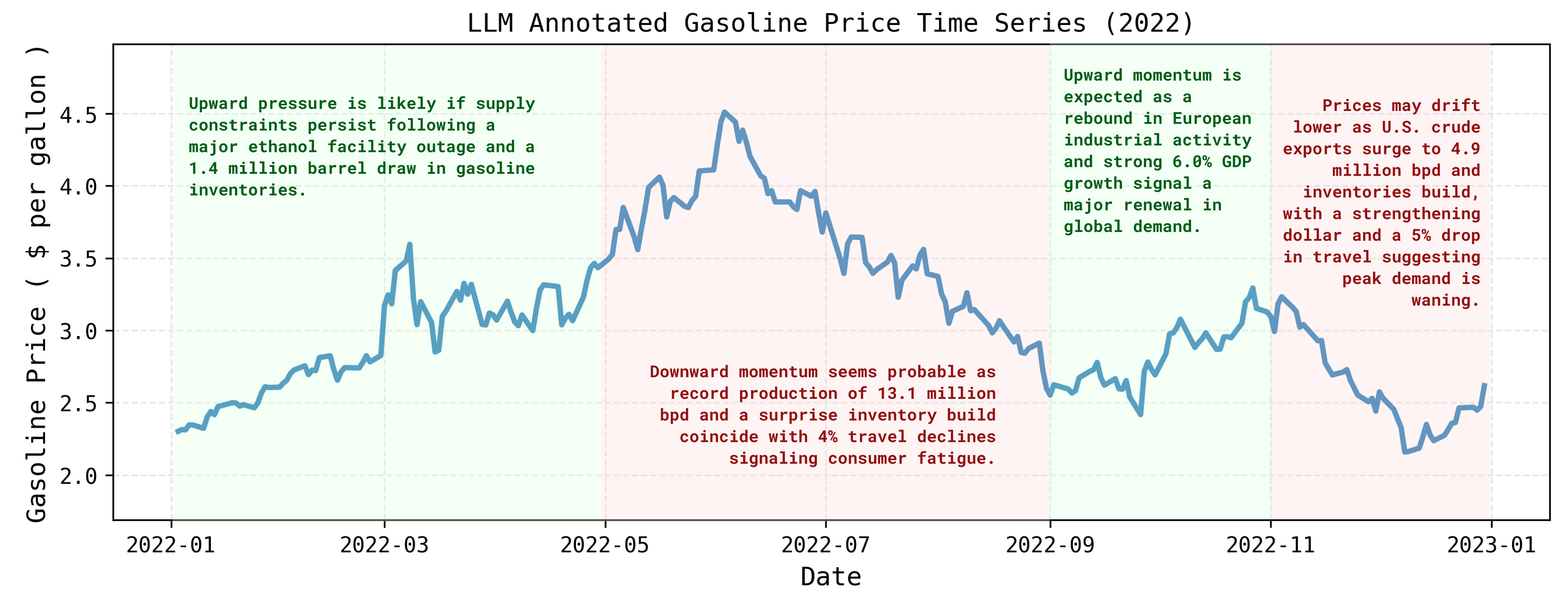}
    \caption{
        \textbf{Qwen3 provides realistic textual annotations.} Here, we show samples of annotations generated by \textit{Qwen3-30b-A3B-Instruct-2507} \cite{qwen3} for gasoline prices in the United States for the year 2022. Note that the text generated is relevant to the domain queried and foreshadows the trend in the numerical data. In the plot, the green portion indicates an increasing trend and the red portion indicates a decreasing trend.
    }
    \label{fig:synthetic_annotation_example}
\end{figure*}

%% file: tables/training_datasets.tex
\begin{table}[H]
\sc
\centering
\small
\renewcommand{\arraystretch}{1.15}

\resizebox{0.35\textwidth}{!}{%
\begin{tabular}{llrr}
\toprule
\textbf{Source} & \textbf{Frequency} & \textbf{Datasets} & \textbf{Samples} \\
\midrule
FRED & Daily & 95 & 107,993 \\
 & Weekly & 91 & 97,010 \\
 & Monthly & 807 & 433,750 \\
 & Quarterly & 93 & 18,891 \\
 & Annual & 52 & 3,176 \\
\midrule
EIA Data & Daily & 118 & 252,581 \\
 & Monthly & 1,596 & 428,449 \\
\midrule
DK Data & Monthly & 26,766 & 5,314,577 \\
\midrule
EU Data & Monthly & 7,001 & 1,471,900 \\
\midrule
Favorita & Daily & 87 & 148,469 \\
\midrule
FNSPID (Real) & Daily & 142 & 140,977 \\
\midrule
Air Quality & Hourly & 120 & 4,135,865 \\
\midrule
\textbf{Total} & & \textbf{36,968} & \textbf{12,553,638} \\
\bottomrule
\end{tabular}
}

\vspace{0.1in}
\caption{Summary of the training datasets used. Datasets from FNSPID contain real annotations}
\label{tab:training_datasets_summary}
\end{table}

%% file: appendix/4_evaluation_datasets.tex
\subsection{Evaluation Datasets}\label{sec:eval_datasets}

A comprehensive summary of the datasets used for evaluation is provided in Table \ref{tab:eval_data_summary}. Our study utilizes existing multimodal benchmarks, specifically the TimeMMD benchmark and stock prices from FNSPID. However, these datasets lack samples beyond 2023. Given that our methodology leverages the recent LLMs for forecasting, evaluating solely on legacy data introduces a risk of temporal data leakage, given that they have knowledge cutoffs of up to late 2024. To ensure a rigorous and ``fair'' evaluation in which the LLM has no prior exposure to test trends, we curated a new multimodal evaluation suite comprising datasets with samples following the 2024 cutoff. We show that CoCo's performance does not degrade when evaluated separately in samples after the cutoff date (Table \ref{tab:post_june_2024_zero_shot_multimodal}).

\textbf{TimeMMD} We evaluate on eight TimeMMD datasets  (spanning Agriculture, Climate, Economy, Energy, Security, Social Good and Traffic). While TimeMMD provides heterogeneous text sources (reports and search queries), we aggregate and summarize all available information for each timestamp to provide a unified context.

\textbf{FNSPID} We include stock price data and timestamped NASDAQ news for a subset of the S\&P 500's top 25 companies from different sectors. To handle the high volume of daily articles, we summarize multiple news entries into a single, daily textual summary.


\textbf{Media Cloud} For the curation of the text for the new evaluation datasets, we leveraged the Media Cloud API\footnote{https://search.mediacloud.org/search} for news articles with timestamps and used the pipeline shown in Figure \ref{fig:media_cloud} to obtain text annotations on a daily basis. The LLM prompts used in the pipeline are provided in \ref{mediacloud_prompts}. We also provide the Media Cloud queries used to fetch news articles from the API for each dataset.

\input{figures_latex/media_cloud}

Target variables were aggregated from three primary sources:
\begin{itemize}[leftmargin=18pt, itemsep=2pt, topsep=2pt]
    \item \textbf{Google Finance}\footnote{https://newsinitiative.withgoogle.com/resources/trainings/access-google-finance-data-in-sheets/}: Daily market pricing and equity data were retrieved using Google Finance, via the Google Sheets integration. This source provides broad accessibility to global exchange data. Specific tickers are provided to ensure exact dataset reconstruction.
    \item \textbf{FRED}: Macroeconomic indices and benchmarks were sourced from the FRED database, a public-domain repository for domestic and international economic indicators maintained by the Federal Reserve Bank of St. Louis.
    \item \textbf{Trading Economics}\footnote{https://tradingeconomics.com}: Commodity prices were obtained from the Trading Economics platform. This source provides curated global economic data. Researchers seeking to replicate this specific subset may require a platform subscription.
\end{itemize}

Below are the specific tickers and Media Cloud queries used for the curation:

\begin{itemize}[leftmargin=18pt, itemsep=2pt, topsep=2pt]
    \item \textbf{S\&P500 Index}
    \begin{itemize}
        \item Source : FRED - SP500\footnote{https://fred.stlouisfed.org/series/SP500}
        \item Media Cloud Query : \textit{``s\&p 500'' OR s\&p500}
    \end{itemize}

    \item \textbf{Crude Oil Price}
    \begin{itemize}
        \item Source : FRED - DCOILWTICO\footnote{https://fred.stlouisfed.org/series/DCOILWTICO}
        \item Media Cloud Query : \textit{``crude oil'' OR wti OR brent}
    \end{itemize}

    \item \textbf{NASDAQ Electricity Benchmark}
    \begin{itemize}
        \item Source : FRED - NASDAQNQUSB651010\footnote{https://fred.stlouisfed.org/series/NASDAQNQUSB651010}
        \item Media Cloud Query : \textit{nasdaq AND (electricity OR energy) AND sector}
    \end{itemize}

    \item \textbf{NVIDIA Stock Price}
    \begin{itemize}
        \item Source : Google Finance, Ticker: \textit{NASDAQ:NVDA}
        \item Media Cloud Query : \textit{nvidia AND analyst}
    \end{itemize}

    \item \textbf{SPDR Gold Shares Price}
    \begin{itemize}
        \item Source : Google Finance, Ticker: \textit{NYSEARCA:GLD}
        \item Media Cloud Query : \textit{gold AND analyst}
    \end{itemize}

     \item \textbf{Egg Prices}
    \begin{itemize}
        \item Source : Trading Economics - Eggs US\footnote{https://tradingeconomics.com/commodity/eggs-us}
        \item Media Cloud Query : \textit{egg AND (sales OR price)}
    \end{itemize}

    \item \textbf{Apple Stock Price}
    \begin{itemize}
        \item Source : Google Finance, Ticker: \textit{NASDAQ:AAPL}
        \item Media Cloud Query : \textit{apple AND tech AND analyst}
    \end{itemize}

    \item \textbf{Silver Price}
    \begin{itemize}
        \item Source : Trading Economics - Silver\footnote{https://tradingeconomics.com/commodity/silver}
        \item Media Cloud Query : \textit{silver AND (sales OR price)}
    \end{itemize}

    \item \textbf{Steel Rebar Futures Price}
    \begin{itemize}
        \item Source : Trading Economics - Steel Rebar Futures\footnote{https://tradingeconomics.com/commodity/steel}
        \item Media Cloud Query : \textit{steel AND (NOT amazon) AND (price OR sale OR export OR import)}
    \end{itemize}

    \item \textbf{Google Stock Price}
    \begin{itemize}
        \item Source : Google Finance, Ticker: \textit{NASDAQ:GOOG}
        \item Media Cloud Query : \textit{(google OR alphabet) AND (analyst)}
    \end{itemize}

    \item \textbf{Copper Futures Price}
    \begin{itemize}
        \item Source : Google Finance, Ticker: \textit{COMEX:HGW00}
        \item Media Cloud Query : \textit{copper AND (NOT amazon) AND (price OR sale OR export OR import)}
    \end{itemize}

\end{itemize}

\input{tables/evaluation_datasets}

%% file: figures_latex/media_cloud.tex
\begin{figure*}[ht]
    \centering
    \includegraphics[width=0.98\linewidth]{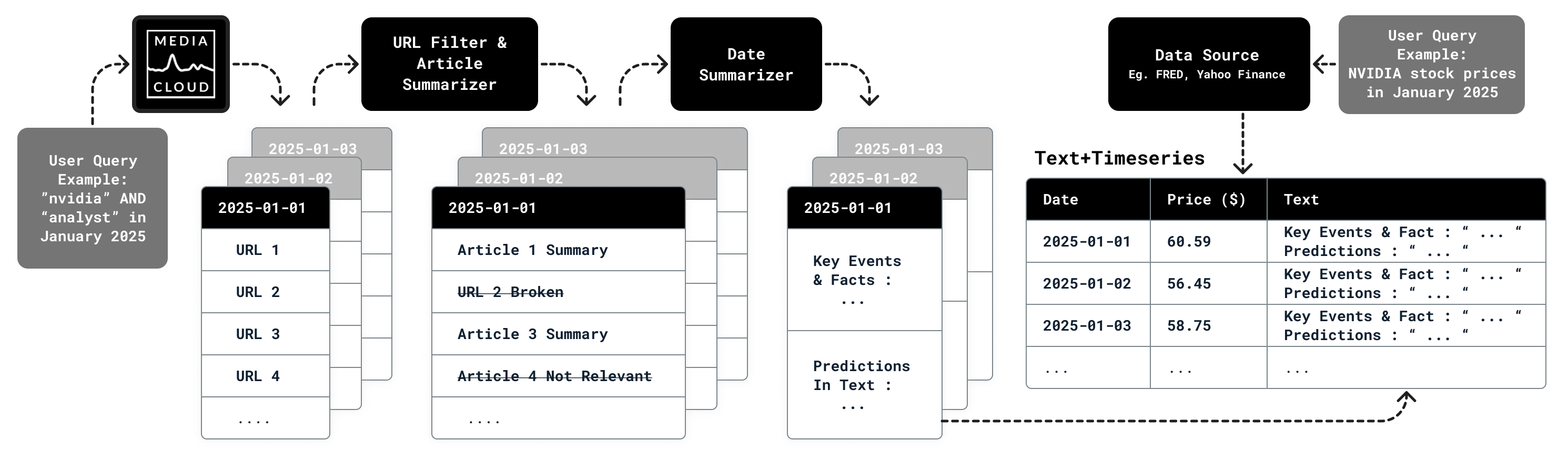}
    \caption{
        \textbf{Our approach to generate textual annotations, for real-world time series datasets, extracted from the news articles that are scraped from the internet.} Let us consider the task of generating textual annotations for the NVIDIA stock price dataset. First, we use the \textbf{Media Cloud} API to extract the relevant news articles between two successive timestamps $t-1$ and $t$. These news articles are summarized into \emph{Key Events and Facts} and \emph{Predictive Information} by an LLM. This summary is treated as the text annotation for the timestamp $t$.
    }
    \label{fig:media_cloud}
\end{figure*}

%% file: tables/evaluation_datasets.tex
\begin{table*}[t]
    \centering
    \small
    \renewcommand{\arraystretch}{1.12}
    \sc
    \resizebox{\textwidth}{!}{%
    \begin{tabular}{llccccc}
    \toprule
    \textbf{Text Source} & \textbf{Dataset} & \textbf{Target Source} & \textbf{Frequency} & \textbf{Date Range} & \textbf{Total} & \textbf{$\geq$2024-06} \\
    \midrule
    \textbf{Media Cloud} & Apple Stock Price & Google Finance & Daily & 2024-01 -- 2025-12 & 503 & 398 \\
     & Copper Futures Price & Google Finance & Daily & 2025-01 -- 2026-01 & 258 & 258 \\
     & Crude Oil Price & FRED & Daily & 2024-01 -- 2026-01 & 500 & 395 \\
     & Egg Price & Trading Economics & Daily & 2025-02 -- 2025-12 & 223 & 223 \\
     & Google Stock Price & Google Finance & Daily & 2024-01 -- 2025-12 & 502 & 398 \\
     & NASDAQ Electricity & FRED & Daily & 2024-01 -- 2025-12 & 523 & 413 \\
     & NVIDIA Stock Price & Google Finance & Daily & 2024-01 -- 2025-12 & 502 & 397 \\
     & S\&P500 Index & FRED & Daily & 2024-01 -- 2025-10 & 463 & 359 \\
     & SPDR Gold Shares Price & Google Finance & Daily & 2024-01 -- 2025-12 & 501 & 396 \\
     & Silver Price & Trading Economics & Daily & 2024-01 -- 2026-01 & 519 & 411 \\
     & Steel Rebar Futures Price & Trading Economics & Daily & 2025-01 -- 2026-01 & 241 & 241 \\
    \midrule
    \textbf{TimeMMD} & Broiler Price (Agriculture) & TimeMMD & Monthly & 1980-01 -- 2024-04 & 532 & 0 \\
     & Drought Level (Climate) & TimeMMD & Weekly & 2000-01 -- 2024-05 & 1,272 & 0 \\
     & US Trade Balance (Economy) & TimeMMD & Monthly & 1987-01 -- 2024-03 & 447 & 0 \\
     & Gasoline Price (Energy) & TimeMMD & Weekly & 1993-04 -- 2024-04 & 1,622 & 0 \\
     & Air Quality (Environment) & TimeMMD & Daily & 2019-09 -- 2023-09 & 1,480 & 0 \\
     & Disaster Grant (Security) & TimeMMD & Monthly & 1998-09 -- 2024-05 & 309 & 0 \\
     & Unemployment Rate (Social Good) & TimeMMD & Monthly & 1980-01 -- 2024-04 & 532 & 0 \\
     & Traffic Volume (Traffic) & TimeMMD & Monthly & 1979-12 -- 2024-03 & 532 & 0 \\
    \midrule
    \textbf{FNSPID} & Bank of America Stock Price & FNSPID & Daily & 2018-06 -- 2020-06 & 499 & 0 \\
     & Boeing Stock Price & FNSPID & Daily & 2021-12 -- 2023-12 & 500 & 0 \\
     & Costco Stock Price & FNSPID & Daily & 2021-12 -- 2023-12 & 500 & 0 \\
     & Exxon Mobil Stock Price & FNSPID & Daily & 2021-12 -- 2023-12 & 500 & 0 \\
     & Goldman Sachs Stock Price & FNSPID & Daily & 2021-12 -- 2023-12 & 500 & 0 \\
     & JPMC Stock Price & FNSPID & Daily & 2018-09 -- 2020-06 & 437 & 0 \\
     & McD Stock Price & FNSPID & Daily & 2018-06 -- 2020-06 & 500 & 0 \\
     & Microsoft Stock Price & FNSPID & Daily & 2022-04 -- 2023-12 & 412 & 0 \\
     & Oracle Stock Price & FNSPID & Daily & 2021-12 -- 2023-12 & 500 & 0 \\
     & Tesla Stock Price & FNSPID & Daily & 2022-05 -- 2023-12 & 410 & 0 \\
     & Walmart Stock Price & FNSPID & Daily & 2021-12 -- 2023-12 & 500 & 0 \\
    \midrule
    \textbf{Total} & & & & & \textbf{16,719} & \textbf{3,889} \\
    \bottomrule
    \end{tabular}
    }
    \caption{Summary of the multimodal evaluation datasets grouped by source.}
    \label{tab:eval_data_summary}
\end{table*}

%% file: appendix/rebuttal/2_baselines.tex
\subsection{Evaluation Baselines}
\label{app:baselines}

We organize the 27 baseline configurations according to whether they incorporate textual context and whether their parameters are fine-tuned. This categorization is also used in Figure~\ref{fig:rank_mae_plot}.

\begin{enumerate}
    \item \textbf{Zero-shot unimodal (8).}
    Chronos2 \citep{chronos2}, TimesFM2.5 \citep{timesfm}, TiRex \citep{auer2025tirex}, Moirai \citep{moirai}, Toto \citep{toto}, TabPFN2.5 \citep{grinsztajn2026tabpfn25advancingstateart}, TimeMoE \citep{shi2025timemoe}, and Panda \citep{lai2026panda}. These methods forecast directly from the numerical history without textual context or task-specific fine-tuning.

    \item \textbf{Zero-shot multimodal (7).}
    GPT-4o, Qwen3-30B, and TimeOmni directly forecast from the numerical history and paired text. Chronos2 MV and Toto MV receive Sentence Transformer embeddings of the text as additional covariates. We also include Chronos2+GPT-4o and TimesFM2.5+GPT-4o mean ensembles \citep{liu2026rethinking}, which average the corresponding TSFM and LLM forecasts.

    \item \textbf{Unimodal fine-tuning (6).}
    Chronos2, PatchTST \citep{patchtst}, iTransformer \citep{itransformer}, AutoTimes \citep{autotimes}, S2IP-LLM \citep{s2ipllm}, and TEST \citep{sun2024test} are fine-tuned using only the numerical histories from the synthetic training data.

    \item \textbf{Multimodal fine-tuning (6).}
    Chronos2 MV, AutoTimes, VoT \citep{vot}, TaTS \citep{li2026language}, S2IP-LLM, and TEST incorporate both numerical history and textual context during fine-tuning and inference. Where a method was originally unimodal, we adapt its input representation to incorporate text.
\end{enumerate}

Many fine-tuning baselines were originally designed for dataset-specific training. To ensure a controlled comparison, we instead train every fine-tuned configuration using the same synthetic text--time-series pairs used to train CoCo and evaluate it on the same held-out real-world datasets.

%% file: appendix/5_evaluation_metrics.tex
\subsection{Evaluation Metrics}\label{app:evaluation_metrics}

\textbf{Normalization}: As described in Appendix~\ref{app:training_description}, we apply z-score normalization during inference, and all reported MAE values are computed in the normalized space.

\textbf{MASE}: Along with the MAE, we also report Mean Absolute Scaled Error (MASE) (Table~\ref{tab:mase_setting_summary}) as a scale-free accuracy metric.
For a forecast $[\hat{x}_{t+1},\hat{x}_{t+2},\ldots,\hat{x}_{t+h}]$ and ground-truth $[x_{t+1}, x_{t+2}, \ldots , x_{t+h}]$ over horizon length $h$, we compute
\[
\mathrm{MASE}
=
\frac{\frac{1}{n}\sum_{t=1}^{n}\left|x_t-\hat{x}_t\right|}
{\frac{1}{n-1}\sum_{t=2}^{n}\left|x_t-x_{t-1}\right|}.
\]

The numerator is the model MAE over the forecast window, while the denominator is a naive
lag-1 scaling term computed on the same window.
Lower is better; values $<1$ indicate performance better than the naive scale baseline.
This follows the MASE usage in the \textbf{GIFT-Eval} benchmark for comparable, scale-normalized
time-series evaluation across heterogeneous datasets. The aggregate MASE over datasets is computed by taking the geometric mean across datasets as per convention.





\textbf{Wilcoxon Signed-Rank p-value}:
For each model pair, we form paired per-setting errors and run the two-sided Wilcoxon signed-rank test. The test ranks absolute paired differences and uses their signs, so it reflects both effect size and direction consistency. The null hypothesis is that the median paired difference is zero (no systematic difference). Small p-values indicate stronger evidence of a systematic difference while mixed positive and negative differences typically yield larger p-values.

\textbf{ELO Ratings}: We rank forecasting models across datasets, context lengths, and forecast horizons using an ELO rating system \cite{elo}, which estimates latent skill levels from relative outcomes. Each evaluation setting induces a multiplayer matchup among all models, converted into relative outcomes based on the MAE (lower values indicate better performance). Model ratings are updated after each matchup using the standard ELO update rule:
\begin{equation}
    R'_i = R_i + K \cdot (S_i - E_i),
\end{equation}
where $R_i$ and $R'_i$ denote the pre- and post-match ratings of model $i$, $S_i$ is the observed score derived from its rank in the matchup, and $E_i$ is the expected score computed from all participating models’ current ratings using a logistic model with deviation parameter $D=400$. All models are initialized with a base rating of 1500, and $K=32$ is the update factor controlling sensitivity.

The conventional ELO system is sensitive to the order of matchups and can become unstable as the number of models and evaluation settings increases \cite{elouncovered}. To mitigate this, we repeat the rating procedure across 32 random orderings of the matchups and report the mean rating for each model. This approach yields stable ratings at our scale of models and settings. This evaluation protocol is in the same spirit as rating-based leaderboards used for LLMs \cite{chatbotarena}, which employ ELO-style updates to aggregate performance comparisons across many evaluation settings rather than relying on simple averages of ranks or win counts.

%% file: appendix/rebuttal/1_stochasticity.tex
\section{CoCo's Robustness to Stochasticity}
\label{sec:stochasticity}

We evaluate the stability of CoCo across all 95 dataset--setting combinations under three sources of stochasticity: (1) LLM rollout sampling, (2) noise injection into the historical covariates, and (3) stochasticity from LLM fine-tuning checkpoints.

\textbf{LLM rollout stochasticity:} We generate LLM forecasts using three different random seeds, with five rollouts per seed at temperature $1.0$, resulting in 15 rollouts per sample. While individual rollout forecasts exhibit substantial variation, aggregating the rollouts using their median substantially reduces this variability. Across the three runs, the LLM rollout median has a coefficient of variation (CV) of only $0.56\%$, with the CV remaining below $2\%$ for most dataset--setting combinations. More importantly, this residual variation is effectively absorbed by the TSFM: CoCo's final forecast has a CV of only $0.01\%$. The corresponding aggregate MAE statistics are shown in Table~\ref{tab:stochasticity_rollouts}.

\begin{table}[t]
    \centering
    \small
    \renewcommand{\arraystretch}{1.12}
    \begin{tabular}{lc}
    \toprule
    \textbf{Forecast} & \textbf{MAE (mean $\pm$ std)} \\
    \midrule
    LLM Rollout ($\times 15$) & $0.7787 \pm 0.0489$ \\
    LLM Rollout Median ($\times 3$) & $0.7860 \pm 0.0044$ \\
    CoCo ($\times 3$) & $0.7194 \pm 0.0001$ \\
    \bottomrule
    \end{tabular}
    \vspace{0.1in}
    \caption{\textbf{CoCo's robustness to LLM rollout stochasticity.}
    Overall MAE across all 95 settings, evaluated across three random seeds. Although individual LLM rollouts exhibit substantial variation, taking the median across five rollouts substantially reduces the variability, and the subsequent CoCo forecast is highly stable.}
    \label{tab:stochasticity_rollouts}
\end{table}

\textbf{Robustness to covariate noise:} CoCo additionally injects Gaussian noise with $\sigma=0.05$ into the historical covariates during Chronos2 inference as a regularization mechanism. To isolate the effect of this stochasticity, we evaluate CoCo using five random seeds while keeping the LLM rollouts fixed. CoCo achieves an overall MAE of $0.7194 \pm 0.0005$, corresponding to a CV of only $0.06\%$. Thus, the small amount of noise introduced during inference does not materially affect the final forecast.

\textbf{Fine-tuning stochasticity:}
Finally, we evaluate the sensitivity of CoCo to stochasticity during LLM fine-tuning. We train the LoRA adapter for Qwen3-4B using three different training seeds, producing three independent checkpoints. At temperature $0$, the resulting LLM forecasts exhibit noticeable variation, with an MAE of $0.818 \pm 0.009$ (CV = $1.05\%$). In contrast, the corresponding CoCo forecasts remain highly consistent, achieving an MAE of $0.7194 \pm 0.0001$ (CV = $0.01\%$). This demonstrates that CoCo is not sensitive to the particular LLM fine-tuning checkpoint used.

Overall, CoCo remains highly robust across all three sources of stochasticity. We primarily attribute this stability to the use of the median across multiple LLM rollouts, which substantially suppresses variability before the resulting covariate is provided to the TSFM. The TSFM further absorbs the small amount of residual variation, resulting in a highly stable final forecast.

%% file: tables/iclr/dataset_mae_zero_shot_unimodal.tex
\begin{table*}[t]
\centering
\scriptsize
\setlength{\tabcolsep}{3.2pt}
\renewcommand{\arraystretch}{1.02}
\resizebox{\textwidth}{!}{%
\begin{tabular}{llrrrrrrrrr}
\toprule
 & & \multicolumn{9}{c}{\textbf{Mean MAE ($\downarrow$)}} \\
\cmidrule(lr){3-11}
\textbf{Group} & \textbf{Dataset} & \cellcolor{gray!15}{\textbf{CoCo}} & \textbf{TiRex1} & \textbf{Chronos2} & \textbf{Moirai} & \textbf{TimesFM2.5} & \textbf{Toto} & \textbf{TabPFN2.5} & \textbf{TimeMoE} & \textbf{PANDA} \\
\midrule
\textbf{Media Cloud} & Crude Oil Price & \cellcolor{gray!15}{0.700} & \underline{0.682} & 0.690 & 0.687 & \textbf{0.663} & 0.703 & 0.777 & 0.770 & 0.774 \\
 & S\&P500 Index & \cellcolor{gray!15}{\textbf{0.594}} & \underline{0.601} & 0.641 & 0.625 & 0.652 & 0.667 & 0.778 & 0.835 & 0.816 \\
 & Apple Stock & \cellcolor{gray!15}{\textbf{0.637}} & 0.666 & 0.666 & \underline{0.651} & 0.676 & 0.683 & 0.760 & 0.797 & 0.846 \\
 & Copper Futures & \cellcolor{gray!15}{\underline{0.682}} & 0.735 & 0.727 & \textbf{0.679} & 0.751 & 0.701 & 0.781 & 0.775 & 0.941 \\
 & Egg Price & \cellcolor{gray!15}{0.604} & 0.587 & \underline{0.583} & 0.607 & 0.587 & \textbf{0.583} & 0.655 & 0.598 & 0.865 \\
 & Electricity Price & \cellcolor{gray!15}{\textbf{0.576}} & 0.613 & 0.605 & 0.608 & \underline{0.592} & 0.625 & 0.710 & 0.707 & 0.751 \\
 & SPDR Gold Price & \cellcolor{gray!15}{\textbf{0.547}} & \underline{0.576} & 0.591 & 0.582 & 0.642 & 0.626 & 0.682 & 0.821 & 0.832 \\
 & Google Stock & \cellcolor{gray!15}{\underline{0.606}} & \textbf{0.600} & 0.624 & 0.627 & 0.636 & 0.640 & 0.659 & 0.809 & 0.790 \\
 & NVIDIA Stock Price & \cellcolor{gray!15}{\textbf{0.613}} & \underline{0.626} & 0.634 & 0.633 & 0.655 & 0.659 & 0.708 & 0.762 & 0.785 \\
 & Silver Price & \cellcolor{gray!15}{\textbf{0.699}} & \underline{0.725} & 0.753 & 0.737 & 0.758 & 0.738 & 0.899 & 0.926 & 0.964 \\
 & Steel Rebar Futures & \cellcolor{gray!15}{\underline{0.607}} & 0.626 & 0.633 & 0.640 & 0.642 & \textbf{0.592} & 0.685 & 0.667 & 0.748 \\
\midrule
\textbf{FNSPID} & Boeing Stock & \cellcolor{gray!15}{\textbf{0.585}} & \underline{0.587} & 0.598 & 0.609 & 0.611 & 0.617 & 0.622 & 0.689 & 0.728 \\
 & Bank of America Stock & \cellcolor{gray!15}{\textbf{0.820}} & \underline{0.826} & 0.850 & 0.843 & 0.831 & 0.871 & 0.890 & 0.941 & 1.005 \\
 & Costco Stock & \cellcolor{gray!15}{\textbf{0.567}} & \underline{0.569} & 0.583 & 0.604 & 0.584 & 0.602 & 0.672 & 0.700 & 0.728 \\
 & Goldman Sachs Stock & \cellcolor{gray!15}{0.743} & 0.741 & 0.743 & 0.767 & \underline{0.725} & 0.767 & 0.731 & \textbf{0.719} & 0.854 \\
 & JPMC Stock & \cellcolor{gray!15}{\textbf{0.799}} & \underline{0.804} & 0.837 & 0.816 & 0.849 & 0.861 & 0.876 & 1.040 & 1.008 \\
 & McDonald's Stock & \cellcolor{gray!15}{\underline{0.684}} & 0.684 & 0.712 & \textbf{0.681} & 0.736 & 0.712 & 0.742 & 0.806 & 0.857 \\
 & Microsoft Stock & \cellcolor{gray!15}{\textbf{0.596}} & \underline{0.610} & 0.646 & 0.625 & 0.645 & 0.654 & 0.712 & 0.801 & 0.830 \\
 & Oracle Stock & \cellcolor{gray!15}{\underline{0.594}} & \textbf{0.582} & 0.600 & 0.624 & 0.626 & 0.648 & 0.675 & 0.781 & 0.807 \\
 & Tesla Stock & \cellcolor{gray!15}{0.566} & 0.577 & 0.593 & \underline{0.562} & \textbf{0.559} & 0.610 & 0.587 & 0.634 & 0.709 \\
 & Walmart Stock & \cellcolor{gray!15}{\textbf{0.591}} & 0.616 & \underline{0.615} & 0.634 & 0.633 & 0.641 & 0.668 & 0.716 & 0.765 \\
 & Exxon Mobil Stock & \cellcolor{gray!15}{0.690} & \textbf{0.664} & \underline{0.684} & 0.706 & 0.730 & 0.712 & 0.772 & 0.731 & 0.766 \\
\midrule
\textbf{TimeMMD} & Broiler Price & \cellcolor{gray!15}{\textbf{0.653}} & \underline{0.674} & 0.705 & 0.682 & 0.735 & 0.712 & 0.809 & 0.853 & 0.934 \\
 & Drought Level & \cellcolor{gray!15}{\textbf{0.689}} & \underline{0.697} & 0.712 & 0.718 & 0.715 & 0.730 & 0.740 & 0.759 & 0.821 \\
 & US Trade Balance & \cellcolor{gray!15}{\textbf{0.650}} & 0.705 & 0.675 & 0.667 & \underline{0.655} & 0.716 & 0.727 & 0.813 & 0.844 \\
 & Gasoline Price & \cellcolor{gray!15}{\textbf{0.671}} & \underline{0.680} & 0.697 & 0.707 & 0.696 & 0.723 & 0.775 & 0.766 & 0.855 \\
 & Air Quality Index & \cellcolor{gray!15}{0.837} & \textbf{0.822} & 0.830 & 0.849 & \underline{0.826} & 0.834 & 0.849 & 0.856 & 0.897 \\
 & Disaster Grants & \cellcolor{gray!15}{\textbf{1.450}} & 1.577 & 1.547 & 1.539 & 1.580 & 1.559 & \underline{1.465} & 1.522 & 1.588 \\
 & Unemployment Rate & \cellcolor{gray!15}{0.550} & \textbf{0.531} & 0.565 & 0.596 & 0.555 & \underline{0.544} & 0.634 & 0.669 & 0.733 \\
 & Travel Volume & \cellcolor{gray!15}{\underline{0.264}} & 0.332 & 0.270 & 0.316 & \textbf{0.252} & 0.271 & 0.292 & 0.315 & 0.602 \\
\midrule
\textbf{TimeMMD Daily} & Air Quality Index & \cellcolor{gray!15}{0.974} & \textbf{0.928} & 0.945 & 0.957 & \underline{0.933} & 0.948 & 0.963 & 1.070 & 0.969 \\
\midrule
\textbf{TimeMMD Weekly} & Drought Level & \cellcolor{gray!15}{\textbf{1.040}} & 1.063 & 1.102 & \underline{1.062} & 1.079 & 1.081 & 1.111 & 1.186 & 1.496 \\
 & Gasoline Price & \cellcolor{gray!15}{\underline{1.041}} & \textbf{1.038} & 1.099 & 1.091 & 1.093 & 1.096 & 1.098 & 1.189 & 1.648 \\
\midrule
\textbf{TimeMMD Monthly} & Broiler Price & \cellcolor{gray!15}{\textbf{1.652}} & 1.682 & 1.681 & 1.712 & 1.707 & 1.774 & 1.715 & 1.752 & \underline{1.671} \\
 & US Trade Balance & \cellcolor{gray!15}{1.356} & 1.436 & 1.390 & 1.522 & 1.371 & \underline{1.343} & 1.443 & \textbf{1.335} & 1.450 \\
 & Disaster Grants & \cellcolor{gray!15}{\underline{1.610}} & \textbf{1.604} & 1.651 & 1.662 & 1.627 & 1.634 & 1.631 & 1.642 & 1.758 \\
 & Unemployment Rate & \cellcolor{gray!15}{\textbf{1.539}} & 1.588 & \underline{1.539} & 1.635 & 1.601 & 1.617 & 1.565 & 1.599 & 1.574 \\
 & Travel Volume & \cellcolor{gray!15}{1.419} & 1.476 & 1.485 & 1.387 & \underline{1.325} & 1.372 & 1.593 & \textbf{1.254} & 1.425 \\
\midrule
 & \textbf{Overall Wins} & \cellcolor{gray!15}{\textbf{38}} & 25 & 0 & 6 & 13 & 2 & 4 & 6 & 1 \\
 & \textbf{Mean} & \cellcolor{gray!15}{\textbf{0.719}} & 0.735 & 0.745 & 0.747 & 0.749 & 0.758 & 0.800 & 0.839 & 0.903 \\
 & \textbf{ELO} & \cellcolor{gray!15}{\textbf{1820}} & 1715 & 1602 & 1582 & 1593 & 1487 & 1344 & 1266 & 1090 \\
 & \textbf{Average Rank} & \cellcolor{gray!15}{\textbf{2.24}} & 2.98 & 4.07 & 4.31 & 4.17 & 5.22 & 6.52 & 7.19 & 8.31 \\
 & \textbf{Head-to-Head Wins} & \cellcolor{gray!15}{--} & 60/95 & 81/95 & 78/95 & 72/95 & 80/95 & 90/95 & 88/95 & 93/95 \\
 & \textbf{Wilcoxon $p$} & \cellcolor{gray!15}{--} & 5.0e-04 & 2.0e-12 & 9.2e-13 & 7.8e-09 & 8.1e-13 & 1.4e-16 & 3.3e-15 & 4.1e-17 \\
\bottomrule
\end{tabular}%
}
\caption{\textbf{Dataset-level MAE for zero-shot unimodal methods.} Each dataset reports the mean over the 64-16, 128-16, and 256-16 settings. TimeMMD Daily, Weekly, and Monthly retain their native settings as separate blocks. Summary statistics are computed across all 95 settings. CoCo outperforms multiple state-of-the-art TSFMs. Bold and underlined values denote the best and second-best method in each row.}
\label{tab:class_mae_pooled_95_zero_shot_unimodal}
\end{table*}

%% file: tables/iclr/dataset_mae_zero_shot_multimodal.tex
\begin{table*}[t]
\centering
\scriptsize
\setlength{\tabcolsep}{3.2pt}
\renewcommand{\arraystretch}{1.02}
\resizebox{\textwidth}{!}{%
\begin{tabular}{llrrrrrrrr}
\toprule
 & & \multicolumn{8}{c}{\textbf{Mean MAE ($\downarrow$)}} \\
\cmidrule(lr){3-10}
\textbf{Group} & \textbf{Dataset} & \cellcolor{gray!15}{\textbf{CoCo}} & \textbf{TimesFM2.5+GPT4o} & \textbf{Chronos2+GPT4o} & \textbf{Chronos2 MV} & \textbf{Toto MV} & \textbf{GPT-4o} & \textbf{Qwen3-30B} & \textbf{TimeOmni} \\
\midrule
\textbf{Media Cloud} & Crude Oil Price & \cellcolor{gray!15}{0.700} & \textbf{0.625} & \underline{0.645} & 0.707 & 0.646 & 0.713 & 0.842 & 0.887 \\
 & S\&P500 Index & \cellcolor{gray!15}{\textbf{0.594}} & \underline{0.634} & 0.649 & 0.693 & 0.724 & 0.723 & 0.655 & 0.785 \\
 & Apple Stock & \cellcolor{gray!15}{\textbf{0.637}} & \underline{0.682} & 0.692 & 0.715 & 0.706 & 0.770 & 0.754 & 0.876 \\
 & Copper Futures & \cellcolor{gray!15}{\textbf{0.682}} & \underline{0.685} & 0.685 & 0.748 & 0.740 & 0.725 & 0.736 & 0.853 \\
 & Egg Price & \cellcolor{gray!15}{0.604} & 0.579 & \textbf{0.572} & 0.594 & \underline{0.577} & 0.599 & 0.672 & 0.809 \\
 & Electricity Price & \cellcolor{gray!15}{\textbf{0.576}} & \underline{0.628} & 0.641 & 0.631 & 0.646 & 0.719 & 0.668 & 0.851 \\
 & SPDR Gold Price & \cellcolor{gray!15}{\underline{0.547}} & 0.568 & \textbf{0.546} & 0.660 & 0.676 & 0.582 & 0.650 & 3.468 \\
 & Google Stock & \cellcolor{gray!15}{\textbf{0.606}} & 0.619 & \underline{0.619} & 0.669 & 0.668 & 0.680 & 0.772 & 0.726 \\
 & NVIDIA Stock Price & \cellcolor{gray!15}{\textbf{0.613}} & \underline{0.658} & 0.658 & 0.683 & 0.688 & 0.758 & 0.885 & 0.734 \\
 & Silver Price & \cellcolor{gray!15}{\textbf{0.699}} & \underline{0.706} & 0.714 & 0.830 & 0.823 & 0.784 & 0.781 & 0.891 \\
 & Steel Rebar Futures & \cellcolor{gray!15}{0.607} & 0.603 & 0.610 & \underline{0.584} & \textbf{0.574} & 0.695 & 0.787 & 1.375 \\
\midrule
\textbf{FNSPID} & Boeing Stock & \cellcolor{gray!15}{\textbf{0.585}} & \underline{0.588} & 0.591 & 0.597 & 0.620 & 0.620 & 0.737 & 0.764 \\
 & Bank of America Stock & \cellcolor{gray!15}{0.820} & \textbf{0.802} & \underline{0.820} & 0.860 & 0.870 & 0.862 & 0.889 & 1.006 \\
 & Costco Stock & \cellcolor{gray!15}{\textbf{0.567}} & \underline{0.567} & 0.577 & 0.621 & 0.629 & 0.627 & 0.632 & 0.713 \\
 & Goldman Sachs Stock & \cellcolor{gray!15}{0.743} & 0.756 & 0.774 & \textbf{0.689} & \underline{0.733} & 0.866 & 0.834 & 0.879 \\
 & JPMC Stock & \cellcolor{gray!15}{0.799} & \textbf{0.763} & \underline{0.768} & 0.865 & 0.862 & 0.780 & 0.840 & 1.077 \\
 & McDonald's Stock & \cellcolor{gray!15}{0.684} & \textbf{0.645} & \underline{0.670} & 0.691 & 0.745 & 0.695 & 0.878 & 0.875 \\
 & Microsoft Stock & \cellcolor{gray!15}{\textbf{0.596}} & \underline{0.607} & 0.615 & 0.678 & 0.694 & 0.632 & 0.626 & 0.662 \\
 & Oracle Stock & \cellcolor{gray!15}{\textbf{0.594}} & 0.635 & \underline{0.634} & 0.651 & 0.658 & 0.714 & 0.689 & 0.744 \\
 & Tesla Stock & \cellcolor{gray!15}{0.566} & \underline{0.565} & 0.583 & \textbf{0.552} & 0.585 & 0.625 & 0.635 & 0.628 \\
 & Walmart Stock & \cellcolor{gray!15}{\textbf{0.591}} & 0.642 & 0.638 & \underline{0.634} & 0.639 & 0.717 & 0.663 & 0.722 \\
 & Exxon Mobil Stock & \cellcolor{gray!15}{0.690} & \underline{0.687} & \textbf{0.679} & 0.718 & 0.725 & 0.741 & 0.795 & 0.826 \\
\midrule
\textbf{TimeMMD} & Broiler Price & \cellcolor{gray!15}{\textbf{0.653}} & 0.669 & \underline{0.663} & 0.786 & 0.753 & 0.725 & 0.779 & 1.394 \\
 & Drought Level & \cellcolor{gray!15}{\textbf{0.689}} & \underline{0.699} & 0.705 & 0.721 & 0.716 & 0.770 & 0.770 & 0.841 \\
 & US Trade Balance & \cellcolor{gray!15}{\textbf{0.650}} & 0.697 & 0.696 & 0.703 & \underline{0.662} & 0.900 & 1.153 & 1.436 \\
 & Gasoline Price & \cellcolor{gray!15}{0.671} & \textbf{0.667} & \underline{0.671} & 0.714 & 0.701 & 0.710 & 0.683 & 0.828 \\
 & Air Quality Index & \cellcolor{gray!15}{0.837} & 0.917 & 0.921 & \textbf{0.833} & \underline{0.833} & 1.126 & 1.158 & 1.146 \\
 & Disaster Grants & \cellcolor{gray!15}{\textbf{1.450}} & 1.640 & 1.629 & \underline{1.552} & 1.567 & 1.847 & 1.764 & 2.908 \\
 & Unemployment Rate & \cellcolor{gray!15}{0.550} & \textbf{0.461} & \underline{0.463} & 0.571 & 0.546 & 0.522 & 0.637 & 0.786 \\
 & Travel Volume & \cellcolor{gray!15}{\textbf{0.264}} & 0.434 & 0.430 & 0.271 & \underline{0.271} & 0.690 & 0.812 & 21.457 \\
\midrule
\textbf{TimeMMD Daily} & Air Quality Index & \cellcolor{gray!15}{0.974} & 1.516 & 1.535 & \textbf{0.943} & \underline{0.947} & 2.506 & 1.910 & 1.275 \\
\midrule
\textbf{TimeMMD Weekly} & Drought Level & \cellcolor{gray!15}{\underline{1.040}} & \textbf{1.035} & 1.055 & 1.112 & 1.081 & 1.101 & 1.100 & 1.225 \\
 & Gasoline Price & \cellcolor{gray!15}{1.041} & \textbf{1.009} & \underline{1.021} & 1.094 & 1.133 & 1.042 & 1.032 & 1.237 \\
\midrule
\textbf{TimeMMD Monthly} & Broiler Price & \cellcolor{gray!15}{\textbf{1.652}} & 1.670 & 1.697 & \underline{1.657} & 1.772 & 1.797 & 2.214 & 2.170 \\
 & US Trade Balance & \cellcolor{gray!15}{1.356} & 1.399 & 1.423 & \underline{1.347} & \textbf{1.316} & 1.515 & 1.700 & 2.373 \\
 & Disaster Grants & \cellcolor{gray!15}{\textbf{1.610}} & 1.663 & 1.674 & 1.639 & \underline{1.611} & 1.767 & 1.773 & 1.981 \\
 & Unemployment Rate & \cellcolor{gray!15}{1.539} & \textbf{1.400} & 1.423 & 1.536 & 1.579 & \underline{1.422} & 1.610 & 1.691 \\
 & Travel Volume & \cellcolor{gray!15}{1.419} & \underline{1.386} & 1.466 & 1.403 & \textbf{1.344} & 1.538 & 1.712 & 2.031 \\
\midrule
 & \textbf{Overall Wins} & \cellcolor{gray!15}{\textbf{41}} & 25 & 7 & 13 & 9 & 0 & 0 & 0 \\
 & \textbf{Mean} & \cellcolor{gray!15}{\textbf{0.719}} & 0.742 & 0.748 & 0.763 & 0.766 & 0.836 & 0.878 & 1.756 \\
 & \textbf{ELO} & \cellcolor{gray!15}{\textbf{1795}} & 1744 & 1676 & 1549 & 1528 & 1365 & 1289 & 1053 \\
 & \textbf{Average Rank} & \cellcolor{gray!15}{\textbf{2.23}} & 2.64 & 3.14 & 4.17 & 4.34 & 5.65 & 6.25 & 7.58 \\
 & \textbf{Head-to-Head Wins} & \cellcolor{gray!15}{--} & 60/95 & 65/95 & 74/95 & 75/95 & 85/95 & 94/95 & 95/95 \\
 & \textbf{Wilcoxon $p$} & \cellcolor{gray!15}{--} & 0.005 & 6.3e-05 & 1.8e-10 & 6.1e-10 & 8.9e-15 & 2.8e-17 & 2.6e-17 \\
\bottomrule
\end{tabular}%
}
\caption{\textbf{Dataset-level MAE for zero-shot multimodal methods.} Each dataset reports the mean over the 64-16, 128-16, and 256-16 settings. TimeMMD Daily, Weekly, and Monthly retain their native settings as separate blocks. Summary statistics are computed across all 95 settings. Mean ensembles are included because they combine zero-shot forecasts. CoCo outperforms state-of-the-art LLM forecasters as well as multivariate TSFMs that directly incorporate Sentence Transformer text embeddings as covariates. Bold and underlined values denote the best and second-best method in each row.}
\label{tab:class_mae_pooled_95_zero_shot_multimodal}
\end{table*}

%% file: tables/iclr/dataset_mae_unimodal_finetuning.tex
\begin{table*}[t]
\centering
\scriptsize
\setlength{\tabcolsep}{3.2pt}
\renewcommand{\arraystretch}{1.02}
\resizebox{\textwidth}{!}{%
\begin{tabular}{llrrrrrrr}
\toprule
 & & \multicolumn{7}{c}{\textbf{Mean MAE ($\downarrow$)}} \\
\cmidrule(lr){3-9}
\textbf{Group} & \textbf{Dataset} & \cellcolor{gray!15}{\textbf{CoCo}} & \textbf{Chronos2 (FT)} & \textbf{AutoTimes} & \textbf{PatchTST} & \textbf{S2IP-LLM} & \textbf{iTransformer} & \textbf{TEST} \\
\midrule
\textbf{Media Cloud} & Crude Oil Price & \cellcolor{gray!15}{0.700} & \textbf{0.681} & \underline{0.691} & 0.707 & 0.723 & 0.763 & 0.769 \\
 & S\&P500 Index & \cellcolor{gray!15}{\textbf{0.594}} & 0.660 & 0.652 & \underline{0.626} & 0.661 & 0.740 & 0.764 \\
 & Apple Stock & \cellcolor{gray!15}{\textbf{0.637}} & 0.683 & \underline{0.669} & 0.683 & 0.698 & 0.772 & 0.827 \\
 & Copper Futures & \cellcolor{gray!15}{\textbf{0.682}} & 0.729 & 0.731 & \underline{0.717} & 0.776 & 0.779 & 0.767 \\
 & Egg Price & \cellcolor{gray!15}{\textbf{0.604}} & \underline{0.623} & 0.705 & 0.698 & 0.774 & 0.765 & 0.805 \\
 & Electricity Price & \cellcolor{gray!15}{\underline{0.576}} & \textbf{0.560} & 0.603 & 0.603 & 0.614 & 0.627 & 0.667 \\
 & SPDR Gold Price & \cellcolor{gray!15}{\textbf{0.547}} & \underline{0.577} & 0.606 & 0.587 & 0.600 & 0.633 & 0.680 \\
 & Google Stock & \cellcolor{gray!15}{\textbf{0.606}} & \underline{0.643} & 0.663 & 0.651 & 0.685 & 0.743 & 0.789 \\
 & NVIDIA Stock Price & \cellcolor{gray!15}{\textbf{0.613}} & 0.659 & 0.667 & \underline{0.651} & 0.683 & 0.738 & 0.759 \\
 & Silver Price & \cellcolor{gray!15}{\textbf{0.699}} & 0.722 & 0.762 & \underline{0.719} & 0.723 & 0.764 & 0.809 \\
 & Steel Rebar Futures & \cellcolor{gray!15}{0.607} & \underline{0.602} & \textbf{0.550} & 0.658 & 0.652 & 0.679 & 0.707 \\
\midrule
\textbf{FNSPID} & Boeing Stock & \cellcolor{gray!15}{\textbf{0.585}} & \underline{0.605} & 0.613 & 0.618 & 0.618 & 0.758 & 0.716 \\
 & Bank of America Stock & \cellcolor{gray!15}{\textbf{0.820}} & \underline{0.848} & 0.855 & 0.862 & 0.871 & 0.938 & 0.998 \\
 & Costco Stock & \cellcolor{gray!15}{\textbf{0.567}} & 0.593 & 0.595 & \underline{0.576} & 0.597 & 0.651 & 0.629 \\
 & Goldman Sachs Stock & \cellcolor{gray!15}{0.743} & \underline{0.737} & \textbf{0.713} & 0.763 & 0.764 & 0.819 & 0.824 \\
 & JPMC Stock & \cellcolor{gray!15}{\textbf{0.799}} & \underline{0.812} & 0.827 & 0.815 & 0.849 & 0.891 & 1.000 \\
 & McDonald's Stock & \cellcolor{gray!15}{\textbf{0.684}} & 0.709 & \underline{0.700} & 0.739 & 0.718 & 0.852 & 0.809 \\
 & Microsoft Stock & \cellcolor{gray!15}{\textbf{0.596}} & \underline{0.613} & 0.638 & 0.616 & 0.620 & 0.685 & 0.761 \\
 & Oracle Stock & \cellcolor{gray!15}{\underline{0.594}} & \textbf{0.589} & 0.619 & 0.622 & 0.604 & 0.697 & 0.690 \\
 & Tesla Stock & \cellcolor{gray!15}{\textbf{0.566}} & 0.572 & \underline{0.571} & 0.598 & 0.608 & 0.681 & 0.722 \\
 & Walmart Stock & \cellcolor{gray!15}{\textbf{0.591}} & 0.631 & \underline{0.615} & 0.634 & 0.653 & 0.699 & 0.762 \\
 & Exxon Mobil Stock & \cellcolor{gray!15}{\underline{0.690}} & 0.736 & \textbf{0.670} & 0.701 & 0.706 & 0.752 & 0.815 \\
\midrule
\textbf{TimeMMD} & Broiler Price & \cellcolor{gray!15}{\textbf{0.653}} & \underline{0.668} & 0.685 & 0.692 & 0.691 & 0.729 & 0.800 \\
 & Drought Level & \cellcolor{gray!15}{\textbf{0.689}} & 0.725 & \underline{0.709} & 0.743 & 0.775 & 0.818 & 0.868 \\
 & US Trade Balance & \cellcolor{gray!15}{\underline{0.650}} & \textbf{0.620} & 0.813 & 0.709 & 0.693 & 0.678 & 0.796 \\
 & Gasoline Price & \cellcolor{gray!15}{\textbf{0.671}} & \underline{0.696} & 0.734 & 0.732 & 0.757 & 0.807 & 0.840 \\
 & Air Quality Index & \cellcolor{gray!15}{\underline{0.837}} & \textbf{0.831} & 0.869 & 0.922 & 0.893 & 0.930 & 0.985 \\
 & Disaster Grants & \cellcolor{gray!15}{\textbf{1.450}} & \underline{1.462} & 1.525 & 1.538 & 1.488 & 1.537 & 1.546 \\
 & Unemployment Rate & \cellcolor{gray!15}{\underline{0.550}} & \textbf{0.501} & 0.740 & 0.689 & 0.647 & 0.810 & 0.794 \\
 & Travel Volume & \cellcolor{gray!15}{\underline{0.264}} & \textbf{0.244} & 0.478 & 0.474 & 0.444 & 0.466 & 0.747 \\
\midrule
\textbf{TimeMMD Daily} & Air Quality Index & \cellcolor{gray!15}{0.974} & \textbf{0.935} & \underline{0.971} & 1.024 & 0.990 & 1.058 & 1.124 \\
\midrule
\textbf{TimeMMD Weekly} & Drought Level & \cellcolor{gray!15}{\underline{1.040}} & 1.072 & \textbf{1.035} & 1.061 & 1.091 & 1.113 & 1.239 \\
 & Gasoline Price & \cellcolor{gray!15}{\textbf{1.041}} & \underline{1.065} & 1.111 & 1.089 & 1.173 & 1.120 & 1.315 \\
\midrule
\textbf{TimeMMD Monthly} & Broiler Price & \cellcolor{gray!15}{\textbf{1.652}} & 1.725 & 1.799 & 1.685 & 1.767 & \underline{1.684} & 2.144 \\
 & US Trade Balance & \cellcolor{gray!15}{1.356} & 1.478 & \textbf{1.294} & \underline{1.300} & 1.343 & 1.316 & 1.589 \\
 & Disaster Grants & \cellcolor{gray!15}{\textbf{1.610}} & 1.621 & 1.636 & 1.616 & 1.630 & \underline{1.613} & 1.980 \\
 & Unemployment Rate & \cellcolor{gray!15}{\underline{1.539}} & \textbf{1.539} & 1.619 & 1.551 & 1.622 & 1.542 & 1.663 \\
 & Travel Volume & \cellcolor{gray!15}{1.419} & 1.680 & \textbf{1.230} & 1.329 & \underline{1.286} & 1.387 & 1.665 \\
\midrule
 & \textbf{Overall Wins} & \cellcolor{gray!15}{\textbf{52}} & 23 & 16 & 1 & 3 & 0 & 0 \\
 & \textbf{Mean} & \cellcolor{gray!15}{\textbf{0.719}} & 0.738 & 0.763 & 0.764 & 0.773 & 0.823 & 0.882 \\
 & \textbf{ELO} & \cellcolor{gray!15}{\textbf{1860}} & 1697 & 1595 & 1549 & 1472 & 1249 & 1078 \\
 & \textbf{Average Rank} & \cellcolor{gray!15}{\textbf{1.69}} & 2.63 & 3.40 & 3.74 & 4.25 & 5.71 & 6.58 \\
 & \textbf{Head-to-Head Wins} & \cellcolor{gray!15}{--} & 64/95 & 78/95 & 89/95 & 87/95 & 92/95 & 94/95 \\
 & \textbf{Wilcoxon $p$} & \cellcolor{gray!15}{--} & 1.2e-05 & 3.2e-10 & 7.2e-15 & 3.5e-15 & 6.1e-17 & 2.7e-17 \\
\bottomrule
\end{tabular}%
}
\caption{\textbf{Dataset-level MAE for unimodal fine-tuning methods.} Each dataset reports the mean over the 64-16, 128-16, and 256-16 settings. TimeMMD Daily, Weekly, and Monthly retain their native settings as separate blocks. Summary statistics are computed across all 95 settings. CoCo outperforms unimodal fine-tuning approaches, indicating that exogenous textual context is essential for improved forecasting and is difficult to recover from numerical history alone. Bold and underlined values denote the best and second-best method in each row.}
\label{tab:class_mae_pooled_95_unimodal_finetuning}
\end{table*}

%% file: tables/iclr/dataset_mae_multimodal_finetuning.tex
\begin{table*}[t]
\centering
\scriptsize
\setlength{\tabcolsep}{3.2pt}
\renewcommand{\arraystretch}{1.02}
\resizebox{\textwidth}{!}{%
\begin{tabular}{llrrrrrrr}
\toprule
 & & \multicolumn{7}{c}{\textbf{Mean MAE ($\downarrow$)}} \\
\cmidrule(lr){3-9}
\textbf{Group} & \textbf{Dataset} & \cellcolor{gray!15}{\textbf{CoCo}} & \textbf{Chronos2 MV (FT)} & \textbf{VoT} & \textbf{TaTS} & \textbf{S2IP-LLM} & \textbf{AutoTimes} & \textbf{TEST} \\
\midrule
\textbf{Media Cloud} & Crude Oil Price & \cellcolor{gray!15}{0.700} & \underline{0.680} & \textbf{0.667} & 0.698 & 0.715 & 0.699 & 0.704 \\
 & S\&P500 Index & \cellcolor{gray!15}{\textbf{0.594}} & \underline{0.641} & 0.679 & 0.694 & 0.666 & 0.682 & 0.727 \\
 & Apple Stock & \cellcolor{gray!15}{\textbf{0.637}} & \underline{0.657} & 0.661 & 0.702 & 0.698 & 0.730 & 0.699 \\
 & Copper Futures & \cellcolor{gray!15}{\underline{0.682}} & \textbf{0.679} & 0.707 & 0.730 & 0.755 & 0.755 & 0.732 \\
 & Egg Price & \cellcolor{gray!15}{\underline{0.604}} & 0.667 & 0.709 & \textbf{0.592} & 0.737 & 0.781 & 0.749 \\
 & Electricity Price & \cellcolor{gray!15}{\textbf{0.576}} & \underline{0.587} & 0.596 & 0.623 & 0.601 & 0.617 & 0.622 \\
 & SPDR Gold Price & \cellcolor{gray!15}{\textbf{0.547}} & \underline{0.561} & 0.642 & 0.646 & 0.603 & 0.661 & 0.660 \\
 & Google Stock & \cellcolor{gray!15}{\textbf{0.606}} & \underline{0.622} & 0.641 & 0.668 & 0.711 & 0.707 & 0.671 \\
 & NVIDIA Stock Price & \cellcolor{gray!15}{\textbf{0.613}} & \underline{0.656} & 0.662 & 0.672 & 0.670 & 0.679 & 0.682 \\
 & Silver Price & \cellcolor{gray!15}{\textbf{0.699}} & \underline{0.726} & 0.775 & 0.810 & 0.746 & 0.828 & 0.798 \\
 & Steel Rebar Futures & \cellcolor{gray!15}{0.607} & 0.608 & 0.585 & \textbf{0.567} & 0.583 & \underline{0.567} & 0.608 \\
\midrule
\textbf{FNSPID} & Boeing Stock & \cellcolor{gray!15}{\textbf{0.585}} & 0.617 & \underline{0.613} & 0.631 & 0.665 & 0.665 & 0.704 \\
 & Bank of America Stock & \cellcolor{gray!15}{\underline{0.820}} & \textbf{0.806} & 0.856 & 0.843 & 0.864 & 0.915 & 0.928 \\
 & Costco Stock & \cellcolor{gray!15}{\underline{0.567}} & \textbf{0.566} & 0.588 & 0.613 & 0.614 & 0.620 & 0.670 \\
 & Goldman Sachs Stock & \cellcolor{gray!15}{0.743} & 0.741 & \underline{0.725} & \textbf{0.684} & 0.765 & 0.727 & 0.793 \\
 & JPMC Stock & \cellcolor{gray!15}{\underline{0.799}} & \textbf{0.796} & 0.813 & 0.841 & 0.820 & 0.892 & 0.895 \\
 & McDonald's Stock & \cellcolor{gray!15}{\underline{0.684}} & \textbf{0.663} & 0.685 & 0.695 & 0.703 & 0.728 & 0.737 \\
 & Microsoft Stock & \cellcolor{gray!15}{\textbf{0.596}} & \underline{0.608} & 0.623 & 0.669 & 0.638 & 0.679 & 0.683 \\
 & Oracle Stock & \cellcolor{gray!15}{\underline{0.594}} & \textbf{0.592} & 0.612 & 0.634 & 0.644 & 0.671 & 0.669 \\
 & Tesla Stock & \cellcolor{gray!15}{0.566} & 0.586 & \underline{0.565} & \textbf{0.552} & 0.664 & 0.595 & 0.676 \\
 & Walmart Stock & \cellcolor{gray!15}{\textbf{0.591}} & 0.617 & \underline{0.606} & 0.627 & 0.634 & 0.637 & 0.670 \\
 & Exxon Mobil Stock & \cellcolor{gray!15}{0.690} & 0.708 & \textbf{0.662} & 0.711 & 0.704 & \underline{0.686} & 0.713 \\
\midrule
\textbf{TimeMMD} & Broiler Price & \cellcolor{gray!15}{\textbf{0.653}} & \underline{0.670} & 0.676 & 0.777 & 0.672 & 0.709 & 0.696 \\
 & Drought Level & \cellcolor{gray!15}{\textbf{0.689}} & 0.723 & \underline{0.710} & 0.722 & 0.740 & 0.773 & 0.745 \\
 & US Trade Balance & \cellcolor{gray!15}{0.650} & \underline{0.637} & 0.728 & \textbf{0.630} & 0.732 & 0.918 & 0.860 \\
 & Gasoline Price & \cellcolor{gray!15}{\textbf{0.671}} & 0.711 & 0.732 & \underline{0.708} & 0.739 & 0.754 & 0.763 \\
 & Air Quality Index & \cellcolor{gray!15}{\textbf{0.837}} & \underline{0.842} & 0.909 & 0.859 & 0.901 & 0.867 & 0.879 \\
 & Disaster Grants & \cellcolor{gray!15}{\textbf{1.450}} & 1.482 & 1.480 & 1.511 & 1.467 & \underline{1.450} & 1.501 \\
 & Unemployment Rate & \cellcolor{gray!15}{\textbf{0.550}} & 0.586 & 0.715 & \underline{0.552} & 0.727 & 0.794 & 0.765 \\
 & Travel Volume & \cellcolor{gray!15}{\underline{0.264}} & 0.303 & 0.406 & \textbf{0.262} & 0.489 & 0.663 & 0.802 \\
\midrule
\textbf{TimeMMD Daily} & Air Quality Index & \cellcolor{gray!15}{0.974} & \textbf{0.941} & 1.013 & 0.961 & 0.999 & \underline{0.960} & 1.010 \\
\midrule
\textbf{TimeMMD Weekly} & Drought Level & \cellcolor{gray!15}{\underline{1.040}} & \textbf{1.027} & 1.065 & 1.071 & 1.067 & 1.111 & 1.104 \\
 & Gasoline Price & \cellcolor{gray!15}{\textbf{1.041}} & 1.086 & 1.102 & \underline{1.066} & 1.149 & 1.193 & 1.177 \\
\midrule
\textbf{TimeMMD Monthly} & Broiler Price & \cellcolor{gray!15}{\textbf{1.652}} & 1.751 & \underline{1.694} & 1.703 & 1.734 & 1.780 & 1.710 \\
 & US Trade Balance & \cellcolor{gray!15}{1.356} & \textbf{1.293} & 1.346 & 1.350 & 1.307 & \underline{1.294} & 1.436 \\
 & Disaster Grants & \cellcolor{gray!15}{\textbf{1.610}} & \underline{1.614} & 1.629 & 1.641 & 1.623 & 1.633 & 1.707 \\
 & Unemployment Rate & \cellcolor{gray!15}{1.539} & \textbf{1.514} & 1.586 & \underline{1.520} & 1.595 & 1.621 & 1.650 \\
 & Travel Volume & \cellcolor{gray!15}{1.419} & 1.446 & 1.275 & 1.433 & \underline{1.229} & \textbf{1.205} & 1.325 \\
\midrule
 & \textbf{Overall Wins} & \cellcolor{gray!15}{\textbf{47}} & 19 & 11 & 13 & 1 & 4 & 0 \\
 & \textbf{Mean} & \cellcolor{gray!15}{\textbf{0.719}} & 0.734 & 0.756 & 0.754 & 0.775 & 0.801 & 0.815 \\
 & \textbf{ELO} & \cellcolor{gray!15}{\textbf{1761}} & 1642 & 1572 & 1499 & 1438 & 1325 & 1263 \\
 & \textbf{Average Rank} & \cellcolor{gray!15}{\textbf{2.07}} & 2.80 & 3.45 & 3.91 & 4.55 & 5.38 & 5.84 \\
 & \textbf{Head-to-Head Wins} & \cellcolor{gray!15}{--} & 68/95 & 71/95 & 70/95 & 85/95 & 83/95 & 91/95 \\
 & \textbf{Wilcoxon $p$} & \cellcolor{gray!15}{--} & 1.8e-05 & 1.5e-08 & 1.3e-08 & 1.3e-13 & 2.3e-13 & 2.7e-16 \\
\bottomrule
\end{tabular}%
}
\caption{\textbf{Dataset-level MAE for multimodal fine-tuning methods.} Each dataset reports the mean over the 64-16, 128-16, and 256-16 settings. TimeMMD Daily, Weekly, and Monthly retain their native settings as separate blocks. Summary statistics are computed across all 95 settings. The CoCo fine-tuning approach for incorporating text into forecasting outperforms other multimodal fine-tuning methods that jointly incorporate numerical history and text. Bold and underlined values denote the best and second-best method in each row.}
\label{tab:class_mae_pooled_95_multimodal_finetuning}
\end{table*}

%% file: tables/iclr/post_june_2024_zero_shot_multimodal.tex
\begin{table*}[t]
\centering
\scriptsize
\setlength{\tabcolsep}{3.2pt}
\renewcommand{\arraystretch}{1.02}
\resizebox{\textwidth}{!}{%
\begin{tabular}{lclrrrrrrrrr}
\toprule
 & & & & \multicolumn{8}{c}{\textbf{Mean MAE ($\downarrow$)}} \\
\cmidrule(lr){5-12}
\textbf{Group} & \textbf{Setting} & \textbf{Dataset} & \textbf{N} & \cellcolor{gray!15}{\textbf{CoCo}} & \textbf{TimesFM2.5+GPT4o} & \textbf{Chronos2+GPT4o} & \textbf{Chronos2 MV} & \textbf{Toto MV} & \textbf{GPT-4o} & \textbf{Qwen3-30B} & \textbf{TimeOmni} \\
\midrule
\textbf{Media Cloud} & 64-16 & Crude Oil Price & 316 & \cellcolor{gray!15}{0.802} & \textbf{0.752} & \underline{0.758} & 0.817 & 0.760 & 0.877 & 0.924 & 1.082 \\
 &  & S\&P500 Index & 280 & \cellcolor{gray!15}{\textbf{0.759}} & \underline{0.797} & 0.805 & 0.904 & 0.929 & 0.883 & 0.859 & 0.978 \\
 &  & Apple Stock & 319 & \cellcolor{gray!15}{\textbf{0.903}} & 0.977 & \underline{0.977} & 1.043 & 1.014 & 1.089 & 1.059 & 1.467 \\
 &  & Copper Futures & 179 & \cellcolor{gray!15}{\underline{0.775}} & 0.784 & \textbf{0.769} & 0.893 & 0.914 & 0.824 & 0.824 & 1.004 \\
 &  & Egg Price & 144 & \cellcolor{gray!15}{0.701} & \underline{0.651} & 0.669 & 0.673 & \textbf{0.644} & 0.714 & 0.778 & 0.902 \\
 &  & Electricity Price & 334 & \cellcolor{gray!15}{\textbf{0.838}} & 0.887 & 0.907 & \underline{0.846} & 0.873 & 1.029 & 0.967 & 1.294 \\
 &  & SPDR Gold Price & 317 & \cellcolor{gray!15}{\textbf{0.883}} & 0.918 & \underline{0.887} & 1.047 & 1.105 & 0.926 & 1.020 & 3.396 \\
 &  & Google Stock & 319 & \cellcolor{gray!15}{\textbf{0.716}} & 0.770 & \underline{0.768} & 0.867 & 0.847 & 0.853 & 0.843 & 0.851 \\
 &  & NVIDIA Stock Price & 318 & \cellcolor{gray!15}{\textbf{0.848}} & \underline{0.914} & 0.934 & 0.934 & 0.937 & 1.104 & 1.416 & 1.001 \\
 &  & Silver Price & 332 & \cellcolor{gray!15}{\underline{0.903}} & \textbf{0.902} & 0.912 & 1.089 & 1.112 & 0.971 & 0.946 & 1.137 \\
 &  & Steel Rebar Futures & 162 & \cellcolor{gray!15}{\textbf{0.732}} & 0.779 & 0.777 & 0.746 & \underline{0.739} & 0.907 & 0.942 & 1.400 \\
\midrule
\textbf{Media Cloud} & 128-16 & Crude Oil Price & 252 & \cellcolor{gray!15}{0.692} & \textbf{0.618} & 0.641 & 0.678 & \underline{0.629} & 0.709 & 0.850 & 0.856 \\
 &  & S\&P500 Index & 216 & \cellcolor{gray!15}{\textbf{0.573}} & \underline{0.609} & 0.641 & 0.632 & 0.646 & 0.727 & 0.663 & 0.824 \\
 &  & Apple Stock & 255 & \cellcolor{gray!15}{\textbf{0.695}} & \underline{0.716} & 0.743 & 0.736 & 0.725 & 0.834 & 0.876 & 0.897 \\
 &  & Copper Futures & 115 & \cellcolor{gray!15}{0.589} & \underline{0.586} & 0.601 & 0.604 & \textbf{0.565} & 0.626 & 0.649 & 0.702 \\
 &  & Egg Price & 80 & \cellcolor{gray!15}{0.506} & 0.506 & \textbf{0.474} & 0.516 & 0.510 & \underline{0.484} & 0.566 & 0.716 \\
 &  & Electricity Price & 270 & \cellcolor{gray!15}{\textbf{0.597}} & 0.668 & 0.704 & \underline{0.653} & 0.670 & 0.810 & 0.712 & 0.884 \\
 &  & SPDR Gold Price & 253 & \cellcolor{gray!15}{\underline{0.512}} & 0.530 & \textbf{0.503} & 0.633 & 0.639 & 0.519 & 0.590 & 0.775 \\
 &  & Google Stock & 255 & \cellcolor{gray!15}{\underline{0.526}} & \textbf{0.526} & 0.528 & 0.559 & 0.572 & 0.572 & 0.650 & 0.659 \\
 &  & NVIDIA Stock Price & 254 & \cellcolor{gray!15}{\textbf{0.548}} & 0.580 & \underline{0.578} & 0.580 & 0.594 & 0.684 & 0.782 & 0.649 \\
 &  & Silver Price & 268 & \cellcolor{gray!15}{\textbf{0.667}} & 0.676 & \underline{0.676} & 0.805 & 0.810 & 0.761 & 0.765 & 0.864 \\
 &  & Steel Rebar Futures & 98 & \cellcolor{gray!15}{0.483} & 0.427 & 0.442 & \underline{0.423} & \textbf{0.408} & 0.483 & 0.631 & 1.349 \\
\midrule
\textbf{Media Cloud} & 256-16 & Crude Oil Price & 124 & \cellcolor{gray!15}{0.453} & \textbf{0.341} & \underline{0.368} & 0.525 & 0.416 & 0.396 & 0.533 & 0.532 \\
 &  & S\&P500 Index & 88 & \cellcolor{gray!15}{\underline{0.240}} & 0.265 & 0.262 & 0.422 & 0.420 & 0.262 & \textbf{0.239} & 0.370 \\
 &  & Apple Stock & 127 & \cellcolor{gray!15}{\textbf{0.405}} & 0.440 & 0.443 & 0.495 & 0.507 & 0.478 & \underline{0.437} & 0.509 \\
 &  & Electricity Price & 142 & \cellcolor{gray!15}{\textbf{0.467}} & \underline{0.501} & 0.502 & 0.548 & 0.530 & 0.556 & 0.573 & 0.575 \\
 &  & SPDR Gold Price & 125 & \cellcolor{gray!15}{\textbf{0.286}} & 0.312 & \underline{0.287} & 0.309 & 0.335 & 0.308 & 0.346 & 1.085 \\
 &  & Google Stock & 127 & \cellcolor{gray!15}{0.547} & 0.540 & 0.544 & 0.676 & 0.686 & \underline{0.538} & \textbf{0.509} & 0.650 \\
 &  & NVIDIA Stock Price & 126 & \cellcolor{gray!15}{\textbf{0.352}} & 0.373 & \underline{0.366} & 0.483 & 0.441 & 0.402 & 0.483 & 0.459 \\
 &  & Silver Price & 140 & \cellcolor{gray!15}{0.608} & 0.604 & \textbf{0.597} & 0.809 & 0.754 & 0.610 & \underline{0.604} & 0.746 \\
\midrule
 &  & \textbf{Overall Wins} &  & \cellcolor{gray!15}{\textbf{16}} & 5 & 4 & 0 & 3 & 0 & 2 & 0 \\
 &  & \textbf{Mean} &  & \cellcolor{gray!15}{\textbf{0.620}} & 0.632 & 0.635 & 0.698 & 0.691 & 0.698 & 0.734 & 0.954 \\
 &  & \textbf{ELO} &  & \cellcolor{gray!15}{\textbf{1727}} & 1667 & 1652 & 1449 & 1475 & 1436 & 1393 & 1200 \\
 &  & \textbf{Average Rank} &  & \cellcolor{gray!15}{\textbf{2.17}} & 2.77 & 2.90 & 5.03 & 4.80 & 5.17 & 5.67 & 7.50 \\
 &  & \textbf{Head-to-Head Wins} &  & \cellcolor{gray!15}{--} & 20/30 & 20/30 & 27/30 & 24/30 & 27/30 & 27/30 & 30/30 \\
 &  & \textbf{Wilcoxon $p$} &  & \cellcolor{gray!15}{--} & 0.084 & 0.058 & 2.3e-06 & 9.9e-05 & 2.8e-06 & 3.5e-08 & 1.9e-09 \\
\bottomrule
\end{tabular}%
}
\caption{\textbf{Dataset-level MAE for samples strictly after June 2024.} We evaluate CoCo against multimodal baselines in a regime where the LLMs have limited or no prior exposure to the test period. The gains are consistent with those in Table~\ref{tab:class_mae_pooled_95_zero_shot_multimodal}, confirming that the improvements stem from the ability of the model to extract predictive signals from text rather than from memorized knowledge. Bold and underlined values denote the best and second-best method in each row.}
\label{tab:post_june_2024_zero_shot_multimodal}
\end{table*}

%% file: tables/iclr/mase_setting_summary.tex
\begin{table*}[t]
\centering
\small
\renewcommand{\arraystretch}{1.15}
\setlength{\tabcolsep}{3pt}
\resizebox{\textwidth}{!}{%
\begin{tabular}{lcccccccccc|ccccc}
\toprule
 & \multicolumn{10}{c|}{\textbf{Geometric Mean MASE ($\downarrow$)}} & \multicolumn{5}{c}{} \\
\cmidrule(lr){2-11}
 & \multicolumn{3}{c}{\textbf{64--16 Setting}} & \multicolumn{3}{c}{\textbf{128--16 Setting}} & \multicolumn{3}{c}{\textbf{256--16 Setting}} & \textbf{Native} & \textbf{} & \textbf{} &  & \textbf{} & \textbf{} \\
\cmidrule(lr){2-4}\cmidrule(lr){5-7}\cmidrule(lr){8-10}\cmidrule(lr){11-11}
\textbf{Model} & \textbf{Media Cloud} & \textbf{FNSPID} & \textbf{TimeMMD} & \textbf{Media Cloud} & \textbf{FNSPID} & \textbf{TimeMMD} & \textbf{Media Cloud} & \textbf{FNSPID} & \textbf{TimeMMD} & \textbf{TimeMMD$^{*}$} & \textbf{Overall MASE} & \textbf{Avg. Rank} & \textbf{ELO} & \textbf{CoCo Wins} & \textbf{Wilcoxon $p$} \\
\midrule
\multicolumn{16}{l}{\textbf{CoCo}} \\
\rowcolor{gray!15} CoCo & \textbf{3.064} & \underline{3.136} & 1.605 & \textbf{3.009} & \textbf{3.056} & \underline{1.533} & \textbf{2.817} & \underline{2.961} & \underline{1.530} & 1.915 & \textbf{2.455} & \textbf{5.21} & \textbf{1845} & -- & -- \\
\addlinespace[2pt]
\multicolumn{16}{l}{\textbf{Zero-shot Unimodal}} \\
TiRex1 & \underline{3.136} & 3.175 & 1.621 & 3.075 & \underline{3.064} & 1.609 & \underline{2.857} & \textbf{2.950} & 1.662 & 1.936 & \underline{2.506} & \underline{6.62} & 1782 & 59/95 & 9.4e-04 \\
Chronos2 & 3.209 & 3.306 & 1.625 & 3.163 & 3.173 & 1.607 & 2.935 & 3.035 & 1.617 & 1.975 & 2.556 & 9.56 & 1668 & 79/95 & 4.9e-12 \\
TimesFM2.5 & 3.230 & 3.251 & \underline{1.598} & 3.184 & 3.225 & 1.561 & 2.994 & 3.131 & 1.606 & \underline{1.900} & 2.553 & 9.60 & 1672 & 65/95 & 1.1e-08 \\
Moirai & 3.209 & 3.204 & 1.693 & 3.092 & 3.231 & 1.631 & 2.955 & 3.104 & 1.638 & 2.002 & 2.571 & 10.15 & 1655 & 79/95 & 7.5e-12 \\
Toto & 3.330 & 3.362 & 1.661 & 3.172 & 3.355 & 1.627 & 3.050 & 3.246 & 1.632 & 1.951 & 2.625 & 13.60 & 1542 & 83/95 & 1.3e-13 \\
TabPFN2.5 & 3.462 & 3.410 & 1.776 & 3.559 & 3.379 & 1.688 & 4.153 & 3.724 & 1.844 & 2.002 & 2.853 & 18.35 & 1381 & 91/95 & 3.9e-16 \\
TimeMoE & 3.949 & 3.753 & 1.920 & 3.548 & 3.609 & 1.832 & 4.361 & 3.894 & 1.804 & 2.050 & 3.019 & 20.96 & 1281 & 88/95 & 1.3e-15 \\
PANDA & 4.311 & 4.016 & 2.311 & 4.378 & 4.130 & 2.162 & 3.701 & 3.732 & 2.045 & 2.260 & 3.295 & 24.03 & 1136 & 93/95 & 4.6e-17 \\
\addlinespace[2pt]
\multicolumn{16}{l}{\textbf{Zero-shot Multimodal}} \\
TimesFM2.5+GPT4o & 3.143 & 3.150 & 1.787 & 3.076 & 3.175 & 1.739 & 2.933 & 3.087 & 1.642 & 2.070 & 2.583 & 8.80 & 1700 & 67/95 & 3.4e-05 \\
Chronos2+GPT4o & 3.171 & 3.222 & 1.808 & 3.096 & 3.199 & 1.742 & 2.951 & 3.115 & 1.638 & 2.143 & 2.611 & 10.45 & 1642 & 76/95 & 7.2e-08 \\
Chronos2 MV & 3.365 & 3.270 & 1.678 & 3.275 & 3.196 & 1.610 & 3.312 & 3.167 & 1.594 & 1.931 & 2.618 & 12.22 & 1583 & 70/95 & 7.3e-09 \\
Toto MV & 3.389 & 3.365 & 1.647 & 3.248 & 3.378 & 1.592 & 3.222 & 3.278 & 1.568 & 1.903 & 2.634 & 13.11 & 1550 & 72/95 & 1.0e-09 \\
GPT-4o & 3.562 & 3.503 & 2.283 & 3.457 & 3.521 & 2.173 & 3.309 & 3.454 & 1.996 & 2.445 & 2.987 & 19.79 & 1332 & 91/95 & 2.2e-16 \\
Qwen3-30B & 3.663 & 3.488 & 2.391 & 3.686 & 3.635 & 2.382 & 3.529 & 3.682 & 2.359 & 2.485 & 3.152 & 21.17 & 1276 & 94/95 & 2.7e-17 \\
TimeOmni & 4.932 & 3.987 & 3.156 & 5.190 & 3.783 & 4.252 & 4.533 & 3.831 & 3.366 & 2.616 & 3.952 & 25.41 & 1063 & 95/95 & 2.6e-17 \\
\addlinespace[2pt]
\multicolumn{16}{l}{\textbf{Unimodal Fine-tuning}} \\
Chronos2 & 3.281 & 3.242 & \textbf{1.583} & 3.118 & 3.195 & \textbf{1.498} & 2.894 & 3.105 & \textbf{1.481} & 2.008 & 2.522 & 8.96 & 1689 & 64/95 & 2.2e-07 \\
AutoTimes & 3.270 & 3.158 & 1.902 & 3.180 & 3.158 & 1.906 & 3.285 & 3.178 & 1.972 & 1.946 & 2.700 & 11.55 & 1602 & 79/95 & 1.4e-10 \\
PatchTST & 3.337 & 3.283 & 1.942 & 3.252 & 3.235 & 1.883 & 3.037 & 3.333 & 1.964 & 1.926 & 2.729 & 14.01 & 1529 & 88/95 & 4.5e-16 \\
S2IP-LLM & 3.493 & 3.342 & 1.910 & 3.387 & 3.227 & 1.805 & 3.144 & 3.370 & 1.963 & 2.011 & 2.769 & 16.07 & 1455 & 90/95 & 9.0e-16 \\
iTransformer & 3.509 & 3.481 & 1.999 & 4.039 & 4.056 & 2.085 & 3.522 & 3.920 & 2.055 & 1.987 & 3.054 & 21.45 & 1257 & 92/95 & 4.5e-17 \\
TEST & 3.978 & 3.805 & 2.301 & 3.916 & 3.829 & 2.221 & 3.889 & 4.273 & 2.519 & 2.428 & 3.322 & 24.94 & 1091 & 95/95 & 2.6e-17 \\
\addlinespace[2pt]
\multicolumn{16}{l}{\textbf{Multimodal Fine-tuning}} \\
Chronos2 MV & 3.224 & 3.142 & 1.662 & \underline{3.074} & 3.136 & 1.605 & 2.917 & 3.103 & 1.580 & \textbf{1.852} & 2.521 & 7.47 & 1742 & 59/95 & 2.7e-05 \\
TaTS & 3.330 & 3.233 & 1.629 & 3.180 & 3.192 & 1.573 & 3.261 & 3.168 & 1.566 & 1.921 & 2.582 & 10.61 & 1635 & 68/95 & 2.5e-07 \\
VoT & 3.190 & \textbf{3.061} & 1.872 & 3.250 & 3.157 & 1.834 & 3.515 & 3.345 & 1.869 & 1.967 & 2.698 & 11.57 & 1604 & 71/95 & 3.9e-09 \\
S2IP-LLM & 3.316 & 3.147 & 1.913 & 3.329 & 3.259 & 1.872 & 3.400 & 3.768 & 1.989 & 1.945 & 2.789 & 14.36 & 1513 & 83/95 & 3.1e-14 \\
AutoTimes & 3.429 & 3.271 & 2.109 & 3.460 & 3.305 & 2.096 & 3.587 & 3.440 & 2.053 & 1.979 & 2.875 & 16.95 & 1425 & 83/95 & 4.6e-14 \\
TEST & 3.334 & 3.364 & 2.129 & 3.531 & 3.596 & 2.111 & 3.693 & 3.772 & 2.099 & 2.064 & 2.968 & 19.04 & 1351 & 91/95 & 2.9e-16 \\
\bottomrule
\end{tabular}%
}
\caption{\textbf{MASE comparison across all 28 evaluated configurations and 95 settings.} Each setting/group column reports the geometric mean over its constituent datasets; TimeMMD$^{*}$ pools the Daily, Weekly, and Monthly native-frequency benchmarks. The relative performance of CoCo under MASE is consistent with its performance under MAE in Tables~\ref{tab:mae_setting_summary_zs} and \ref{tab:mae_setting_summary_ft}. Bold and underlined values denote the best and second-best models.}
\label{tab:mase_setting_summary}
\end{table*}

%% file: tables/iclr/average_rank_all_models.tex
\begin{table*}[t]
\centering
\scriptsize
\setlength{\tabcolsep}{5pt}
\renewcommand{\arraystretch}{1.05}
\begin{tabular}{lrrrrr}
\toprule
 & \multicolumn{5}{c}{\textbf{Average Rank ($\downarrow$)}} \\
\cmidrule(lr){2-6}
\textbf{Model} & \textbf{MAE} & \textbf{MSE} & \textbf{MAPE} & \textbf{MAE Unscaled} & \textbf{MASE} \\
\midrule
\multicolumn{6}{l}{\textbf{CoCo}} \\
\rowcolor{gray!15} CoCo & \textbf{5.24} & \textbf{5.98} & \textbf{6.39} & \textbf{5.81} & \textbf{5.21} \\
\addlinespace[2pt]
\multicolumn{6}{l}{\textbf{Zero-shot Unimodal}} \\
TiRex1 & \underline{7.26} & \underline{8.63} & \underline{6.92} & \underline{6.97} & \underline{6.62} \\
Chronos2 & 10.18 & 11.16 & 9.81 & 10.11 & 9.56 \\
Moirai & 10.57 & 11.15 & 10.43 & 10.41 & 10.15 \\
TimesFM2.5 & 10.78 & 12.09 & 9.35 & 9.78 & 9.60 \\
Toto & 13.36 & 13.97 & 12.82 & 13.54 & 13.60 \\
TabPFN2.5 & 18.27 & 18.71 & 17.39 & 17.85 & 18.35 \\
TimeMoE & 20.82 & 19.95 & 20.46 & 20.32 & 20.96 \\
PANDA & 24.21 & 23.53 & 23.76 & 23.99 & 24.03 \\
\addlinespace[2pt]
\multicolumn{6}{l}{\textbf{Zero-shot Multimodal}} \\
TimesFM2.5+GPT4o & 8.35 & 8.97 & 8.49 & 8.76 & 8.80 \\
Chronos2+GPT4o & 9.65 & 10.16 & 10.58 & 10.37 & 10.45 \\
Chronos2 MV & 13.42 & 14.45 & 11.09 & 11.41 & 12.22 \\
Toto MV & 14.04 & 14.92 & 11.58 & 12.60 & 13.11 \\
GPT-4o & 18.23 & 17.69 & 19.34 & 18.94 & 19.79 \\
Qwen3-30B & 20.96 & 21.16 & 22.02 & 21.45 & 21.17 \\
TimeOmni & 25.67 & 27.04 & 25.55 & 25.62 & 25.41 \\
\addlinespace[2pt]
\multicolumn{6}{l}{\textbf{Unimodal Fine-tuning}} \\
Chronos2 & 9.18 & 10.57 & 9.80 & 9.55 & 8.96 \\
AutoTimes & 11.75 & 11.17 & 11.56 & 11.41 & 11.55 \\
PatchTST & 13.06 & 12.61 & 15.11 & 14.96 & 14.01 \\
S2IP-LLM & 14.98 & 13.61 & 16.93 & 16.57 & 16.07 \\
iTransformer & 20.49 & 18.01 & 21.84 & 21.58 & 21.45 \\
TEST & 24.14 & 21.93 & 24.97 & 24.98 & 24.94 \\
\addlinespace[2pt]
\multicolumn{6}{l}{\textbf{Multimodal Fine-tuning}} \\
Chronos2 MV & 8.02 & 9.40 & 8.25 & 7.72 & 7.47 \\
VoT & 11.06 & 9.65 & 10.85 & 10.91 & 11.57 \\
TaTS & 11.79 & 13.08 & 9.37 & 9.95 & 10.61 \\
S2IP-LLM & 14.18 & 12.99 & 14.99 & 14.53 & 14.36 \\
AutoTimes & 17.40 & 16.51 & 16.97 & 16.83 & 16.95 \\
TEST & 18.93 & 16.93 & 19.39 & 19.12 & 19.04 \\
\bottomrule
\end{tabular}
\caption{\textbf{CoCo performs consistently across raw and normalized evaluation metrics.} Average ranks are computed over the same 95 settings, similar to Tables~\ref{tab:mae_setting_summary_zs}, \ref{tab:mae_setting_summary_ft}, and~\ref{tab:mase_setting_summary}. Bold and underlined values denote the best and second-best method for each metric.}
\label{tab:average_rank_all_models}
\end{table*}

%% file: tables/iclr/gpt5_zero_shot_multimodal.tex
\begin{table*}[t]
\centering
\scriptsize
\setlength{\tabcolsep}{3.2pt}
\renewcommand{\arraystretch}{1.02}
\resizebox{\textwidth}{!}{%
\begin{tabular}{lclrrrrrrrrrr}
\toprule
 & & & & \multicolumn{9}{c}{\textbf{Mean MAE ($\downarrow$)}} \\
\cmidrule(lr){5-13}
\textbf{Group} & \textbf{Setting} & \textbf{Dataset} & \textbf{N} & \cellcolor{gray!15}{\textbf{CoCo}} & \textbf{TimesFM2.5+GPT4o} & \textbf{Chronos2+GPT4o} & \textbf{Chronos2 MV} & \textbf{Toto MV} & \textbf{GPT-4o} & \textbf{Qwen3-30B} & \textbf{TimeOmni} & \textbf{GPT-5} \\
\midrule
\textbf{Media Cloud} & 64-16 & Crude Oil Price & 421 & \cellcolor{gray!15}{0.877} & \underline{0.817} & 0.831 & 0.853 & \textbf{0.816} & 0.930 & 1.027 & 1.160 & 0.859 \\
 &  & S\&P500 Index & 384 & \cellcolor{gray!15}{\underline{0.797}} & 0.827 & 0.843 & 0.944 & 0.998 & 0.896 & 0.855 & 0.979 & \textbf{0.766} \\
 &  & Apple Stock & 424 & \cellcolor{gray!15}{\underline{0.872}} & 0.948 & 0.945 & 1.008 & 1.003 & 1.043 & 0.961 & 1.325 & \textbf{0.854} \\
 &  & Copper Futures & 179 & \cellcolor{gray!15}{\underline{0.775}} & 0.784 & \textbf{0.769} & 0.893 & 0.914 & 0.824 & 0.824 & 1.004 & 0.787 \\
 &  & Egg Price & 144 & \cellcolor{gray!15}{0.701} & \underline{0.651} & 0.669 & 0.673 & \textbf{0.644} & 0.714 & 0.778 & 0.902 & 0.729 \\
 &  & Electricity Price & 444 & \cellcolor{gray!15}{\textbf{0.798}} & 0.854 & 0.858 & 0.831 & 0.872 & 0.948 & 0.907 & 1.189 & \underline{0.829} \\
 &  & SPDR Gold Price & 422 & \cellcolor{gray!15}{0.849} & 0.866 & \underline{0.844} & 1.010 & 1.066 & 0.895 & 0.976 & 4.565 & \textbf{0.824} \\
 &  & Google Stock & 423 & \cellcolor{gray!15}{\textbf{0.730}} & 0.757 & \underline{0.748} & 0.881 & 0.858 & 0.822 & 0.869 & 0.876 & 0.757 \\
 &  & NVIDIA Stock Price & 423 & \cellcolor{gray!15}{\textbf{0.883}} & \underline{0.939} & 0.950 & 0.966 & 0.984 & 1.084 & 1.330 & 1.021 & 0.948 \\
 &  & Silver Price & 440 & \cellcolor{gray!15}{\textbf{0.886}} & 0.895 & 0.907 & 1.003 & 1.028 & 0.982 & 0.966 & 1.131 & \underline{0.888} \\
 &  & Steel Rebar Futures & 162 & \cellcolor{gray!15}{\textbf{0.732}} & 0.779 & 0.777 & 0.746 & \underline{0.739} & 0.907 & 0.942 & 1.400 & 0.828 \\
\midrule
\textbf{Media Cloud} & 128-16 & Crude Oil Price & 357 & \cellcolor{gray!15}{0.714} & \textbf{0.634} & \underline{0.657} & 0.727 & 0.658 & 0.721 & 0.900 & 0.904 & 0.710 \\
 &  & S\&P500 Index & 320 & \cellcolor{gray!15}{\textbf{0.562}} & 0.607 & 0.617 & 0.641 & 0.656 & 0.700 & 0.650 & 0.811 & \underline{0.574} \\
 &  & Apple Stock & 360 & \cellcolor{gray!15}{\textbf{0.610}} & 0.639 & 0.657 & 0.657 & 0.642 & 0.737 & 0.757 & 0.768 & \underline{0.630} \\
 &  & Copper Futures & 115 & \cellcolor{gray!15}{0.589} & \underline{0.586} & 0.601 & 0.604 & \textbf{0.565} & 0.626 & 0.649 & 0.702 & 0.602 \\
 &  & Egg Price & 80 & \cellcolor{gray!15}{0.506} & 0.506 & \textbf{0.474} & 0.516 & 0.510 & \underline{0.484} & 0.566 & 0.716 & 0.520 \\
 &  & Electricity Price & 380 & \cellcolor{gray!15}{\textbf{0.531}} & 0.585 & 0.609 & 0.606 & 0.613 & 0.686 & 0.631 & 0.789 & \underline{0.579} \\
 &  & SPDR Gold Price & 358 & \cellcolor{gray!15}{\textbf{0.486}} & 0.510 & 0.492 & 0.617 & 0.610 & 0.526 & 0.585 & 3.884 & \underline{0.488} \\
 &  & Google Stock & 359 & \cellcolor{gray!15}{\textbf{0.546}} & 0.566 & 0.575 & \underline{0.549} & 0.573 & 0.643 & 0.678 & 0.685 & 0.568 \\
 &  & NVIDIA Stock Price & 359 & \cellcolor{gray!15}{\textbf{0.541}} & 0.588 & 0.590 & \underline{0.578} & 0.595 & 0.698 & 0.764 & 0.647 & 0.580 \\
 &  & Silver Price & 376 & \cellcolor{gray!15}{\textbf{0.665}} & \underline{0.680} & 0.689 & 0.762 & 0.771 & 0.776 & 0.791 & 0.837 & 0.695 \\
 &  & Steel Rebar Futures & 98 & \cellcolor{gray!15}{0.483} & 0.427 & 0.442 & \underline{0.423} & \textbf{0.408} & 0.483 & 0.631 & 1.349 & 0.482 \\
\midrule
\textbf{Media Cloud} & 256-16 & Crude Oil Price & 229 & \cellcolor{gray!15}{0.509} & \textbf{0.423} & \underline{0.448} & 0.540 & 0.463 & 0.488 & 0.599 & 0.598 & 0.483 \\
 &  & S\&P500 Index & 192 & \cellcolor{gray!15}{\textbf{0.422}} & 0.468 & 0.488 & 0.493 & 0.519 & 0.574 & 0.459 & 0.564 & \underline{0.435} \\
 &  & Apple Stock & 232 & \cellcolor{gray!15}{\textbf{0.428}} & 0.459 & 0.472 & 0.480 & 0.473 & 0.531 & 0.543 & 0.535 & \underline{0.440} \\
 &  & Electricity Price & 252 & \cellcolor{gray!15}{\textbf{0.399}} & \underline{0.445} & 0.456 & 0.456 & 0.454 & 0.523 & 0.467 & 0.575 & 0.445 \\
 &  & SPDR Gold Price & 230 & \cellcolor{gray!15}{0.305} & 0.330 & \underline{0.303} & 0.354 & 0.353 & 0.326 & 0.389 & 1.955 & \textbf{0.293} \\
 &  & Google Stock & 231 & \cellcolor{gray!15}{0.540} & \underline{0.534} & 0.534 & 0.578 & 0.574 & 0.573 & 0.768 & 0.618 & \textbf{0.519} \\
 &  & NVIDIA Stock Price & 231 & \cellcolor{gray!15}{\textbf{0.416}} & 0.446 & \underline{0.435} & 0.503 & 0.485 & 0.494 & 0.561 & 0.533 & 0.457 \\
 &  & Silver Price & 248 & \cellcolor{gray!15}{\underline{0.547}} & \textbf{0.544} & 0.547 & 0.723 & 0.670 & 0.592 & 0.584 & 0.704 & 0.561 \\
\midrule
 &  & \textbf{Overall Wins} &  & \cellcolor{gray!15}{\textbf{16}} & 3 & 2 & 0 & 4 & 0 & 0 & 0 & 5 \\
 &  & \textbf{Mean} &  & \cellcolor{gray!15}{\textbf{0.623}} & 0.636 & 0.641 & 0.687 & 0.684 & 0.708 & 0.747 & 1.124 & 0.638 \\
 &  & \textbf{ELO} &  & \cellcolor{gray!15}{\textbf{1745}} & 1677 & 1633 & 1440 & 1487 & 1381 & 1311 & 1173 & 1654 \\
 &  & \textbf{Average Rank} &  & \cellcolor{gray!15}{\textbf{2.27}} & 2.97 & 3.47 & 5.67 & 5.17 & 6.43 & 7.23 & 8.60 & 3.20 \\
 &  & \textbf{Head-to-Head Wins} &  & \cellcolor{gray!15}{--} & 22/30 & 20/30 & 27/30 & 24/30 & 28/30 & 30/30 & 30/30 & 21/30 \\
 &  & \textbf{Wilcoxon $p$} &  & \cellcolor{gray!15}{--} & 0.067 & 0.028 & 3.2e-06 & 3.4e-04 & 1.0e-07 & 1.9e-09 & 1.9e-09 & 0.015 \\
\bottomrule
\end{tabular}%
}
\caption{\textbf{Dataset-level MAE for 16-step forecasts across context lengths 64, 128, and 256 on the Media Cloud subset, including GPT-5 as an additional LLM baseline.} Media Cloud, comprising datasets from 2024--2025, is the only subset where GPT-5 has limited prior exposure to test trends. GPT-5 performs competitively with strong TSFMs, yet CoCo outperforms GPT-5 in 70\% (21/30) of settings, demonstrating that a fine-tuned 4B LLM providing forecast-space covariates to Chronos2 surpasses a closed-source model used as a standalone forecaster. Bold and underlined values denote the best and second-best method in each row.}
\label{tab:gpt5_zero_shot_multimodal}
\end{table*}

%% file: tables/adversial_text.tex
\begin{table}[ht]
\centering
\small
\renewcommand{\arraystretch}{1.12}
\resizebox{0.50\textwidth}{!}{%
\begin{tabular}{cccc}
\toprule
\multicolumn{3}{c}{MAE ($\downarrow$)} \\
\midrule
 CoCo Fine-tuned
 & CoCo
 & \textbf{CoCo}
 \\
  Chronos2
 & (With Misleading Text)
 & \textbf{(With Truthful Text)}
 \\
\midrule

 2.929
 & 3.025
 & \textbf{2.802}
 \\
\bottomrule
\end{tabular}
}%
\vspace{0.1in}
\caption{\textbf{CoCo's performance degrades when text is adversarial.} 
We evaluate on a synthetically generated dataset of 1000 samples 
where the same numerical series is paired with two kinds of 
annotations: \textit{truthful} text describing the correct future 
trend, and \textit{misleading} text deliberately describing the 
opposite trend. CoCo improves over the fine-tuned Chronos2 baseline 
with truthful text but degrades with misleading text, consistent 
with the gating analysis in Figure~\ref{fig:text_use_ablation} (right), where confidently wrong LLM forecasts are the most harmful regime.
}
\label{tab:adversial_text}
\end{table}

%% file: tables/covariate_usefulness.tex
\begin{table}[t]
    \centering
    \small
    \renewcommand{\arraystretch}{1.12}
    \resizebox{\textwidth}{!}{%
    \begin{tabular}{lcccccccccccc}
    \toprule
     & \multicolumn{2}{c}{LLM Forecast}
     & \multicolumn{2}{c}{Chronos2 Forecast}
     & \multicolumn{2}{c}{CoCo}
     & \multicolumn{2}{c}{CoCo} \\

     & \multicolumn{2}{c}{$\theta_\mathrm{LLM}(X^H_t, S_t)$}
     & \multicolumn{2}{c}{$\theta_\mathrm{M-TSFM}(X^H_t)$}
     & \multicolumn{2}{c}{$\theta_\mathrm{M-TSFM}(X^H_t||M_t)$}
     & \multicolumn{2}{c}{$\theta_\mathrm{M-TSFM}(X^H_t||M_t||C_t)$} \\

    \cmidrule(lr){2-3} \cmidrule(lr){4-5} \cmidrule(lr){6-7} \cmidrule(lr){8-9}
    \textbf{LLM}
     & MAE ($\downarrow$)
     & Rank ($\downarrow$)
     & MAE ($\downarrow$)
     & Rank ($\downarrow$)
     & MAE ($\downarrow$)
     & Rank ($\downarrow$)
     & MAE ($\downarrow$)
     & Rank ($\downarrow$)
     \\
    \midrule
    Zeroshot
     & 1.270 & 8.842
     & 0.980 & - 
     & 1.028 & 2.579 
     & 1.110 & 4.676 
     \\
    Qwen3-4b Fine-tuned
     & 1.078 & 3.816 
     & 0.980 & - 
     & 0.978 & 2.632 
     & 0.964 & 2.684 
     \\
    Chronos2 Fine-tuned
     & 1.270 & 8.842 
     & 0.959 & 2.342 
     & 0.956 & 2.459 
     & 0.955 & 2.459 
     \\
    Both Fine-tuned
     & 1.078 & 3.816 
     & 0.959 & 2.342 
     & 0.954 & 2.263
     & \textbf{0.947} & \textbf{2.184} 
     \\
    \bottomrule
    \end{tabular}
    }%
    \vspace{0.1in}
    \caption{\textbf{Effect of fine-tuning and the usefulness of the covariates.} Extended version of Table~\ref{tab:finetuning_ablation_sumary} showing MAE and average rank over the 30 datasets for the 64-16 setting and the 8 TimeMMD datasets with their native settings (38 settings in total) for different fine-tuning settings and covariate inputs. `-' indicates that the rank is not valid since Chronos2 is one of the baselines. 
    }
    \label{tab:covariate_finetuning_usefulness}
\end{table}

%% file: tables/rollout_k_effect.tex
\begin{table}[t]
    \centering
    \small
    \renewcommand{\arraystretch}{1.12}
    \resizebox{0.4\textwidth}{!}{%
    \begin{tabular}{lcc}
    \toprule
     & \multicolumn{2}{c}{CoCo Forecast}
     \\
     & \multicolumn{2}{c}{$\theta_\mathrm{M-TSFM}(X^H_t||M_t||C_t)$}
     \\
    \cmidrule(lr){2-3} 
    \textbf{$N_{\mathrm{rollouts}}$}
     & MAE ($\downarrow$)
     & Rank ($\downarrow$)
     \\
    \midrule
    5
     & \textbf{0.845} & \textbf{1.833}  
     \\
    9
     & \textbf{0.845} & 1.867 
     \\
    15
     & 0.846 & 1.900 
     \\
    \bottomrule
    \end{tabular}
    }%
    \vspace{0.1in}
    \caption{\textbf{Effect of $N_{\mathrm{rollouts}}$, the number of rollouts used to compute confidence covariate on performance.} The MAE and the average rank are shown over the 30 datasets for 16-step forecasts with a context length of 64.}
    \label{tab:covariate_baselines_combined}
\end{table}

%% file: tables/text_in_finetuning.tex
\begin{table}[t]
    \centering
    \small
    \renewcommand{\arraystretch}{1.12}
    \resizebox{\textwidth}{!}{%
    \begin{tabular}{lcccccccccccc}
    \toprule
     & \multicolumn{4}{c}{LLM Forecast ($\theta_\mathrm{LLM}(X^H_t,S_t)$)}
     \\
    \cmidrule(lr){2-5} 
    Qwen3-4b Fine-tuning
     & Overall MAE ($\downarrow$)
     & TimeMMD MAE ($\downarrow$)
     & FNSPID MAE ($\downarrow$)
     & Rank ($\downarrow$)
     \\
    \midrule
    Supervised Fine-tuning
     & 1.187 & 1.635 & 1.050 & 8.289
    \\
    CoCo Fine-tuning without Text
     & 1.172 & 1.543 & 0.954 & 4.395  
     \\
    CoCo Fine-tuning with Synthetic Text
     & 1.087 & 1.365 & 0.903 & 3.974 
     \\
    CoCo Fine-tuning with Real Text
     & 1.140 & 1.528 & \textbf{0.891} & \textbf{3.605} 
    \\
    CoCo Fine-tuning with Real and Synthetic Text
     & \textbf{1.078} & \textbf{1.354} & 0.894 & 3.737 
     \\
    \bottomrule
    \end{tabular}
    }%
    \vspace{0.1in}
    \caption{\textbf{Effect of using text and the kind of text used in fine-tuning the LLM.} The results are for the \texttt{Qwen3-4B-Instruct-2507} model. Real text here denotes FNSPID data and synthetic text refers to the dataset described in Appendix~\ref{sec:training_datasets}. Though the model trained completely with FNSPID data has a marginally better rank (the wins are concentrated in the FNSPID subset, which was seen during training), we observe that it fails in the TimeMMD dataset, which is out of distribution and hence does not generalize. On the other hand, models trained on completely synthetic text or a mixture of both provide improved performance in TimeMMD as well. The MAE and average ranks reported are over the same 38 settings as Table~\ref{tab:covariate_finetuning_usefulness}.}
    \label{tab:text_in_finetuning}
\end{table}

%% file: tables/rebuttal/tsfm_llm_generalization.tex
\begin{table}[t]
    \centering
    \small
    \renewcommand{\arraystretch}{1.12}
    \resizebox{\textwidth}{!}{%
    \begin{tabular}{llcccc}
    \toprule
     &  & \multicolumn{4}{c}{Mean MAE ($\downarrow$)} \\
\cmidrule(lr){3-6}
    M-TSFM & LLM
    & $\theta_\mathrm{LLM}(X^H_t,S_t)$
    & $\theta_\mathrm{\mathrm{M-TSFM}}(X^H_t)$
    & $\theta_\mathrm{\mathrm{M-TSFM}}(X^H_t\Vert M_t)$
    & $\theta_\mathrm{\mathrm{M-TSFM}}(X^H_t\Vert M_t\Vert C_t)$ \\
    \midrule

    \multirow{5}{*}{\textbf{Chronos2}}
    & GPT-OSS-120B & 0.885 & 0.847 & \textbf{0.828} & 0.842 \\
    & Qwen3-30B   & 0.949 & 0.847 & \textbf{0.816} & 0.851 \\
    & Gemma3-27B  & 0.871 & 0.847 & \textbf{0.821} & 0.827 \\
    & Qwen3-4B    & 1.019 & 0.847 & \textbf{0.824} & 0.895 \\
    & Qwen3-4B (FT) & 0.862 & 0.847 & 0.818 & \textbf{0.801} \\

    \midrule

    \multirow{5}{*}{\textbf{Toto}}
    & GPT-OSS-120B & 0.885 & 0.865 & 0.864 & \textbf{0.854} \\
    & Qwen3-30B   & 0.949 & 0.865 & 0.859 & \textbf{0.851} \\
    & Gemma3-27B  & 0.871 & 0.865 & 0.859 & \textbf{0.855} \\
    & Qwen3-4B    & 1.019 & 0.865 & 0.863 & \textbf{0.858} \\
    & Qwen3-4B (FT) & 0.862 & 0.865 & 0.862 & \textbf{0.857} \\

    \midrule

    \multirow{5}{*}{\textbf{Moirai}}
    & GPT-OSS-120B & 0.885 & 0.843 & \textbf{0.838} & 0.844 \\
    & Qwen3-30B   & 0.949 & 0.843 & 0.841 & \textbf{0.839} \\
    & Gemma3-27B  & 0.871 & 0.843 & \textbf{0.839} & 0.843 \\
    & Qwen3-4B    & 1.019 & 0.843 & \textbf{0.839} & 0.841 \\
    & Qwen3-4B (FT) & 0.862 & 0.843 & \textbf{0.841} & 0.845 \\

    \midrule

    \multirow{5}{*}{\textbf{Chronos2 (LoRA FT)}}
    & GPT-OSS-120B & 0.885 & 0.824 & 0.816 & \textbf{0.813} \\
    & Qwen3-30B   & 0.949 & 0.824 & 0.814 & \textbf{0.811} \\
    & Gemma3-27B  & 0.871 & 0.824 & 0.814 & \textbf{0.810} \\
    & Qwen3-4B    & 1.019 & 0.824 & 0.815 & \textbf{0.815} \\
    & Qwen3-4B (FT) & 0.862 & 0.824 & 0.815 & \textbf{0.809} \\

    \bottomrule
    \end{tabular}
    }%
    \vspace{0.1in}
    \caption{\textbf{CoCo generalizes across LLM and M-TSFM choices.}
Mean MAE is reported on the Media Cloud datasets for the 64-16 context--horizon setting. Across five LLM configurations and three M-TSFM backbones, the forecast covariate $M_t$ improves over the corresponding M-TSFM baseline in every evaluated combination. The confidence covariate $C_t$ provides further gains with the zero-shot Toto backbone and after fine-tuning Chronos2. Importantly, Chronos2 is fine-tuned using rollouts pre-generated only by the fine-tuned Qwen3-4B, yet it generalizes to covariates from every evaluated zero-shot LLM. These results demonstrate that CoCo is not reliant on properties specific to Qwen3-4B or Chronos2. \textbf{Bold} denotes the best result in each row.}
    \label{tab:additional_llms_tsfms}
\end{table}

%% file: tables/coco_with_toto.tex
\begin{table*}[t]
\centering
\small
\renewcommand{\arraystretch}{1.15}

\resizebox{\textwidth}{!}{%
\begin{tabular}{lllccccc}
\toprule
 &  &  &  & \multicolumn{4}{c}{Mean MAE ($\downarrow$)} \\
\cmidrule(lr){5-8}

 &  &  &  & TSFM & TSFM MV & CoCo (With Toto) & \textbf{CoCo (With Toto)} \\
\textbf{Group} & \textbf{Setting} & \textbf{Dataset} & \textbf{N} & Toto & Toto & Only Forecast Covariate & \textbf{Both Covariates} \\
\midrule

\textbf{Suite} & 64-16 & Crude Oil Price & 421 & 0.900 & \textbf{\textcolor{black}{0.816}} & 0.899 & \cellcolor[HTML]{EFEFEF}{\underline{\textcolor{black}{0.899}}} \\
 &  & S\&P500 Index & 384 & \textbf{\textcolor{black}{0.926}} & 0.998 & 0.959 & \cellcolor[HTML]{EFEFEF}{\underline{\textcolor{black}{0.948}}} \\
 &  & Apple Stock & 424 & 0.959 & 1.003 & \underline{\textcolor{black}{0.951}} & \cellcolor[HTML]{EFEFEF}{\textbf{\textcolor{black}{0.941}}} \\
 &  & Copper Futures & 179 & 0.827 & 0.914 & \textbf{\textcolor{black}{0.814}} & \cellcolor[HTML]{EFEFEF}{\underline{\textcolor{black}{0.815}}} \\
 &  & Egg Price & 144 & \underline{\textcolor{black}{0.666}} & \textbf{\textcolor{black}{0.644}} & 0.738 & \cellcolor[HTML]{EFEFEF}{0.694} \\
 &  & Electricity Price & 444 & 0.847 & 0.872 & \underline{\textcolor{black}{0.840}} & \cellcolor[HTML]{EFEFEF}{\textbf{\textcolor{black}{0.838}}} \\
 &  & SPDR Gold Price & 422 & 0.987 & 1.066 & \underline{\textcolor{black}{0.968}} & \cellcolor[HTML]{EFEFEF}{\textbf{\textcolor{black}{0.953}}} \\
 &  & Google Stock & 423 & 0.794 & 0.858 & \underline{\textcolor{black}{0.780}} & \cellcolor[HTML]{EFEFEF}{\textbf{\textcolor{black}{0.775}}} \\
 &  & NVIDIA Stock Price & 423 & 0.935 & 0.984 & \underline{\textcolor{black}{0.926}} & \cellcolor[HTML]{EFEFEF}{\textbf{\textcolor{black}{0.914}}} \\
 &  & Silver Price & 440 & \textbf{\textcolor{black}{0.924}} & 1.028 & 0.925 & \cellcolor[HTML]{EFEFEF}{\underline{\textcolor{black}{0.925}}} \\
 &  & Steel Rebar Futures & 162 & 0.747 & \underline{\textcolor{black}{0.739}} & 0.740 & \cellcolor[HTML]{EFEFEF}{\textbf{\textcolor{black}{0.739}}} \\
\midrule

\textbf{FNSPID} & 64-16 & Boeing Stock & 421 & 0.834 & \underline{\textcolor{black}{0.822}} & 0.824 & \cellcolor[HTML]{EFEFEF}{\textbf{\textcolor{black}{0.813}}} \\
 &  & Bank of America Stock & 420 & 1.103 & \textbf{\textcolor{black}{1.084}} & 1.111 & \cellcolor[HTML]{EFEFEF}{\underline{\textcolor{black}{1.101}}} \\
 &  & Costco Stock & 421 & \underline{\textcolor{black}{0.842}} & 0.875 & 0.843 & \cellcolor[HTML]{EFEFEF}{\textbf{\textcolor{black}{0.829}}} \\
 &  & Goldman Sachs Stock & 421 & 0.968 & \textbf{\textcolor{black}{0.935}} & 0.959 & \cellcolor[HTML]{EFEFEF}{\underline{\textcolor{black}{0.951}}} \\
 &  & JPMC Stock & 358 & \underline{\textcolor{black}{1.131}} & \textbf{\textcolor{black}{1.117}} & 1.134 & \cellcolor[HTML]{EFEFEF}{1.132} \\
 &  & McDonald's Stock & 421 & 0.957 & 1.005 & \underline{\textcolor{black}{0.946}} & \cellcolor[HTML]{EFEFEF}{\textbf{\textcolor{black}{0.938}}} \\
 &  & Microsoft Stock & 333 & 0.947 & 0.987 & \underline{\textcolor{black}{0.943}} & \cellcolor[HTML]{EFEFEF}{\textbf{\textcolor{black}{0.938}}} \\
 &  & Oracle Stock & 421 & 0.913 & 0.926 & \underline{\textcolor{black}{0.906}} & \cellcolor[HTML]{EFEFEF}{\textbf{\textcolor{black}{0.894}}} \\
 &  & Tesla Stock & 331 & 0.848 & \underline{\textcolor{black}{0.829}} & 0.830 & \cellcolor[HTML]{EFEFEF}{\textbf{\textcolor{black}{0.817}}} \\
 &  & Walmart Stock & 421 & 0.987 & 0.991 & \underline{\textcolor{black}{0.970}} & \cellcolor[HTML]{EFEFEF}{\textbf{\textcolor{black}{0.958}}} \\
 &  & Exxon Mobil Stock & 421 & 0.972 & 1.008 & \underline{\textcolor{black}{0.956}} & \cellcolor[HTML]{EFEFEF}{\textbf{\textcolor{black}{0.952}}} \\
\midrule

\textbf{TimeMMD} & 64-16 & Broiler Price & 453 & 0.984 & 1.063 & \underline{\textcolor{black}{0.978}} & \cellcolor[HTML]{EFEFEF}{\textbf{\textcolor{black}{0.974}}} \\
 &  & Drought Level & 1193 & 0.909 & 0.920 & \underline{\textcolor{black}{0.907}} & \cellcolor[HTML]{EFEFEF}{\textbf{\textcolor{black}{0.904}}} \\
 &  & US Trade Balance & 368 & \underline{\textcolor{black}{0.882}} & \textbf{\textcolor{black}{0.837}} & 0.901 & \cellcolor[HTML]{EFEFEF}{0.899} \\
 &  & Gasoline Price & 1543 & 0.926 & 0.937 & \textbf{\textcolor{black}{0.907}} & \cellcolor[HTML]{EFEFEF}{\underline{\textcolor{black}{0.912}}} \\
 &  & Air Quality Index & 1401 & 0.886 & \textbf{\textcolor{black}{0.882}} & 0.884 & \cellcolor[HTML]{EFEFEF}{\underline{\textcolor{black}{0.884}}} \\
 &  & Disaster Grants & 230 & \underline{\textcolor{black}{1.139}} & 1.144 & \textbf{\textcolor{black}{1.136}} & \cellcolor[HTML]{EFEFEF}{1.140} \\
 &  & Unemployment Rate & 453 & \textbf{\textcolor{black}{0.659}} & \underline{\textcolor{black}{0.660}} & 0.670 & \cellcolor[HTML]{EFEFEF}{0.680} \\
 &  & Travel Volume & 453 & \textbf{\textcolor{black}{0.304}} & \underline{\textcolor{black}{0.308}} & 0.319 & \cellcolor[HTML]{EFEFEF}{0.316} \\

\midrule
 &  & \textbf{Overall Wins} &  & 4 & 7 & 3 & \textbf{16} \\
 &  & \textbf{Mean} &  & 0.890 & 0.908 & 0.889 & \textbf{0.882} \\
 &  & \textbf{ELO} &  & 1458 & 1415 & 1486 & \textbf{1640} \\
 &  & \textbf{Average Rank} &  & 2.767 & 2.967 & 2.567 & \textbf{1.700} \\
 &  & \textbf{Head-to-Head Wins} &  & 22/30 & 21/30 & 26/30 & - \\
 &  & \textbf{Wilcoxon $\mathbf{p}$} &  & 1.28e-02 & 9.3e-03 & 4.97e-05 & - \\
\bottomrule
\end{tabular}
}%

\caption{\textbf{Results with Toto as the TSFM backbone, using covariates generated by the fine-tuned Qwen3-4b LLM without any Toto-specific fine-tuning.} CoCo (with Toto) outperforms Toto in $73\%\ (22/30)$ of datasets and outperforms sentence transformer augmented Toto (TSFM MV) in $70\%\ (21/30)$ of the datasets, demonstrating that the covariate design transfers across TSFM backbones and is not specific to Chronos2. \textbf{Bold} and \underline{underlined} indicate the best and second best per dataset respectively.}
\label{tab:coco_with_toto}
\end{table*}

%% file: tables/rebuttal/leakage_probe_analysis.tex
\begin{table}[t]
    \centering
    \small
    \renewcommand{\arraystretch}{1.12}
    \resizebox{\textwidth}{!}{%
    \begin{tabular}{lccc}
    \toprule
    \textbf{Annotations} & \textbf{Direction} & \textbf{Magnitude Tertile} & \textbf{Argmax Bucket} \\
    \midrule
    Media Cloud (Real) & 0.533 & 0.344 & 0.301 \\
    FNSPID (Real) & 0.544 & 0.340 & 0.279 \\
    Synthetic (LLM generated with ground truth future values) & 0.555 & 0.316 & 0.272 \\
    Synthetic (LLM generated without ground truth future values) & 0.491 & 0.376 & 0.346 \\
    \bottomrule
    \end{tabular}
    }%
    \vspace{0.1in}
    \caption{\textbf{Synthetic textual annotations that depend on numeric history alone provide incorrect information about the true future.} In this table, we show the accuracy of linear probes for predicting the domain-agnostic forecast statistics, such as the direction, magnitude tertile, and the argmax location/bucket. Note that the linear probes take an LLM's embeddings of the context as input. In our training, we utilize synthetic annotations/context that is synthetically generated (Section \ref{sec:dataset_gen}). Here, we ablate the synthetic annotation generation strategy. We show two approaches: one where the numeric future is used to generate the annotation (row 3) and one where the numeric future is not used (row 4). We compare the accuracies of predicting domain-agnostic forecast statistics from these two variants of synthetic annotations with the accuracies of predicting the same from real-annotations. Prefix-only (without the numeric future, row 4) annotations are generated using only domain information and the observed history, without access to forecast values. Compared with real and future-conditioned annotations, they yield lower direction accuracy but substantially higher magnitude-tertile and argmax-bucket accuracies, indicating an out-of-distribution probe profile relative to the real text. This mismatch suggests that prefix-only generation may weaken sim-to-real transfer and motivates our randomized-context design (Section \ref{sec:dataset_gen}) as a middle ground between limiting future leakage and preserving real-world text statistics. However, one key limitation of this experiment is that the probes are linear, and more non-linear approaches could extract better forecast statistics from prefix-only annotations as well.}
    \label{tab:annotation_probe_accuracy}
\end{table}

%% file: tables/rebuttal/finetuned_component_ablation.tex
\begin{table}[t]
    \centering
    \small
    \renewcommand{\arraystretch}{1.12}
    \resizebox{\textwidth}{!}{%
    \begin{tabular}{lccccc}
    \toprule
    \textbf{Fine-tuned Components}
    & \textbf{Mean MAE ($\downarrow$)}
    & \textbf{Avg. Rank among Baselines ($\downarrow$)}
    & \textbf{ELO among Baselines ($\uparrow$)}
    & \textbf{Avg. Rank among Variants ($\downarrow$)}
    & \textbf{ELO among Variants ($\uparrow$)} \\
    \midrule
    M-TSFM \& LLM & \textbf{0.719} & \textbf{5.24} & \textbf{1829} & \textbf{1.87} & \textbf{1535} \\
    LLM         & 0.732          & 6.44          & 1785          & 1.93          & 1508 \\
    M-TSFM        & 0.724          & 7.16          & 1758          & 2.19          & 1457 \\
    \bottomrule
    \end{tabular}
    }%
    \vspace{0.1in}
    \caption{\textbf{Ablation of fine-tuned components in CoCo across all 95 settings.} We compare variants with only the LLM fine-tuned, only the M-TSFM fine-tuned, and both components fine-tuned, evaluating each against all the 27 methods as well as against the other CoCo variants. Fine-tuning both components jointly (in a sequential manner as described in Section \ref{sec:model}) achieves the best mean MAE, average rank, and ELO in both comparisons. Fine-tuning either component individually also outperforms the zero-shot baselines, demonstrating that both stages contribute independently. Notably, the advantage of jointly fine-tuning both components is more pronounced when evaluated against the full baseline pool than in the direct three-way comparison.}
    \label{tab:finetuned_components_ablation}
\end{table}

%% file: figures_latex/standalone_llm_forecast.tex
\begin{figure}[t]
    \centering
    \includegraphics[width=0.65\linewidth]{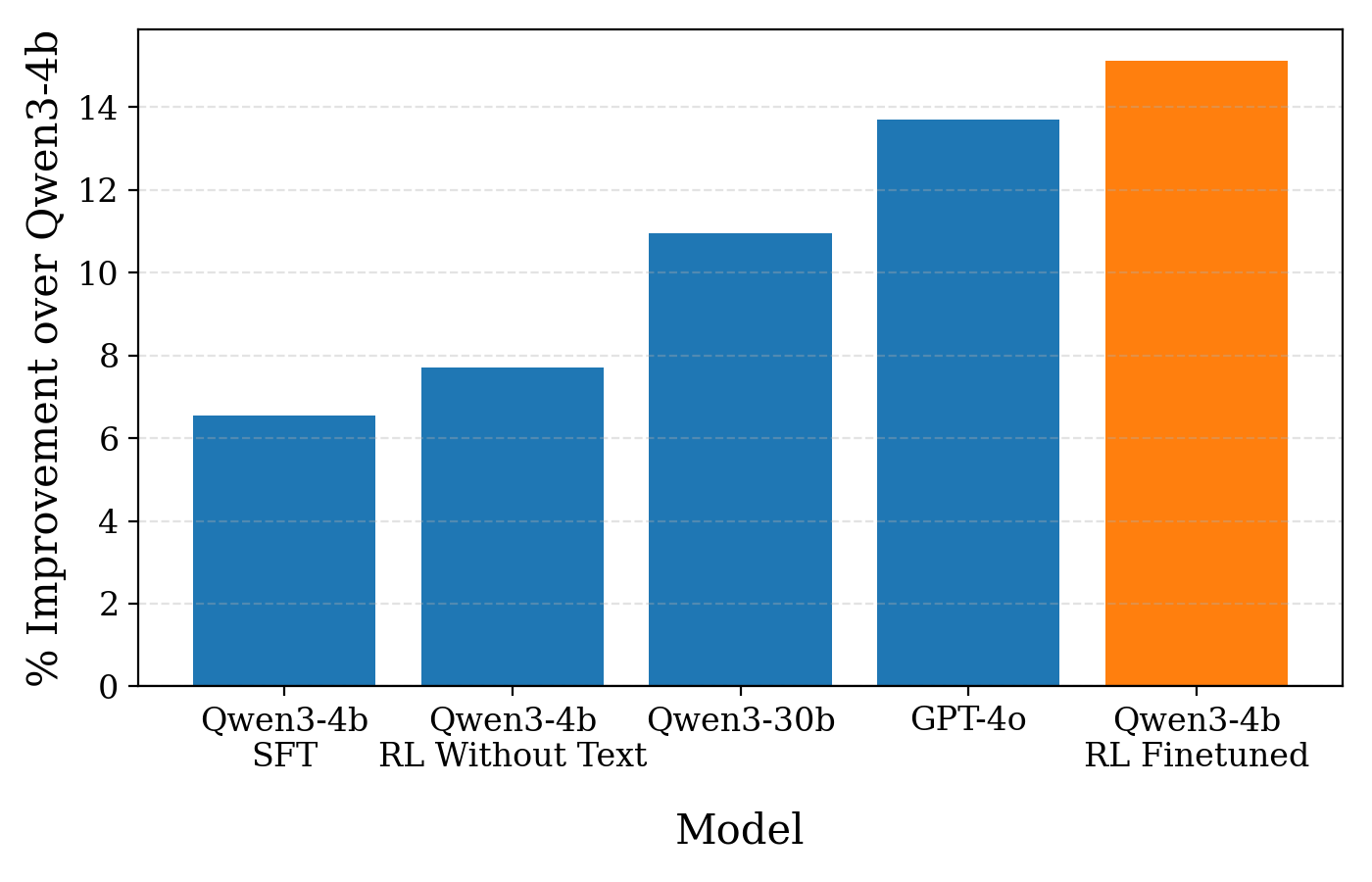}
    \caption{\textbf{Comparison of standalone forecasting performance of LLMs.} The plot shows the \%-improvement in the MAE over the base Qwen3-4B model across the same 38 settings as Table~\ref{tab:covariate_finetuning_usefulness}. We observe that the CoCo fine-tuned Qwen3-4B outperforms larger and closed-source LLMs as a standalone forecaster.
}
    \label{fig:standalone_llm_forecast}
\end{figure}

%% file: tables/latency.tex
\begin{table}[t]
\centering
\small
\sc
\begin{tabular}{lr}
\toprule
Model & Latency (ms/sample) \\
\midrule
Chronos2 & \textbf{11.43 $\pm$ 7.18} \\
Toto & 31.15 $\pm$ 39.67 \\
TimesFM 2.5 & 37.08 $\pm$ 52.43 \\
Chronos MV & 75.77 $\pm$ 54.26 \\
TabPFN 2.5 & 270.17 $\pm$ 28.62 \\
\cellcolor[HTML]{EFEFEF}{CoCo} & \cellcolor[HTML]{EFEFEF}{280.07 $\pm$ 21.96} \\
Toto MV & 312.14 $\pm$ 2.47 \\
\bottomrule
\end{tabular}
\vspace{0.1in}
\caption{\textbf{Inference latency (ms/sample) for CoCo and all TSFM baselines, measured on a single H100 GPU.} Latency is computed by processing 128 samples (16-step forecasts with a context length of 64) in batches of 16 and averaging over batches. For CoCo, the fine-tuned Qwen3-4b-Instruct-2507 is served via a vLLM inference server on the same GPU, with $k=5$ rollouts generated per sample for the confidence covariate $C_t$. Despite requiring LLM inference with multiple rollouts, CoCo 
achieves $280$ ms/sample, comparable to TabPFN 2.5. The latency overhead is a direct 
consequence of incorporating an LLM, which we view as necessary for extracting predictive signals from text, a capability that unimodal TSFMs fundamentally lack.}
\label{tab:latency}
\end{table}

%% file: appendix/6_llm_prompts.tex
\clearpage

\section{LLM Prompts}
\label{llm_prompts}

\subsection{Synthetic Annotation Prompt}
\label{sec:synthetic_annotation_gen} 

The following prompt is used to generate the large-scale training dataset in Section \ref{sec:training_datasets}.
\begin{tcolorbox}[
  colback=gray!10,
  colframe=gray!35,
  boxrule=0.4pt,
  arc=2pt,
  left=6pt,
  right=6pt,
  top=6pt,
  bottom=6pt,
  breakable,
  listing only
]
\footnotesize\ttfamily
\begin{Verbatim}[
  fontsize=\small,
  breaklines=true,
  breakanywhere=true,
  breaksymbolleft={},
  breaksymbolright={}
]
You are given a time series for the economic indicator: { Series Title / Information }. The data consists of dates (t) and corresponding values in appropriate units. Your task is to annotate each data point with a text column that serves as an analytical briefing note. For each data point (t, y_t), follow this structured thinking process to generate the text:

1.  Observation: State the current value and its immediate change. Is it an increase or decrease? Is the pace of change accelerating, decelerating, or stable compared to the prior few periods?

2.  Contextualization: Explain this movement by citing 2-3 plausible, interconnected macroeconomic drivers. Make sure to use factors that align and are relevant to the time period / date of the data point.

3.  Outlook: Provide a speculative, forward-looking statement. This outlook must NOT reveal future data. Instead, frame it in terms of risks, potential, and key indicators to monitor for confirmation. Use cautious but insightful language.

You don't need to have explicit sections in your response. Just the text.

Key Directives:
-   Report-Like Tone: The text must be formal, objective, and analytical.
-   Predictive Value: The combination of context and outlook should provide a rich narrative basis for forecasting without leaking future values.
-   Dynamic Narrative: Do not use repetitive sentence structures or explanations. Each annotation should feel like a unique daily or monthly report.
-   Conciseness: Keep each annotation under 150 words.
-   Output Format: Your entire output must be in CSV format with three columns:`t`, `y_t`, `text`. Do not add headers, markdown, or any text outside of the CSV rows.

Begin generating the CSV output now.

Data:
{Data as rows containing timestamp and value.}
\end{Verbatim}
\end{tcolorbox}

\subsection{LLM Forecast Prompt}
\label{sec:gpt_oss_forecast}
The following prompt is used to obtain forecasts from LLMs, given the numerical history and textual annotations. It is used to both evaluate LLMs for forecasting performance and to obtain the covariates required for CoCo.
\label{sec:gpt_oss_eval_prompt}
\begin{tcolorbox}[
  colback=gray!10,
  colframe=gray!35,
  boxrule=0.4pt,
  arc=2pt,
  left=6pt,
  right=6pt,
  top=6pt,
  bottom=6pt,
  breakable
]
\small\ttfamily
\begin{Verbatim}[
  fontsize=\small,
  breaklines=true,
  breakanywhere=true,
  breaksymbolleft={},
  breaksymbolright={}
]
You are a time series forecasting expert. Analyze the historical data with text annotations to predict future values.

FULL TIME SERIES ({l: context length} timesteps):
[
{Data as rows containing timestamps and values}
]

DETAILED CONTEXT (most recent 16 timesteps with text annotations):
[
{Textual context as rows containing timestamps and annotations}
]

TASK:
Analyze the full time series pattern and the text annotations for:
1.  FACTUAL PATTERNS: What trends, events, or behaviors are described in the recent history?
2.  PREDICTIVE SIGNALS: Are there any forward-looking statements, expectations, or indicators of future change?

Based on this analysis, provide a forecast for the next {h: prediction horizon} steps. FORECAST TARGET TIMESTAMPS (next steps, in order):
[
{Timestamps to forecast}
]

IMPORTANT: Your forecast numbers MUST correspond to these timestamps in the same order. Output ONLY {h: prediction horizon} comma-separated numbers representing your forecast. Do not include any explanations or other text.

Forecast:
\end{Verbatim}
\end{tcolorbox}

\subsection{Media Cloud Dataset Collection}\label{mediacloud_prompts}

\subsubsection{Prompt used to filter irrelevant articles and summarize relevant articles}

\label{sec:media_cloud_filter}
\begin{tcolorbox}[
  colback=gray!10,
  colframe=gray!35,
  boxrule=0.4pt,
  arc=2pt,
  left=6pt,
  right=6pt,
  top=6pt,
  bottom=6pt,
  breakable
]
\small\ttfamily
\begin{Verbatim}[
  fontsize=\small,
  breaklines=true,
  breakanywhere=true,
  breaksymbolleft={},
  breaksymbolright={}
]
First, determine if the following article is related to {topic} in any way.

If the article is NOT related to {topic}, respond with exactly: NA

If the article IS related to {topic}, provide a concise summary (3-4 sentences) focusing on:
1. Key facts and events
2. Forward-looking statements or predictions
Only use the provided article to create the summary.

Title: {article title}

Content:
{article content}

Response:
\end{Verbatim}
\end{tcolorbox}

\subsubsection{Prompt used to generate summaries from collection of articles}

\label{sec:media_cloud_daily}
\begin{tcolorbox}[
  colback=gray!10,
  colframe=gray!35,
  boxrule=0.4pt,
  arc=2pt,
  left=6pt,
  right=6pt,
  top=6pt,
  bottom=6pt,
  breakable
]
\small\ttfamily
\begin{Verbatim}[
  fontsize=\small,
  breaklines=true,
  breakanywhere=true,
  breaksymbolleft={},
  breaksymbolright={}
]
You are analyzing news articles from {date} to provide a comprehensive daily summary for financial forecasting.

Below are summaries of {N} articles from this date:

{Article 1 Summary: ....}
{Article 2 Summary: ....}
...
{Article N Summary: ....}

Based on these article summaries, provide a comprehensive daily summary with TWO sections:

SECTION 1 - KEY EVENTS AND FACTS:
Synthesize the main events, developments, and factual information from the day. Focus on market movements, economic data, corporate announcements, and significant news. (3-5 sentences)

SECTION 2 - FORWARD-LOOKING SIGNALS:
Identify and synthesize any forward-looking information, predictions, market expectations, or signals that could indicate future trends. Include analyst forecasts, anticipated policy changes, and emerging patterns. (3-5 sentences)

Please provide your response in exactly this format:

KEY EVENTS AND FACTS:
[Your synthesis here]

FORWARD-LOOKING SIGNALS:
[Your synthesis here]
\end{Verbatim}
\end{tcolorbox}

%% file: appendix/12_broader_impacts.tex
\section{Broader Societal Impacts}
\label{app:broader_impacts}
Our main contribution in this work is \textbf{Co}ntext as \textbf{Co}variates (CoCo), which is a novel forecasting pipeline that combines the strengths of LLMs and TSFMs to perform accurate multimodal time series forecasting. We also propose a novel approach to generate synthetic multimodal annotations, thereby enabling \textbf{sim-to-real} transfer in domains that lack multimodal time series datasets. We note that improved forecasting using exogenous text has a lot of positive  impacts in the time series domain, as it can enable accurate decision making based on precise forecasts. Additionally, CoCo unlocks the ability to perform accurate scenario simulation by providing speculative text annotations for the future. As such, the societal consequences of
our work are limited to those arising from improved quality of generated synthetic data. Apart from that, one of the limitations (Section \ref{sec:experiments}) of CoCo is that it is sensitive to incorrect or misleading text. In other words, if the text is deliberately misleading or false, CoCo can generate erroneous forecasts.

%% file: checklist.tex
\clearpage

\section*{NeurIPS Paper Checklist}

\begin{enumerate}

\item {\bf Claims}
    \item[] Question: Do the main claims made in the abstract and introduction accurately reflect the paper's contributions and scope?
    \item[] Answer: \answerYes{} 
    \item[] Justification: We prove the effectiveness of CoCo to interpret text accurately and improve multimodal forecasting in Section \ref{sec:experiments}. Additionally, we also showcase the effectiveness of training on synthetic multimodal data that is generated using the approach outlined in Section \ref{sec:dataset_gen} (see Table \ref{tab:text_in_finetuning}). 
    \item[] Guidelines:
    \begin{itemize}
        \item The answer \answerNA{} means that the abstract and introduction do not include the claims made in the paper.
        \item The abstract and/or introduction should clearly state the claims made, including the contributions made in the paper and important assumptions and limitations. A \answerNo{} or \answerNA{} answer to this question will not be perceived well by the reviewers. 
        \item The claims made should match theoretical and experimental results, and reflect how much the results can be expected to generalize to other settings. 
        \item It is fine to include aspirational goals as motivation as long as it is clear that these goals are not attained by the paper. 
    \end{itemize}

\item {\bf Limitations}
    \item[] Question: Does the paper discuss the limitations of the work performed by the authors?
    \item[] Answer: \answerYes{} 
    \item[] Justification: In Section \ref{sec:experiments}, we list three key limitations of our approach pertaining to inference latency, lack of evaluating the reliability of text, and limitations in benchmarks where numerical history has the least significance for forecasting.
    \item[] Guidelines:
    \begin{itemize}
        \item The answer \answerNA{} means that the paper has no limitation while the answer \answerNo{} means that the paper has limitations, but those are not discussed in the paper. 
        \item The authors are encouraged to create a separate ``Limitations'' section in their paper.
        \item The paper should point out any strong assumptions and how robust the results are to violations of these assumptions (e.g., independence assumptions, noiseless settings, model well-specification, asymptotic approximations only holding locally). The authors should reflect on how these assumptions might be violated in practice and what the implications would be.
        \item The authors should reflect on the scope of the claims made, e.g., if the approach was only tested on a few datasets or with a few runs. In general, empirical results often depend on implicit assumptions, which should be articulated.
        \item The authors should reflect on the factors that influence the performance of the approach. For example, a facial recognition algorithm may perform poorly when image resolution is low or images are taken in low lighting. Or a speech-to-text system might not be used reliably to provide closed captions for online lectures because it fails to handle technical jargon.
        \item The authors should discuss the computational efficiency of the proposed algorithms and how they scale with dataset size.
        \item If applicable, the authors should discuss possible limitations of their approach to address problems of privacy and fairness.
        \item While the authors might fear that complete honesty about limitations might be used by reviewers as grounds for rejection, a worse outcome might be that reviewers discover limitations that aren't acknowledged in the paper. The authors should use their best judgment and recognize that individual actions in favor of transparency play an important role in developing norms that preserve the integrity of the community. Reviewers will be specifically instructed to not penalize honesty concerning limitations.
    \end{itemize}

\item {\bf Theory assumptions and proofs}
    \item[] Question: For each theoretical result, does the paper provide the full set of assumptions and a complete (and correct) proof?
    \item[] Answer: \answerYes{} 
    \item[] Justification: We validate Assumptions \ref{assump:semantic} and \ref{assump:lipschitz} through real-world experiments. Lemma \ref{lemma:mismatch} and Theorem \ref{thm:transfer} are proved in Appendix \ref{appendix:theory_proofs}.
    \item[] Guidelines:
    \begin{itemize}
        \item The answer \answerNA{} means that the paper does not include theoretical results. 
        \item All the theorems, formulas, and proofs in the paper should be numbered and cross-referenced.
        \item All assumptions should be clearly stated or referenced in the statement of any theorems.
        \item The proofs can either appear in the main paper or the supplemental material, but if they appear in the supplemental material, the authors are encouraged to provide a short proof sketch to provide intuition. 
        \item Inversely, any informal proof provided in the core of the paper should be complemented by formal proofs provided in appendix or supplemental material.
        \item Theorems and Lemmas that the proof relies upon should be properly referenced. 
    \end{itemize}

    \item {\bf Experimental result reproducibility}
    \item[] Question: Does the paper fully disclose all the information needed to reproduce the main experimental results of the paper to the extent that it affects the main claims and/or conclusions of the paper (regardless of whether the code and data are provided or not)?
    \item[] Answer: \answerYes{} 
    \item[] Justification: All training and inference hyperparameters are clearly discussed in Appendix \ref{app:training_description}.
    \item[] Guidelines:
    \begin{itemize}
        \item The answer \answerNA{} means that the paper does not include experiments.
        \item If the paper includes experiments, a \answerNo{} answer to this question will not be perceived well by the reviewers: Making the paper reproducible is important, regardless of whether the code and data are provided or not.
        \item If the contribution is a dataset and\slash or model, the authors should describe the steps taken to make their results reproducible or verifiable. 
        \item Depending on the contribution, reproducibility can be accomplished in various ways. For example, if the contribution is a novel architecture, describing the architecture fully might suffice, or if the contribution is a specific model and empirical evaluation, it may be necessary to either make it possible for others to replicate the model with the same dataset, or provide access to the model. In general. releasing code and data is often one good way to accomplish this, but reproducibility can also be provided via detailed instructions for how to replicate the results, access to a hosted model (e.g., in the case of a large language model), releasing of a model checkpoint, or other means that are appropriate to the research performed.
        \item While NeurIPS does not require releasing code, the conference does require all submissions to provide some reasonable avenue for reproducibility, which may depend on the nature of the contribution. For example
        \begin{enumerate}
            \item If the contribution is primarily a new algorithm, the paper should make it clear how to reproduce that algorithm.
            \item If the contribution is primarily a new model architecture, the paper should describe the architecture clearly and fully.
            \item If the contribution is a new model (e.g., a large language model), then there should either be a way to access this model for reproducing the results or a way to reproduce the model (e.g., with an open-source dataset or instructions for how to construct the dataset).
            \item We recognize that reproducibility may be tricky in some cases, in which case authors are welcome to describe the particular way they provide for reproducibility. In the case of closed-source models, it may be that access to the model is limited in some way (e.g., to registered users), but it should be possible for other researchers to have some path to reproducing or verifying the results.
        \end{enumerate}
    \end{itemize}

\item {\bf Open access to data and code}
    \item[] Question: Does the paper provide open access to the data and code, with sufficient instructions to faithfully reproduce the main experimental results, as described in supplemental material?
    \item[] Answer: \answerYes{} 
    \item[] Justification: We have provided the code, data, and instructions to reproduce the results in Section \ref{sec:experiments}, as part of our supplementary material. Additionally, we have publicly released the code and data, with the links provided in the manuscript.
    \item[] Guidelines:
    \begin{itemize}
        \item The answer \answerNA{} means that paper does not include experiments requiring code.
        \item Please see the NeurIPS code and data submission guidelines (\url{https://neurips.cc/public/guides/CodeSubmissionPolicy}) for more details.
        \item While we encourage the release of code and data, we understand that this might not be possible, so \answerNo{} is an acceptable answer. Papers cannot be rejected simply for not including code, unless this is central to the contribution (e.g., for a new open-source benchmark).
        \item The instructions should contain the exact command and environment needed to run to reproduce the results. See the NeurIPS code and data submission guidelines (\url{https://neurips.cc/public/guides/CodeSubmissionPolicy}) for more details.
        \item The authors should provide instructions on data access and preparation, including how to access the raw data, preprocessed data, intermediate data, and generated data, etc.
        \item The authors should provide scripts to reproduce all experimental results for the new proposed method and baselines. If only a subset of experiments are reproducible, they should state which ones are omitted from the script and why.
        \item At submission time, to preserve anonymity, the authors should release anonymized versions (if applicable).
        \item Providing as much information as possible in supplemental material (appended to the paper) is recommended, but including URLs to data and code is permitted.
    \end{itemize}

\item {\bf Experimental setting/details}
    \item[] Question: Does the paper specify all the training and test details (e.g., data splits, hyperparameters, how they were chosen, type of optimizer) necessary to understand the results?
    \item[] Answer: \answerYes{} 
    \item[] Justification: Appendices \ref{app:training_description} and \ref{sec:eval_datasets} contain the exact training and testing related hyperparameters.
    \item[] Guidelines:
    \begin{itemize}
        \item The answer \answerNA{} means that the paper does not include experiments.
        \item The experimental setting should be presented in the core of the paper to a level of detail that is necessary to appreciate the results and make sense of them.
        \item The full details can be provided either with the code, in appendix, or as supplemental material.
    \end{itemize}

\item {\bf Experiment statistical significance}
    \item[] Question: Does the paper report error bars suitably and correctly defined or other appropriate information about the statistical significance of the experiments?
    \item[] Answer: \answerYes{} 
    \item[] Justification: In all the tables, we report the Wilcoxon p-value for statistical significance.
    \item[] Guidelines:
    \begin{itemize}
        \item The answer \answerNA{} means that the paper does not include experiments.
        \item The authors should answer \answerYes{} if the results are accompanied by error bars, confidence intervals, or statistical significance tests, at least for the experiments that support the main claims of the paper.
        \item The factors of variability that the error bars are capturing should be clearly stated (for example, train/test split, initialization, random drawing of some parameter, or overall run with given experimental conditions).
        \item The method for calculating the error bars should be explained (closed form formula, call to a library function, bootstrap, etc.)
        \item The assumptions made should be given (e.g., Normally distributed errors).
        \item It should be clear whether the error bar is the standard deviation or the standard error of the mean.
        \item It is OK to report 1-sigma error bars, but one should state it. The authors should preferably report a 2-sigma error bar than state that they have a 96\% CI, if the hypothesis of Normality of errors is not verified.
        \item For asymmetric distributions, the authors should be careful not to show in tables or figures symmetric error bars that would yield results that are out of range (e.g., negative error rates).
        \item If error bars are reported in tables or plots, the authors should explain in the text how they were calculated and reference the corresponding figures or tables in the text.
    \end{itemize}

\item {\bf Experiments compute resources}
    \item[] Question: For each experiment, does the paper provide sufficient information on the computer resources (type of compute workers, memory, time of execution) needed to reproduce the experiments?
    \item[] Answer: \answerYes{} 
    \item[] Justification: Appendix \ref{app:training_description} describes the required compute resources for the LLM fine-tuning experiments.
    \item[] Guidelines:
    \begin{itemize}
        \item The answer \answerNA{} means that the paper does not include experiments.
        \item The paper should indicate the type of compute workers CPU or GPU, internal cluster, or cloud provider, including relevant memory and storage.
        \item The paper should provide the amount of compute required for each of the individual experimental runs as well as estimate the total compute. 
        \item The paper should disclose whether the full research project required more compute than the experiments reported in the paper (e.g., preliminary or failed experiments that didn't make it into the paper). 
    \end{itemize}
    
\item {\bf Code of ethics}
    \item[] Question: Does the research conducted in the paper conform, in every respect, with the NeurIPS Code of Ethics \url{https://neurips.cc/public/EthicsGuidelines}?
    \item[] Answer: \answerYes{} 
    \item[] Justification: Yes, our research conforms with the NeurIPS Code of Ethics in every respect.
    \item[] Guidelines:
    \begin{itemize}
        \item The answer \answerNA{} means that the authors have not reviewed the NeurIPS Code of Ethics.
        \item If the authors answer \answerNo, they should explain the special circumstances that require a deviation from the Code of Ethics.
        \item The authors should make sure to preserve anonymity (e.g., if there is a special consideration due to laws or regulations in their jurisdiction).
    \end{itemize}

\item {\bf Broader impacts}
    \item[] Question: Does the paper discuss both potential positive societal impacts and negative societal impacts of the work performed?
    \item[] Answer: \answerYes{} 
    \item[] Justification: Our work mainly focuses on improving time series forecasting with multimodal context. As explained in the limitations (see Section \ref{sec:experiments}), we note that our approach utilizes the textual context without guardrails on how much the text should be trusted. We provided a detailed discussion regarding the same in Appendix \ref{app:broader_impacts}. 
    \item[] Guidelines:
    \begin{itemize}
        \item The answer \answerNA{} means that there is no societal impact of the work performed.
        \item If the authors answer \answerNA{} or \answerNo, they should explain why their work has no societal impact or why the paper does not address societal impact.
        \item Examples of negative societal impacts include potential malicious or unintended uses (e.g., disinformation, generating fake profiles, surveillance), fairness considerations (e.g., deployment of technologies that could make decisions that unfairly impact specific groups), privacy considerations, and security considerations.
        \item The conference expects that many papers will be foundational research and not tied to particular applications, let alone deployments. However, if there is a direct path to any negative applications, the authors should point it out. For example, it is legitimate to point out that an improvement in the quality of generative models could be used to generate Deepfakes for disinformation. On the other hand, it is not needed to point out that a generic algorithm for optimizing neural networks could enable people to train models that generate Deepfakes faster.
        \item The authors should consider possible harms that could arise when the technology is being used as intended and functioning correctly, harms that could arise when the technology is being used as intended but gives incorrect results, and harms following from (intentional or unintentional) misuse of the technology.
        \item If there are negative societal impacts, the authors could also discuss possible mitigation strategies (e.g., gated release of models, providing defenses in addition to attacks, mechanisms for monitoring misuse, mechanisms to monitor how a system learns from feedback over time, improving the efficiency and accessibility of ML).
    \end{itemize}
    
\item {\bf Safeguards}
    \item[] Question: Does the paper describe safeguards that have been put in place for responsible release of data or models that have a high risk for misuse (e.g., pre-trained language models, image generators, or scraped datasets)?
    \item[] Answer: \answerNA{} 
    \item[] Justification: We believe our work does not suffer from high-risk misuse.
    \item[] Guidelines:
    \begin{itemize}
        \item The answer \answerNA{} means that the paper poses no such risks.
        \item Released models that have a high risk for misuse or dual-use should be released with necessary safeguards to allow for controlled use of the model, for example by requiring that users adhere to usage guidelines or restrictions to access the model or implementing safety filters. 
        \item Datasets that have been scraped from the Internet could pose safety risks. The authors should describe how they avoided releasing unsafe images.
        \item We recognize that providing effective safeguards is challenging, and many papers do not require this, but we encourage authors to take this into account and make a best faith effort.
    \end{itemize}

\item {\bf Licenses for existing assets}
    \item[] Question: Are the creators or original owners of assets (e.g., code, data, models), used in the paper, properly credited and are the license and terms of use explicitly mentioned and properly respected?
    \item[] Answer: \answerYes{} 
    \item[] Justification: All datasets and models used in this work are properly credited via citations in the main paper. Specifically: (1) Qwen3-4b~\cite{qwen3} 
(\texttt{Qwen/Qwen3-4B-Instruct-2507}) is released under 
Apache 2.0; (2) Chronos2~\cite{chronos2} 
(\texttt{amazon/chronos-2}, version \texttt{2.1.0}) is released under Apache 2.0; 
(3) TimeMMD~\cite{timemmd} is released under CC BY 4.0; 
(4) FNSPID~\cite{fnspid} is released under CC BY-NC 4.0, 
and is used exclusively for research purposes consistent 
with its non-commercial license. Media Cloud data is 
accessed via the public Media Cloud API~\cite{mediacloud} 
in accordance with its terms of service. All model 
checkpoints are accessed from their official HuggingFace 
repositories.
    \item[] Guidelines:
    \begin{itemize}
        \item The answer \answerNA{} means that the paper does not use existing assets.
        \item The authors should cite the original paper that produced the code package or dataset.
        \item The authors should state which version of the asset is used and, if possible, include a URL.
        \item The name of the license (e.g., CC-BY 4.0) should be included for each asset.
        \item For scraped data from a particular source (e.g., website), the copyright and terms of service of that source should be provided.
        \item If assets are released, the license, copyright information, and terms of use in the package should be provided. For popular datasets, \url{paperswithcode.com/datasets} has curated licenses for some datasets. Their licensing guide can help determine the license of a dataset.
        \item For existing datasets that are re-packaged, both the original license and the license of the derived asset (if it has changed) should be provided.
        \item If this information is not available online, the authors are encouraged to reach out to the asset's creators.
    \end{itemize}

\item {\bf New assets}
    \item[] Question: Are new assets introduced in the paper well documented and is the documentation provided alongside the assets?
    \item[] Answer: \answerYes{} 
    \item[] Justification: Our paper introduces three new assets: (1) a synthetic multimodal training dataset, (2) a curated real-world evaluation dataset, and (3) CoCo fine-tuned TSFM and LLM checkpoints. We provide detailed documentation for all assets to ensure reproducibility. Appendix~\ref{sec:training_datasets} and Appendix~\ref{llm_prompts} describe the synthetic data generation pipeline, along with the prompts used. Appendix~\ref{sec:eval_datasets} provides details on the numerical time series and textual annotations used to curate the real-world evaluation dataset. Appendix~\ref{app:training_description} documents the fine-tuning procedures, hyperparameters, and training configurations for both the TSFM and LLM components. We additionally provide the fine-tuned checkpoints and inference pipeline as supplementary material, and have open-sourced the curated datasets (check the main manuscript for the links).
    \item[] Guidelines:
    \begin{itemize}
        \item The answer \answerNA{} means that the paper does not release new assets.
        \item Researchers should communicate the details of the dataset\slash code\slash model as part of their submissions via structured templates. This includes details about training, license, limitations, etc. 
        \item The paper should discuss whether and how consent was obtained from people whose asset is used.
        \item At submission time, remember to anonymize your assets (if applicable). You can either create an anonymized URL or include an anonymized zip file.
    \end{itemize}

\item {\bf Crowdsourcing and research with human subjects}
    \item[] Question: For crowdsourcing experiments and research with human subjects, does the paper include the full text of instructions given to participants and screenshots, if applicable, as well as details about compensation (if any)? 
    \item[] Answer: \answerNA{} 
    \item[] Justification: Our paper does not involve crowdsourcing nor research with human subjects.
    \item[] Guidelines:
    \begin{itemize}
        \item The answer \answerNA{} means that the paper does not involve crowdsourcing nor research with human subjects.
        \item Including this information in the supplemental material is fine, but if the main contribution of the paper involves human subjects, then as much detail as possible should be included in the main paper. 
        \item According to the NeurIPS Code of Ethics, workers involved in data collection, curation, or other labor should be paid at least the minimum wage in the country of the data collector. 
    \end{itemize}

\item {\bf Institutional review board (IRB) approvals or equivalent for research with human subjects}
    \item[] Question: Does the paper describe potential risks incurred by study participants, whether such risks were disclosed to the subjects, and whether Institutional Review Board (IRB) approvals (or an equivalent approval/review based on the requirements of your country or institution) were obtained?
    \item[] Answer: \answerNA{} 
    \item[] Justification: Our paper does not involve crowdsourcing nor research with human subjects.
    \item[] Guidelines:
    \begin{itemize}
        \item The answer \answerNA{} means that the paper does not involve crowdsourcing nor research with human subjects.
        \item Depending on the country in which research is conducted, IRB approval (or equivalent) may be required for any human subjects research. If you obtained IRB approval, you should clearly state this in the paper. 
        \item We recognize that the procedures for this may vary significantly between institutions and locations, and we expect authors to adhere to the NeurIPS Code of Ethics and the guidelines for their institution. 
        \item For initial submissions, do not include any information that would break anonymity (if applicable), such as the institution conducting the review.
    \end{itemize}

\item {\bf Declaration of LLM usage}
    \item[] Question: Does the paper describe the usage of LLMs if it is an important, original, or non-standard component of the core methods in this research? Note that if the LLM is used only for writing, editing, or formatting purposes and does \emph{not} impact the core methodology, scientific rigor, or originality of the research, declaration is not required.
    \item[] Answer: \answerYes{}
    \item[] Justification: Our work utilizes LLMs to understand and interpret the effect of text on time series forecasting. Our approach involves fine-tuning LLMs for multimodal forecasting. We have described the fine-tuning procedure in Section \ref{sec:model} and described all the prompts that we used in the Appendix.
    \item[] Guidelines:
    \begin{itemize}
        \item The answer \answerNA{} means that the core method development in this research does not involve LLMs as any important, original, or non-standard components.
        \item Please refer to our LLM policy in the NeurIPS handbook for what should or should not be described.
    \end{itemize}

\end{enumerate}